\documentclass[11pt]{article}

\usepackage[preprint]{acl}

\usepackage{times}
\usepackage{latexsym}
\usepackage{amsmath}

\usepackage{wasysym} 
\usepackage{amssymb} 
\usepackage[dvipsnames]{xcolor} 

\usepackage{booktabs}   
\usepackage{graphicx}  
\usepackage{array}     

\usepackage{tabularx}
\usepackage{xspace}
\usepackage{booktabs} 

\usepackage[table]{xcolor}
\usepackage{multirow}      

\usepackage[T1]{fontenc}

\usepackage[utf8]{inputenc}

\usepackage{microtype}

\usepackage{inconsolata}

\usepackage{graphicx}

\usepackage{longtable}
\usepackage{booktabs}

\newcommand{\method}{\textit{Population Aligned Language Models}\xspace}
\newcommand{\wums}{\textit{PALMs}\xspace}
\newcommand{\wum}{\textit{PALM}\xspace}

\title{PALMs: Using Multi Construct-Grounded Rationales for Modeling Population Preferences in LLMs}

\author{
Priyanka Dey\textsuperscript{$1$,$2$} \quad
Brihi Joshi\textsuperscript{$1$} \quad
Preyashi Poddar\textsuperscript{$1$} \quad
\textbf{Jieyu Zhao}\textsuperscript{$1$} \quad
\textbf{Emilio Ferrara}\textsuperscript{$1$,$2$} \\ 
\textsuperscript{$1$}University of Southern California \quad \textsuperscript{$2$}Information Sciences Institute \\
\texttt{\{deyp\}@usc.edu}
}

\begin{document}
\maketitle
\begin{abstract}

Large language models are being extensively used to simulate individual user behavior, yet faithfully representing a \textit{population} requires capturing the systematic variation in values, beliefs, and cultural norms that distinguish one group from another.
We introduce \method\ (\wums), a suite of models each aligned to specific populations, covering five countries: USA, India, Brazil, France and Italy.
\wums\ are created by synthesizing rationales grounded in psychological and cultural constructs and using these as latent supervision during preference tuning for population-specific alignment.
Evaluated across four dimensions: personality, values and beliefs, cultural norms, and morality, \wums\ consistently outperform baselines, including culture-specialized models, achieving an average of 8.59\% relative improvement over the best baseline across all five populations.
Notably, construct-grounded rationales outperform both demographic prompting and survey-based fine-tuning, suggesting that grounding preference learning in psychology and culture provides a richer inductive signal than surface-level response distributions.
We further demonstrate strong generalization to downstream applications without task-specific supervision: outperforming best baselines by 5.19\% in personalized reward modeling, 6.34\% in population simulation, and showing strong transfer to social reasoning tasks. Datasets and code are available at: \href{https://github.com/limenlp/PALMs} https://github.com/limenlp/PALMs.

\end{abstract}

\section{Introduction}

While LLMs serve billions of users worldwide, a core challenge in alignment remains: moving beyond a single, aggregate notion of human preference toward models that reflect the plurality of values and preferences across cultures \citep{sorensen2024roadmap, santurkar2023whose, aroyo2023dices, prabhakaran2022cultural, durmus2023towards}. User simulation --- modeling how specific users or populations respond to language --- has emerged as a key testbed, enabling applications such as personalized reward modeling, opinion prediction, and social reasoning \citep{argyle2023, wright2024llm, park2022social, krsteski2025validsurveysimulationslimited}. Yet existing frameworks condition models on shallow demographic attributes or short persona descriptions that fail to capture the values, beliefs, and cultural norms distinguishing one population from another \citep{santurkar2023whose, alkhamissi2024investigating, dominguez2024questioning, sun2025sociodemographic, gupta2024bias}. Fine-tuning directly on behavioral or survey data does not reliably solve this either, as surface-level optimization fails to preserve the richer compositional structure of human preferences \citep{ft_no_help, li2024culturellm, suh2025language, meister2025benchmarking}.

Human preferences are not flat outputs of demographics \citep{alipour2025robustness, torelli2009values, hwang2023aligning}. They emerge from the interaction of multiple psychological and cultural factors: personality traits \citep{john1999big}, cultural dimensions \citep{hofstede2001culture}, moral foundations \citep{graham2013moral}, personal values \citep{schwartz1992universals}, and world beliefs \citep{clifton2020testing}. While many such constructs exist, we focus on these five as they span individual-level psychology and group-level cultural variation, providing complementary lenses for reasoning about population-level preferences. Existing approaches either prompt models with demographic data or fine-tune on observed responses, neither of which forces LLMs to represent the underlying constructs that jointly shape human judgment.

In this work, \textit{population} refers to country-level human groups whose preferences reflect shared cultural, social, and behavioral patterns.\footnote{We acknowledge that meaningful variation exists within countries across regional, linguistic, and socioeconomic lines \citep{orlikowski2023ecological}; we adopt country-level abstraction as a practical and widely-used unit of analysis \citep{li2024culturellm, khanuja-etal-2024-image, liu2025can}.} We introduce \textbf{\method (\wums)}, a suite of LLMs each aligned to a specific population across five countries: USA, India, Brazil, France, and Italy. \wums are trained using \textit{multi-construct rationales}: synthetic rationales grounded in five psychological and cultural constructs --- personality traits (OCEAN; \citealt{goldberg1993structure}), cultural dimensions (Hofstede; \citealt{hofstede2001culture}), human values (Schwartz; \citealt{schwartz1992universals}), moral foundations \citep{graham2013moral}, and world beliefs \citep{clifton2020testing} --- as latent supervision during preference optimization via DPO \citep{rafailov2023direct}, with rationale tokens masked from the loss. At inference, \wums generate rationales before producing outputs, requiring no external construct annotation. Figure~\ref{fig:training_pipeline} shows the training and inference pipeline.

We evaluate \wums across four dimensions of population alignment --- personality, values and beliefs, cultural norms, and moral values --- by sampling users conditioned on demographic backgrounds drawn from each target population and aggregating their responses against ground-truth human distributions. \wums consistently outperform demographic prompting baselines and culture-specialized models across all five populations, achieving an average $8.59\%$ relative improvement over the best baseline. Notably, fine-tuning on survey data can underperform zero-shot baselines, suggesting that surface-level optimization may collapse rather than enrich population representations, whereas \wums avoid this by grounding preference reasoning in latent constructs. Beyond population alignment, \wums also show strong performance in three downstream applications without task-specific supervision: personalized reward modeling ($5.19\%$ improvement), population simulation ($6.34\%$ improvement), and social reasoning.

Our contributions are as follows: (1) We introduce \textbf{\method (\wums)}, a suite of population-aligned LLMs built on multi-construct rationales grounded in established psychological, cultural, and values theory. (2) We demonstrate that construct-grounded rationales outperform both demographic prompting and behavioral fine-tuning across five countries and four alignment dimensions. (3) We show that \wums generalize without task-specific supervision to several downstream tasks, establishing multi-construct rationales as a principled foundation for modeling populations.
\section{Related Work}

\paragraph{Persona-based user modeling.}
A central line of work in LLM personalization conditions model behavior on user personas — descriptions comprising demographic attributes, prior judgments, or interaction history~\citep{mazare2018training, madotto2019personalizing}. Zero-shot demographic prompting is conceptually appealing but produces inconsistent results: models tend to over-represent majority viewpoints and reflect known training biases~\citep{santurkar2023whose, hu2025generative}, and sociodemographic prompting does not consistently improve — and in some cases worsens — alignment with specific subpopulations~\citep{sun2025sociodemographic}. More recent approaches incorporate prior user judgments into the prompt~\citep{hwang2023aligning}. \citet{joshi2025improving}, \citet{koncel-kedziorski-etal-2025-primex} augment personas with psychologically scaffolded rationales to explain user judgments at inference time, and \citet{dey2026gravity} generate synthetic profile-grounded preference pairs using cultural and psychological frameworks to improve personalized content generation. Our work differs from both: rather than enriching inference-time inputs, we use construct-grounded reasoning as \textit{training-time} latent supervision, shaping the model's internal representations through masked scratchpads during preference optimization.
 
\paragraph{Cultural and population alignment.}
A growing body of work studies how well LLMs represent diverse human populations. \citep{santurkar2023whose} and \citep{argyle2023} show that LLMs can simulate survey responses but tend to over-represent majority viewpoints. Culture-specialized models such as CultureLLM~\citep{li2024culturellm} and CulturePark~\citep{li2024culturepark} train directly on large-scale cultural survey data to improve cross-cultural representation. \citep{meister2025benchmarking} and \citep{dey-etal-2025-llms} propose frameworks for evaluating LLM alignment with human personality and opinion distributions. Our work builds on this evaluation paradigm but identifies a failure mode these approaches share: optimizing directly on behavioral outputs narrows representational diversity rather than enriching it, a pattern we observe consistently across our experiments.
 
\paragraph{Reasoning as training supervision.} Chain-of-thought prompting~\citep{wei2022chain} and its variants have shown that eliciting intermediate reasoning steps improves LLM performance across a range of tasks. STaR~\citep{zelikman2022star} and related methods use model-generated rationales as self-improvement signals during training. Scratchpad-based approaches~\citep{nye2021show} similarly use intermediate computation as latent structure. In preference learning, \citet{lightman2024let} show that process-level supervision outperforms outcome-level supervision for mathematical reasoning. Our approach is in this spirit but targets a different goal: we use reasoning not to improve task accuracy, but to inject structured psychological knowledge into preference representations, masking it from the optimization objective so it functions purely as inductive bias.

\paragraph{Psychological constructs in language models.} Humans routinely use everyday explanations to rationalize each other's actions---a practice known as folk psychology~\citep{malle2006mind, churchland2013folk}. Such explanations are prevalent enough in text corpora that LLMs can generate plausible psychological rationales zero-shot~\citep{binz2023using}, making synthetic construct-grounded reasoning chains a tractable source of structured supervision. Prior work has applied individual constructs to NLP tasks: Big Five traits have been used to model user behavior~\citep{goldberg1993structure}, Schwartz values to study opinion formation~\citep{schwartz1992universals, kiesel2022identifying}, and moral foundations to analyze political discourse~\citep{haidt2004intuitive}. \method unifies these into a single scaffold space, using them jointly as lenses for reasoning about preference decisions.
\section{\method}

\begin{figure*}[!t]
\centering
\includegraphics[width=\textwidth]{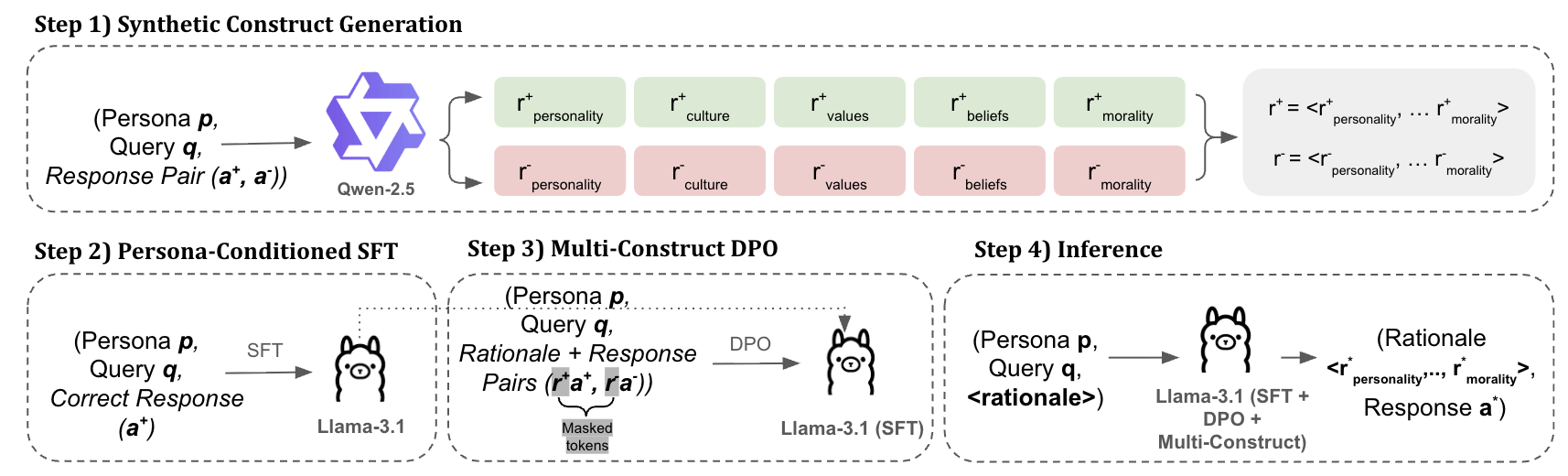}
\caption{Overview of the \method training and inference pipeline. We generate multi-construct post-hoc rationales grounded in $\mathcal{C}$ (see Table~\ref{tab:construct_families}), then run SFT on persona-conditioned preference instances. The SFT model is further optimized with DPO on rationale-augmented preference pairs, with rationale tokens masked from the loss and used only as structured context. At inference, given a persona, query, and \texttt{<rationale>} token, the model generates a construct-grounded rationale followed by a preference prediction. }
\label{fig:training_pipeline}
\end{figure*}

\subsection{Problem Formulation}

Consider a user from a country (e.g., USA, India), described by an individual demographic persona $p$ (e.g. their ethnicity, age, gender, education level, and political leaning). Given a query $q$ paired with two candidate responses $a^{+}$ and $a^{-}$, our goal is to predict the user's preferred response $a^{*}$. 

Standard preference learning optimizes a direct mapping $(p, q) \to a^{*}$, conditioning only on surface-level demographic descriptors. \method instead introduces an intermediate multi-construct rationale $r$, decomposing each decision across five psychological and cultural constructs (\S\ref{sec:construct_space}) before predicting the preferred response:
$
(p, q) \;\to\; r \;\to\; a^{*}.$
The rationale $r$ is supplied as latent supervision during training and masked from the loss, so the model internalizes construct-grounded rationale without learning to generate the rationale token-by-token. At inference time, the model generates $r*$ before producing $a^{*}$, requiring no external annotation. 

Generating this rationale $r$ allows the model to capture the psychological and cultural mechanisms behind a preference, rather than fitting surface-level correlations between demographic descriptors and outputs. A direct $(p, q) \to a^{*}$ mapping does not provide a mechanism to represent \textit{why} a person with a given background prefers one response over another. By generating $r$ first, the model reasons through an individual's personality traits, values, moral intuitions, and world beliefs, situating them within the broader cultural tendencies that shape how those traits are expressed~\citep{sagiv2022personal, vauclair2015kinds}. 

\vspace{-0.75pt}

\subsection{Construct Space}
\label{sec:construct_space}

We ground our rationales in five families of established social science theory that together span individual-level psychology and group-level cultural variation. 
These construct families act as interpretive lenses through which the model reasons about each preference decision (See Table~\ref{tab:construct_families}).

\begin{table}[t]
\centering
\scriptsize 
\setlength{\tabcolsep}{2pt} 
\begin{tabular}{@{} >{\raggedright\arraybackslash}p{0.4\columnwidth} >{\raggedright\arraybackslash}p{0.6\columnwidth} @{}}
\toprule
\textbf{Construct Family} & \textbf{Description} \\
\midrule
Personality (OCEAN)\newline\citep{goldberg1993structure} & Individual-level traits (e.g., openness, conscientiousness) that shape communication and decision style. \\
\addlinespace[4pt]
Cultural dimensions (Hofstede)\newline\citep{hofstede2001culture} & Group-level tendencies (e.g., individualism, power distance) that govern how people relate to authority, community, and social expectations. \\
\addlinespace[4pt]
Human values (Schwartz)\newline\citep{schwartz1992universals} & Individual-level motivational priorities (e.g., conservation, self-transcendence) that drive trade-offs between competing choices. \\
\addlinespace[4pt]
Primal world beliefs\newline\citep{clifton2020testing} & Individual-level assumptions about the world (e.g., safe vs.\ dangerous) that determine situational interpretation. \\
\addlinespace[4pt]
Moral foundations\newline\citep{haidt2004intuitive} & Individual-level moral intuitions (e.g., fairness, authority) that govern ethical evaluation. \\
\bottomrule
\end{tabular}
\caption{Construct families spanning the latent space $\mathcal{C}$.}
\label{tab:construct_families}
\end{table}

\begin{figure*}[!t]
\centering
\includegraphics[width=\textwidth]{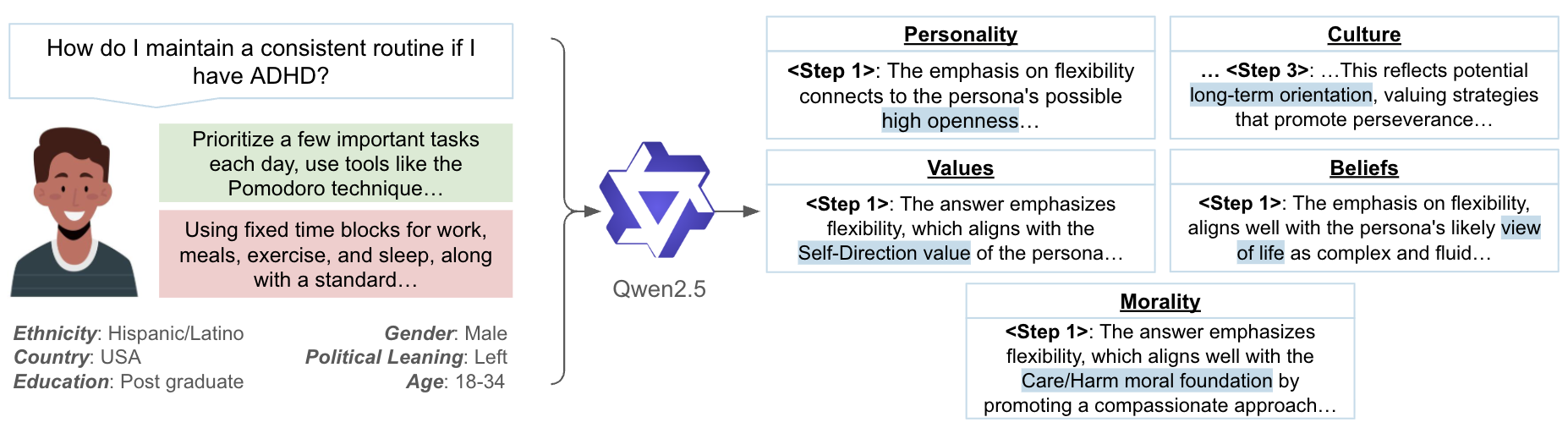}
\caption{During training, we prompt \texttt{Qwen2.5-7B-Instruct} to produce synthetic construct-grounded rationales for each preference instance. Each rationale decomposes the preference decision across five psychological and cultural construct families, providing structured label-conditioned supervision during DPO optimization.}
\label{fig:reasoning_chains}
\end{figure*}

Figure~\ref{fig:reasoning_chains} illustrates an example synthetic rationale generated from these five families for a user's preference instance.

\subsection{Construct-Grounded Rationale Generation}
\label{sec:scratchpad_generation}
For each training instance $(p, q, a^{+}, a^{-})$, we prompt an LLM $f_{\phi}$ to generate 
construct-grounded rationales that explain the observed preference as a function of the user's 
demographic profile. These post-hoc rationalizations provide a justification for \textit{why} 
a user with profile $p$ might plausibly favor $a^{+}$ over $a^{-}$ through each construct 
lens---not inferred ground truths about the user's inner states, but plausible explanations 
consistent with the observed demographics and choice. For each construct $C_k \in \mathcal{C}$ 
and each candidate response $a \in \{a^{+}, a^{-}\}$, we condition $f_{\phi}$ on the 
demographics, query, candidate response, and preference label 
$\ell \in \{\text{chosen}, \text{rejected}\}$ to produce a construct-specific reasoning 
chain $r^{(k)}$, each consisting of 4--6 steps. The full rationale concatenates all 
construct-specific chains:
$r = \langle\, r^{\textsc{pers}},\; r^{\textsc{cult}},\;
              r^{\textsc{val}},\; r^{\textsc{mor}},\;
              r^{\textsc{bel}} \,\rangle
$.

\subsection{Training}
\label{sec:training}
We train \method in two stages, with a separate model per country (see Fig~\ref{fig:training_pipeline}).

\paragraph{Stage 1: Supervised Fine-Tuning.}
We fine-tune \texttt{Llama-3.1-8B-Instruct} on preference instances $(p, q, a^{+})$ 
with a standard causal LM objective to initialize the model on persona-conditioned 
preference prediction before preference optimization.

\paragraph{Stage 2: Preference Optimization.}
We then augment each preference pair with the rationales from \S\ref{sec:scratchpad_generation}, 
where $r^{+}$ rationalizes $a^{+}$ as the chosen response and $r^{-}$ rationalizes $a^{-}$ 
as the rejected one:
$
c^{\pm} = \langle \texttt{<rationale>}\; r^{\pm}\; \texttt{</rationale>},\; a^{\pm} \rangle$.
We optimize $c^{+}$ and $c^{-}$ using Direct Preference Optimization 
(DPO;~\citealt{rafailov2023direct}), masking all rationale tokens from the loss and compute loss over over $a^{+}$ and $a^{-}$. As multiple rationales are equally plausible 
for any given instance, optimizing directly over them would overfit the model to arbitrary 
explanations. Masking ensures the rationales instead shape preference learning as structured 
context rather than explicit prediction targets.

\paragraph{Inference.}
During inference, \wums are prompted with the \texttt{<rationale>} token and generate a construct-grounded rationale before producing a final preference decision. Only the final output is used for evaluation. 

\begin{table*}[t]
\centering
\scriptsize 
\renewcommand{\arraystretch}{1.1} 
\setlength{\tabcolsep}{6pt} 

\begin{tabular*}{\textwidth}{c | l | c | @{\extracolsep{\fill}} cccc @{}}
\toprule
\tiny\textbf{Model} & \tiny\textbf{Country} & \tiny\textbf{Overall} $\uparrow$ & \tiny\textbf{Personality} $\uparrow$ & \tiny\textbf{Culture} $\uparrow$ & \tiny\textbf{Values} $\uparrow$ & \tiny\textbf{Morality} $\uparrow$ \\ 
\midrule
\rowcolor[gray]{0.92} \multicolumn{7}{l}{\textbf{Baselines}} \\
\multirow{5}{*}{Llama-3.1-8B-Instruct} & USA    & $0.522_{\pm 0.014}$ & $0.541_{\pm 0.007}$ & $\underline{0.527}_{\pm 0.029}$ & $0.519_{\pm 0.046}$ & $0.502_{\pm 0.004}$ \\
                                       & India  & $0.490_{\pm 0.033}$ & $0.487_{\pm 0.010}$ & $\underline{0.475}_{\pm 0.123}$ & $0.524_{\pm 0.044}$ & $0.475_{\pm 0.004}$ \\
                                       & Brazil & $0.468_{\pm 0.030}$ & $0.431_{\pm 0.017}$ & $\underline{0.436}_{\pm 0.108}$ & $0.511_{\pm 0.044}$ & $0.492_{\pm 0.004}$ \\
                                       & France & $0.491_{\pm 0.018}$ & $0.508_{\pm 0.014}$ & $0.495_{\pm 0.054}$ & $0.532_{\pm 0.044}$ & $0.428_{\pm 0.004}$ \\
                                       & Italy  & $0.487_{\pm 0.016}$ & $0.524_{\pm 0.020}$ & $\underline{0.455}_{\pm 0.046}$ & $0.516_{\pm 0.040}$ & $0.452_{\pm 0.004}$ \\
\midrule
\multirow{5}{*}{Gemini-2.5-Flash}      & USA    & $\underline{0.642}_{\pm 0.025}$ & $\underline{0.644}_{\pm 0.006}$ & $0.507_{\pm 0.029}$ & $\underline{0.630}_{\pm 0.094}$ & $\underline{0.787}_{\pm 0.004}$ \\
                                       & India  & $\underline{0.646}_{\pm 0.032}$ & $\underline{0.719}_{\pm 0.008}$ & $0.390_{\pm 0.120}$ & $\underline{0.760}_{\pm 0.046}$ & $\underline{0.717}_{\pm 0.004}$ \\
                                       & Brazil & $\underline{0.592}_{\pm 0.030}$ & $\underline{0.550}_{\pm 0.016}$ & $0.397_{\pm 0.106}$ & $\underline{0.694}_{\pm 0.054}$ & $\textbf{\underline{0.726}}_{\pm 0.004}$ \\
                                       & France & $\underline{0.634}_{\pm 0.019}$ & $\underline{0.640}_{\pm 0.013}$ & $0.480_{\pm 0.054}$ & $\textbf{\underline{0.735}}_{\pm 0.051}$ & $\textbf{\underline{0.681}}_{\pm 0.004}$ \\
                                       & Italy  & $\underline{0.643}_{\pm 0.017}$ & $\underline{0.709}_{\pm 0.015}$ & $0.401_{\pm 0.045}$ & $\underline{0.710}_{\pm 0.048}$ & $\underline{0.754}_{\pm 0.004}$ \\
\midrule
\multirow{5}{*}{CultureLLM}            & USA    & $0.480_{\pm 0.012}$ & $0.499_{\pm 0.006}$ & $0.513_{\pm 0.029}$ & $0.479_{\pm 0.040}$ & $0.430_{\pm 0.004}$ \\
                                       & India  & $0.452_{\pm 0.031}$ & $0.568_{\pm 0.009}$ & $0.373_{\pm 0.119}$ & $0.454_{\pm 0.036}$ & $0.413_{\pm 0.004}$ \\
                                       & Brazil & $0.415_{\pm 0.029}$ & $0.427_{\pm 0.017}$ & $0.359_{\pm 0.104}$ & $0.474_{\pm 0.044}$ & $0.400_{\pm 0.004}$ \\
                                       & France & $0.457_{\pm 0.016}$ & $0.478_{\pm 0.015}$ & $\underline{0.517}_{\pm 0.053}$ & $0.421_{\pm 0.036}$ & $0.413_{\pm 0.004}$ \\
                                       & Italy  & $0.436_{\pm 0.015}$ & $0.506_{\pm 0.019}$ & $0.414_{\pm 0.045}$ & $0.420_{\pm 0.035}$ & $0.405_{\pm 0.004}$ \\
\midrule
\rowcolor[gray]{0.92} \multicolumn{7}{l}{\textbf{PALMs}} \\
\multirow{5}{*}{SFT}                  & USA    & $0.655_{\pm 0.009}$ & $0.590_{\pm 0.010}$ & $0.650_{\pm 0.022}$ & $0.620_{\pm 0.024}$ & $0.760_{\pm 0.008}$ \\
                                       & India  & $0.620_{\pm 0.030}$ & $0.560_{\pm 0.012}$ & $0.550_{\pm 0.115}$ & $0.690_{\pm 0.032}$ & $0.680_{\pm 0.007}$ \\
                                       & Brazil & $0.558_{\pm 0.026}$ & $0.490_{\pm 0.014}$ & $0.410_{\pm 0.098}$ & $0.660_{\pm 0.030}$ & $0.670_{\pm 0.008}$ \\
                                       & France & $0.580_{\pm 0.014}$ & $0.590_{\pm 0.011}$ & $0.460_{\pm 0.048}$ & $0.660_{\pm 0.028}$ & $0.610_{\pm 0.009}$ \\
                                       & Italy  & $0.563_{\pm 0.014}$ & $0.490_{\pm 0.016}$ & $0.420_{\pm 0.042}$ & $0.650_{\pm 0.031}$ & $0.690_{\pm 0.007}$ \\
\midrule
\multirow{5}{*}{SFT + DPO}            & USA    & $0.683_{\pm 0.008}$ & $0.610_{\pm 0.009}$ & $0.700_{\pm 0.020}$ & $0.640_{\pm 0.022}$ & $0.780_{\pm 0.007}$ \\
                                       & India  & $0.653_{\pm 0.029}$ & $0.580_{\pm 0.011}$ & $0.600_{\pm 0.112}$ & $0.720_{\pm 0.030}$ & $0.710_{\pm 0.006}$ \\
                                       & Brazil & $0.588_{\pm 0.025}$ & $0.520_{\pm 0.013}$ & $0.440_{\pm 0.095}$ & $0.690_{\pm 0.028}$ & $0.700_{\pm 0.007}$ \\
                                       & France & $0.613_{\pm 0.013}$ & $0.620_{\pm 0.010}$ & $0.500_{\pm 0.045}$ & $0.690_{\pm 0.026}$ & $0.640_{\pm 0.008}$ \\
                                       & Italy  & $0.590_{\pm 0.013}$ & $0.510_{\pm 0.015}$ & $0.450_{\pm 0.040}$ & $0.680_{\pm 0.029}$ & $0.720_{\pm 0.006}$ \\
\midrule
\multirow{5}{*}{SFT + DPO + Generic}  & USA    & $0.699_{\pm 0.008}$ & $0.622_{\pm 0.008}$ & $0.729_{\pm 0.019}$ & $0.650_{\pm 0.021}$ & $0.795_{\pm 0.006}$ \\
                                       & India  & $0.676_{\pm 0.028}$ & $0.590_{\pm 0.010}$ & $0.644_{\pm 0.108}$ & $0.740_{\pm 0.029}$ & $\textbf{0.730}_{\pm 0.005}$ \\
                                       & Brazil & $0.600_{\pm 0.024}$ & $0.530_{\pm 0.012}$ & $0.455_{\pm 0.092}$ & $0.700_{\pm 0.026}$ & $0.715_{\pm 0.006}$ \\
                                       & France & $0.632_{\pm 0.012}$ & $0.638_{\pm 0.009}$ & $0.520_{\pm 0.042}$ & $0.715_{\pm 0.024}$ & $0.656_{\pm 0.007}$ \\
                                       & Italy  & $0.610_{\pm 0.012}$ & $0.526_{\pm 0.014}$ & $0.470_{\pm 0.038}$ & $0.700_{\pm 0.027}$ & $0.745_{\pm 0.005}$ \\
\midrule
\multirow{5}{*}{SFT + DPO + Multi-Construct} & USA    & $\textbf{0.726}_{\pm 0.008}$ & $\textbf{0.656}_{\pm 0.006}$ & $\textbf{0.772}_{\pm 0.024}$ & $\textbf{0.672}_{\pm 0.018}$ & $\textbf{0.802}_{\pm 0.004}$ \\
                                       & India  & $\textbf{0.738}_{\pm 0.026}$ & $\textbf{0.728}_{\pm 0.008}$ & $\textbf{0.714}_{\pm 0.102}$ & $\textbf{0.788}_{\pm 0.025}$ & $0.722_{\pm 0.004}$ \\
                                       & Brazil & $\textbf{0.618}_{\pm 0.023}$ & $\textbf{0.565}_{\pm 0.012}$ & $\textbf{0.490}_{\pm 0.088}$ & $\textbf{0.708}_{\pm 0.024}$ & $0.708_{\pm 0.004}$ \\
                                       & France & $\textbf{0.647}_{\pm 0.011}$ & $\textbf{0.650}_{\pm 0.009}$ & $\textbf{0.558}_{\pm 0.038}$ & $0.730_{\pm 0.022}$ & $0.650_{\pm 0.004}$ \\
                                       & Italy  & $\textbf{0.703}_{\pm 0.011}$ & $\textbf{0.720}_{\pm 0.012}$ & $\textbf{0.608}_{\pm 0.035}$ & $\textbf{0.715}_{\pm 0.024}$ & $\textbf{0.771}_{\pm 0.004}$ \\
\bottomrule
\end{tabular*}
\caption{In-domain alignment results across five target populations. Higher scores are better; subscripts denote bootstrapped 95\% confidence intervals. \textbf{Bold} indicates best result per country--dimension; \underline{underlines} indicate best baseline. \method consistently achieves the highest overall alignment, with each training stage contributing incrementally. \texttt{CultureLLM} frequently underperforms zero-shot baselines, suggesting that survey-based fine-tuning does not reliably improve population alignment.}
\label{tab:baseline_comparison}
\end{table*}

\section{Experiments}
\label{sec:experiments}

 
\paragraph{Training Details.}
We train \method on the Community Alignment dataset \cite{zhang2025cultivating}, a large-scale preference dataset designed capturing population-grounded human judgments. 
Each sample consists of a user persona with associated demographic attributes, a query, and a pair of candidate responses with a preferred choice. 
We augment each preference pair with synthetic construct-grounded rationales generated by \texttt{Qwen2.5-7B-Instruct} as described in \S\ref{sec:scratchpad_generation}.
We train five separate models, one per target population: USA, India, Brazil, France, and Italy (see App.~\ref{app:training_details} for dataset statistics and training details). 
We train five separate models, one per target population: USA, India, Brazil, France, and Italy (see App.~\ref{app:training_details} for dataset statistics and training details), initialized from \texttt{Llama-3.1-8B-Instruct}~\citep{grattafiori2024llama} and trained following the two-step pipeline in \S\ref{sec:training}.

 
\paragraph{In-Domain Evaluation.}
We evaluate population alignment across four dimensions: \textit{personality, 
culture, values and beliefs}, and \textit{morality}. 
For each dimension, we select an established survey instrument and measure alignment between model response distributions (via sampling users by conditioning on demographic backgrounds) and ground-truth human distributions from the target population. 

For \textit{personality}, we follow \citet{dey-etal-2025-llms} and \citet{meister2025benchmarking}.
We prompt each \wum to produce response distributions over items from the IPIP-120 questionnaire~\citep{goldberg1999ipip}. 
We measure alignment via Wasserstein distance averaged across the five OCEAN traits. 
For \textit{culture}, we use open-ended cultural norm questions from 
CultureBank~\citep{shi2024culturebank} and measure alignment via 
entailment accuracy using GPT-4o~\cite{hurst2024gpt} as an LLM judge~\cite{zheng2023judging}. 
For \textit{values and beliefs} and \textit{morality}, we use opinion questions from the Pew Global Attitudes Survey \cite{pew2023globalattitudes} and the World Values Survey~\citep{haerpfer2022world} respectively, measuring population-level alignment via $L_1$ distance between model and human response distributions over unordered categorical options.
We normalize all metrics to $[0, 1]$ and apply a $1 - \text{score}$ transformation to lower-is-better metrics (Wasserstein, $L_1$) before averaging, so that higher scores consistently reflect better alignment across all four dimensions.
 
\paragraph{Out-of-Domain Evaluation.}
To assess whether construct-grounded representations generalize beyond the training objective, we evaluate \method on three downstream applications where modeling diverse human perspectives is critical but existing approaches remain limited: (1)~\textit{personalized reward modeling}, (2)~\textit{population simulation}, and (3)~\textit{social intelligence}.

For \textit{personalized reward modeling}, general-purpose reward models are not tailored to specific user populations, yet collecting group-specific preference data is often infeasible. 
We evaluate on \textit{PersonalizedRewardBench}~\citep{ma2026personalized}, a personalized reward benchmark spanning datasets across varied topics like lifestyle, culture, and entertainment, and \textit{PRISM}~\citep{kirk2024prism}, a globally diverse preference dataset focused on values and opinions. 
We benchmark \wums against Skywork-Reward-V2~\citep{liu2025skywork}, an 8B model fine-tuned specifically for reward modelling and a top performer on RewardBench~\citep{lambert2025rewardbench}, as a strong baseline.

For \textit{population simulation}, LLMs are increasingly used to simulate human behavior, but struggle to capture variation across real human populations. 
We evaluate on \textsc{SimBench}~\citep{hu2025simbench}, a large-scale benchmark spanning opinion, personality, moral reasoning, humor, and decision-making. 
We report SimBench's normalized score derived from Total Variation Distance, where 100 indicates perfect alignment, 0 equals random guessing, and negative scores indicate worse-than-random performance.

For \textit{social intelligence}, interpretations of social situations vary systematically across cultural norms, values, and personality --- yet existing benchmarks largely treat social reasoning as uniform across individuals. 
We conduct a case study using \wum-USA on \textsc{Social IQa}~\citep{sap2019social}, a benchmark for reasoning about human intent, emotion, and interpersonal expectations.\footnote{We restrict this evaluation to the USA population as \textsc{Social IQa} does not provide cross-cultural annotations for our other target populations, making a controlled single-population analysis the most meaningful comparison.}
\vspace{-0.5pt}
\paragraph{Baselines.}
All models receive demographics per user: ethnicity, age group, gender, education, and political leaning, as provided by Community Alignment. We use \texttt{Llama-3.1-8B-Instruct} and \texttt{Gemini-2.5-Flash} as open- and closed-source zero-shot baselines. We also include \texttt{CultureLLM}~\citep{li2024culturellm}, a culture-specialized model trained on large-scale human survey data.

To isolate the contribution of construct-grounded rationales, we compare four configurations: \textit{SFT}, fine-tuned on the community dataset alone; \textit{SFT+DPO}, which adds preference optimization without rationales; \textit{SFT+DPO+Generic}, which incorporates post-hoc rationales \textit{without} construct grounding; and \method, the full model. For out-of-domain evaluation, we restrict comparison to \method and \textit{SFT+DPO+Generic}.

\vspace{-1pt} 
\section{Results}
\label{sec:results}

\begin{figure}[!t]
\centering
\includegraphics[width=1\columnwidth]{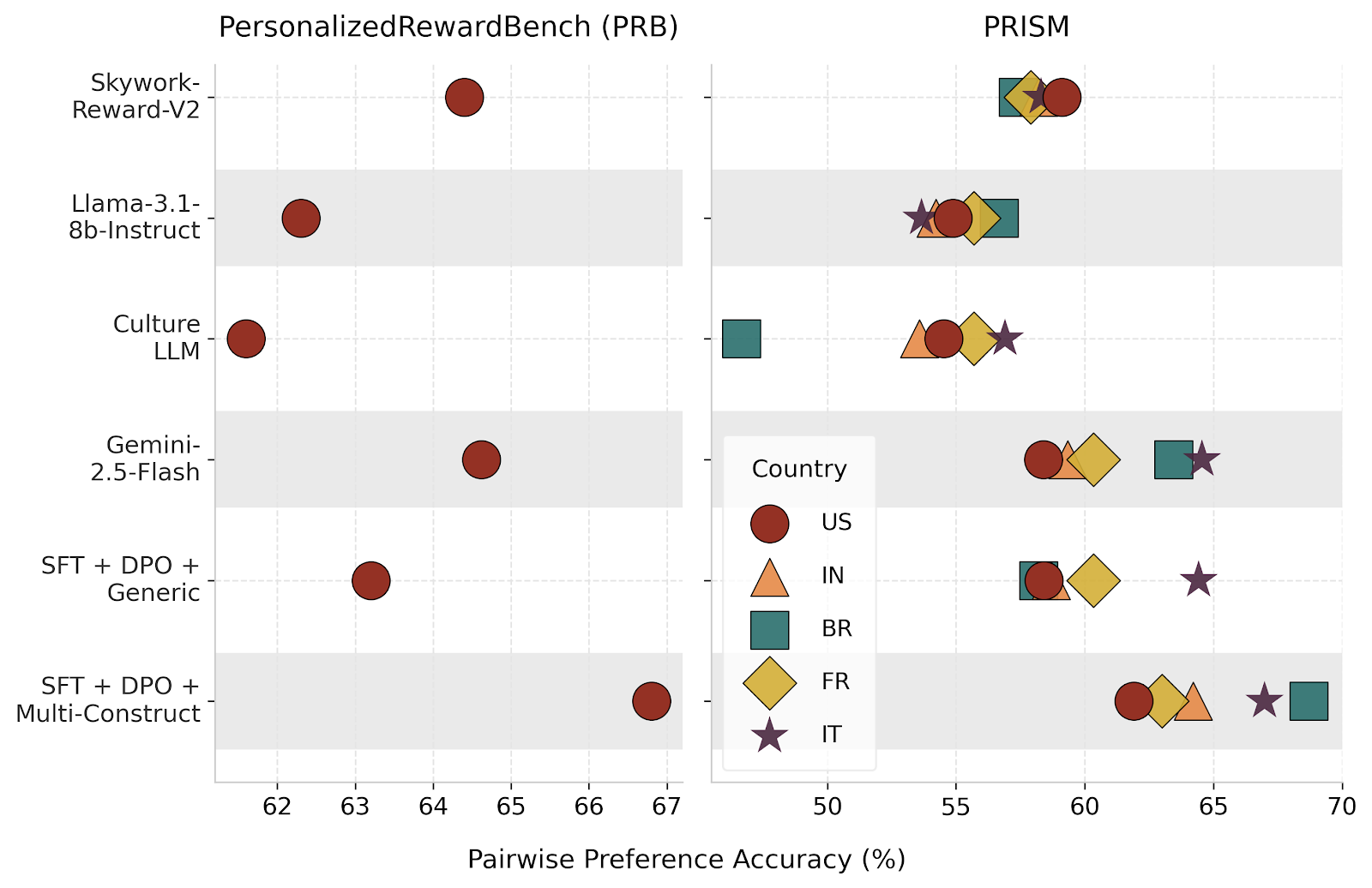}
\caption{Pairwise preference prediction accuracy (\%) on  \textit{PersonalizedRewardBench} (PRB) and \textit{PRISM}, aggregated by country. \wums outperforms baselines across both benchmarks and all five countries.} 
\label{fig:reward_models}
\end{figure}
 
\subsection{In-Domain Population Alignment}

Table~\ref{tab:baseline_comparison} presents results on in-domain
population alignment across all five target countries.

\paragraph{\wums achieves the strongest alignment across populations and dimensions.} \wums consistently achieves the highest overall alignment across all countries, outperforming both baselines and intermediate training configurations. Within the constructs used, \textit{Culture} shows the largest relative improvement over baselines, while \textit{personality} shows the most consistent gains across the training progression. \textit{Morality} is the most variable construct, where intermediate models occasionally match \wums, suggesting that moral alignment may require construct scaffolds beyond those captured by our current taxonomy\footnote{Unlike personality, values, and cultural norms, moral judgments are highly sensitive to situational framing~\citep{haidt2004intuitive}, such that demographic and construct-level signals may be insufficient to capture within-population moral variation.}.

\paragraph{Construct-grounded rationales provide the strongest alignment signal at each training stage.} Moving from \textit{SFT} to \textit{SFT+DPO} yields consistent but modest gains, showing that contrastive preference optimization sharpens population-specific alignment beyond supervised training alone. Adding generic rationales (\textit{SFT+DPO+Generic}) produces a further improvement across most dimensions, showing that structured intermediate reasoning helps even without psychological grounding. The gap between \textit{SFT+DPO+Generic} and \wums isolates what construct grounding specifically contributes: for the USA, culture alignment improves from $0.729$ to $0.772$ and personality from $0.622$ to $0.656$, with similar patterns across all five populations, suggesting that the construct families provide a decomposition vocabulary that generic reasoning alone cannot replicate.

\vspace{-0.5pt}

\paragraph{Fitting directly to user behavioral data does not always
improve population alignment.} Models optimized on user-level survey responses without construct grounding show diminishing returns and in some cases underperform zero-shot baselines, particularly on culture and morality. Directly fitting observed response distributions does not reliably capture the diversity of human preferences across populations. Rationalizing through latent psychological and cultural constructs instead appears to provide a more robust path to alignment across diverse populations.

\begin{figure}[!t]
\centering
\includegraphics[width=1\columnwidth]{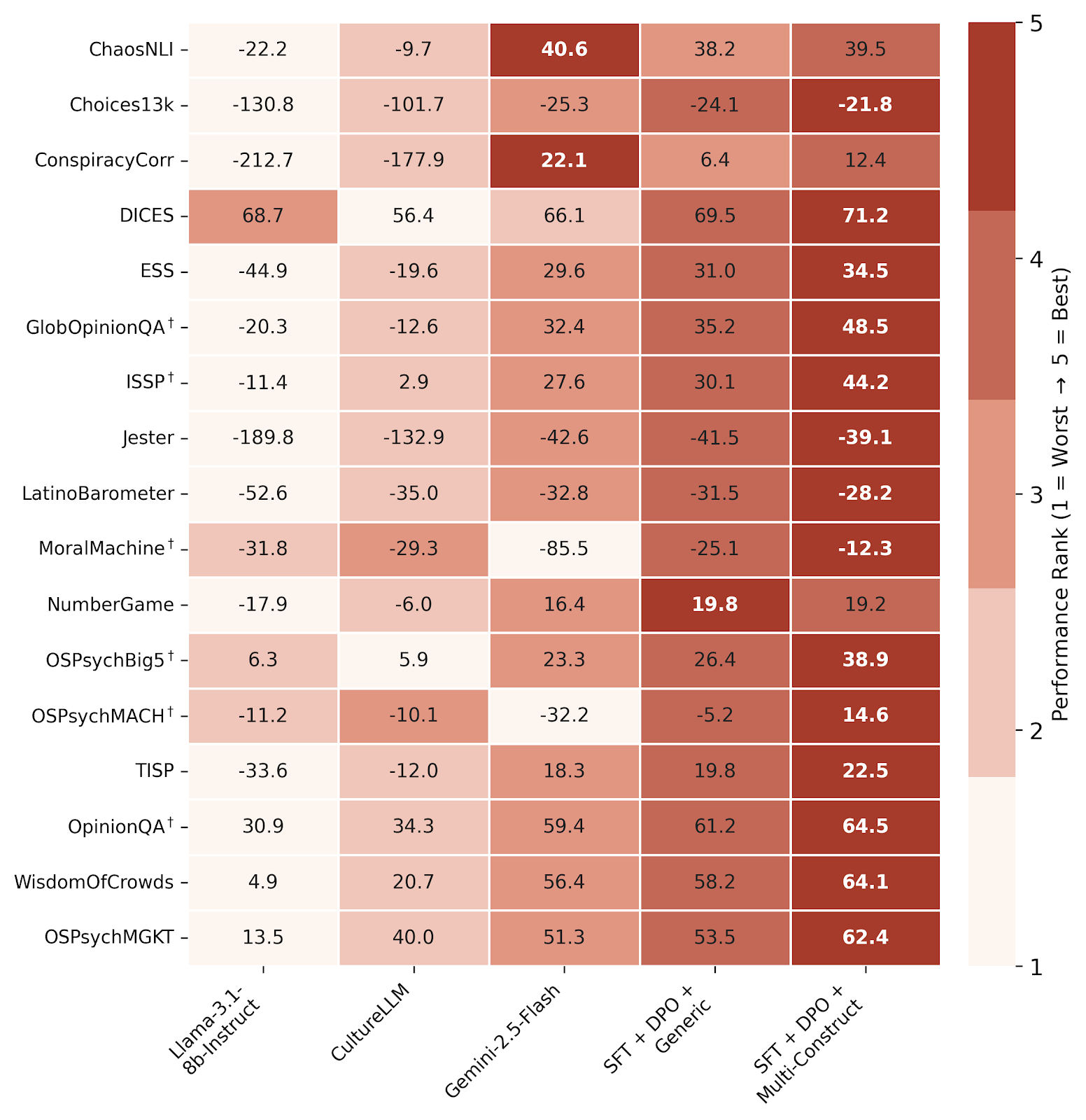}
\caption{Population simulation results on \textsc{SimBench}, macro-averaged over countries (color indicates best performance rank). $^\dagger$ denotes tasks aligned with construct families (personality, culture, values, and morality). \wums outperforms baselines on opinion, personality, and value-aligned tasks.} 

\label{fig:simbench}
\end{figure}
 
\subsection{\textit{Application 1}: Personalized Reward Modeling}
 
 
\paragraph{Construct-grounded rationales transfer to personalized reward modeling without any task-specific supervision.} As shown in Fig~\ref{fig:reward_models}, \wums outperforms all baselines across both benchmarks and all five countries. On PRB, \wums reaches $66.89\%$ against Skywork's $64.40\%$. We further observe consistent gains across PRB topic categories (Table~\ref{tab:prb_breakdown}), with the largest improvements on society-related topics spanning religious beliefs, law, philosophy, and politics --- domains which directly link to world beliefs, values, and morality. With construct rationales, \wums achieves a $5.2\%$ gain over baselines.

We also observe strong performance gains on PRISM for non-US populations, particularly Brazil ($68.70\%$ vs.\ $63.45\%$ for Gemini) and India ($64.20\%$ vs.\ $59.34\%$), suggesting that construct-grounded rationales can be especially effective at capturing preference variation across culturally diverse groups. These results suggest that \wums can serve as an effective personalized reward model, generalizing to preference prediction across diverse cultural groups without any task-specific training.

\subsection{\textit{Application 2}: Population Simulation}

\paragraph{Construct-grounded rationales generalize to behavioral tasks well outside the training distribution.} As shown in Fig~\ref{fig:simbench}, \wums achieves the best simulation alignment on 15 out of 17 \textsc{SimBench} datasets, spanning opinion, personality, moral reasoning, and decision-making. Gains are largest on construct-aligned tasks ($^\dagger$); our full method shows large improvements over generic rationales by $13.3\%$ on \textit{OpinionQA}, $14.1\%$ on \textit{ISSP}, and nearly $20\%$ on both \textit{OSPsychMACH} and \textit{MoralMachine}. These results highlight that construct-grounded rationales help capture latent structure in how people make ethical trade-offs under uncertainty and how deeper personality dimensions, such as Machiavellianism, shape preference judgments.

Beyond construct-aligned tasks, \wums also generalizes to subjective human judgment tasks. On \textit{DICES}, which captures demographic variability in safety assessments of conversational AI responses, \wums achieves $3\%$ gains over best baselines, suggesting that construct-grounded representations of human diversity also transfer to safety perception judgments. Similarly, on \textit{TISP}, which measures trust in science and science-related populism, \wums yields a $4\%$ improvement, indicating that value and belief-grounded rationales generalize to perception-based tasks beyond those directly covered by our construct families. On crowd knowledge tasks such as \textit{WisdomOfCrowds} and \textit{OSPsychMGKT}, \wums also shows consistent gains of $6\%$ and $9\%$, indicating that construct rationales help in tasks relying on aggregated human expertise. 


\begin{figure}[!t]
\centering
\includegraphics[width=1\columnwidth]{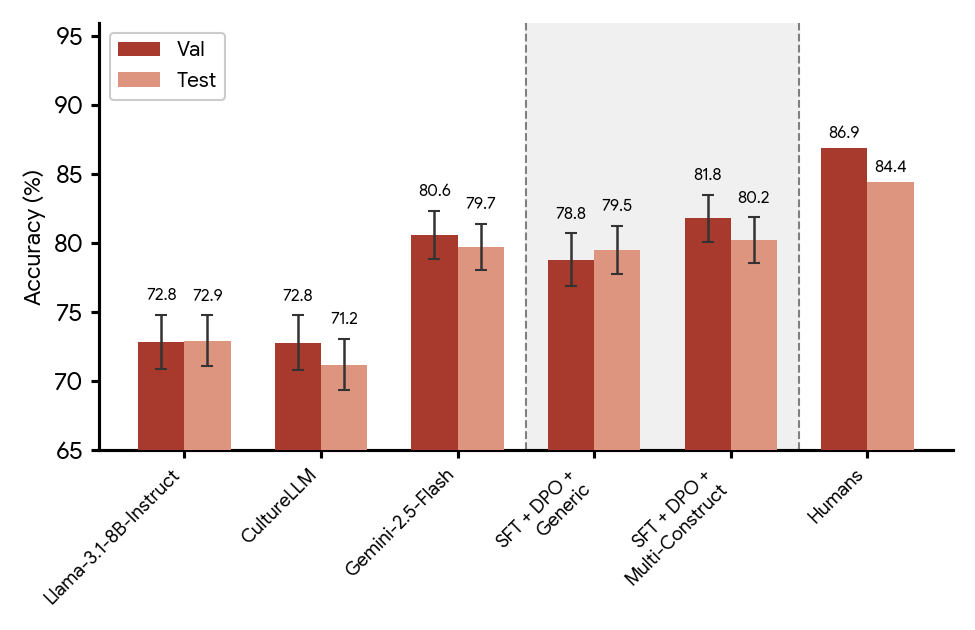}
\caption{Zero-shot performance (Accuracy \% w/ 95\% CI) on \textsc{SocialIQA} validation and test data splits. \wums shows strong transfer to social intelligence tasks.}
\label{fig:social_iqa}
\end{figure} 
 
\subsection{\textit{Case Study}: Social Intelligence}
\label{subsec:social_intelligence}
  
\paragraph{Construct-grounded rationales improves social intelligence, with particular gains on understanding user intents and motivations.}
Table~\ref{fig:social_iqa} presents zero-shot performance on \textsc{SocialIQa} as a case study of whether construct-grounded rationales generalizes to social intelligence beyond the preference learning objective. \textsc{SocialIQa} requires reasoning about human intent, emotion, and interpersonal expectations, making it a natural test of whether construct-grounded rationales transfer beyond preference prediction. \wums achieves $80.23\%$ on test, outperforming all baselines including Gemini-2.5-Flash ($79.72\%$) despite being an 8B model with no task-specific supervision. \textit{CultureLLM}, struggles to perform well on this task ($71.18\%$), consistent with patterns observed in in-domain evaluation where models trained on user-level responses struggle to generalize. 

We further observe performance gains in various social intellgence question categories (see Tables~\ref{tab:socialiqa_val} and \ref{tab:socialiqa_test}). \wums shows the largest gains on categories that require reasoning about human intent and motivation, for example, improving over the strongest baseline by $3.33\%$ on \textit{motivations}, suggesting that construct-grounded rationales captures something meaningful interpretations about why people behave as they do. Although models still fail to achieve human-level performances ($84.40\%$), these results suggest that rationales transfer to broader social reasoning in models.

\section{Conclusions}

In this work, we introduced \method\ (\wums), a framework for population-level modeling that uses construct-grounded rationales as latent supervision during preference optimization. Our central finding is that rationalizing through established psychological and cultural constructs provides a richer inductive signal for preference learning than demographic conditioning or survey-based training data alone. Across five culturally diverse populations, \wums\ consistently improves population alignment, and these gains transfer to personalized reward modeling, population simulation, and social reasoning without any task-specific supervision. We hope this work encourages the broader adoption of social science theory as structured inductive bias in alignment research, and opens new directions in multilingual, fine-grained, population modeling.

\section{Limitations}


\paragraph{Country as a Proxy for Population.}
Following a long line of NLP work that uses country-level annotations as a proxy for cultural population~\citep{li2024culture, khanuja-etal-2024-image, liu2025can}, we adopt country-level abstraction as our unit of analysis. While countries are not monoliths, this granularity is motivated by both conceptual and practical considerations: nations represent long-standing units of shared history, institutional influence, and social norms, and country-level geo-tagging remains the most common form of demographic partitioning in large-scale datasets. That said, this abstraction inevitably flattens meaningful within-country variation — regional, linguistic, and socioeconomic differences that a single population model cannot capture. Future work should explore finer-grained population definitions, including regional and individual-level modeling.

\paragraph{Choice of Psychological Constructs.}
The five construct families used in this work --- personality traits (OCEAN), cultural dimensions (Hofstede), human values (Schwartz), moral foundations, and primal world beliefs, represent a principled but non-exhaustive selection from a broader landscape of psychological and cultural theory. Other well-established frameworks \cite{howe2012attachment, brockner2001regulatory} may capture additional dimensions of human preference that our current construct space does not cover. The choice of these five was motivated by their complementary coverage of individual-level psychology and group-level cultural variation, as well as their prevalence in prior NLP work. Future work should explore whether alternative or expanded construct sets yield further improvements in population alignment.

\paragraph{Single Dataset.}
\method is trained and evaluated on the Community Alignment dataset, which, while large-scale and culturally diverse, represents one particular operationalization of human preference. Other opinion-based resources, such as social media corpora or alternative survey datasets, may surface different dimensions of population variation. Future work can further explore broader data sources and alternative representations of user identity beyond structured demographic profiles.

\paragraph{Multilingual Modeling.}
To isolate the effect of construct-grounded reasoning on population alignment without introducing language as a confounding variable, we translated all non-English training instances to English. This means \method does not model native-language preference expression, which may itself carry culturally meaningful signal. Extending \wums to natively multilingual settings is an important direction for future work.

\paragraph{Model Scale and Architecture.}
All \method models are built on \texttt{Llama-3.1-8B-Instruct}. While our results demonstrate that multi-construct rationales are effective at this scale, it remains an open question how construct-grounded preference learning interacts with larger models or different base architectures. Future work should examine whether these gains scale and generalize across model families.

\clearpage

\bibliography{custom}

\clearpage
\appendix
\label{sec:appendix}

\section{Dataset Statistics}
\label{app:dataset_stats}

\subsection{Training Data}

We construct preference pairs from country-specific opinion data from the Community Alignment dataset, with one \method instance trained per population. Table~\ref{tab:training_stats} summarizes the size of the training set for each country. Each preference pair consists of a persona, a query, a chosen response, and a rejected response, augmented with a multi-construct rationale (\S\ref{sec:construct_space}) for \method{} and a generic rationale for the \textsc{Generic} baseline. Non-English instances 
(Hindi, Portuguese, French, and Italian) were translated to English using \texttt{Qwen2.5-7B-Instruct} to allow for controlled comparisons across populations without introducing language as a confounding variable; we leave multilingual modeling as future work.

\begin{table}[h]
\centering
\small
\begin{tabular}{lr}
\toprule
\textbf{Population} & \textbf{\# Preference Pairs} \\
\midrule
USA    & 105{,}999 \\
India  & 192{,}696 \\
Brazil &  94{,}380 \\
France &  92{,}348 \\
Italy  &  97{,}252 \\
\bottomrule
\end{tabular}
\caption{Training preference pairs per population.}
\label{tab:training_stats}
\end{table}

\subsection{Evaluation Datasets}

We evaluate \method{} along four dimensions of population alignment: personality, cultural norms, values/beliefs, and morality. Table~\ref{tab:eval_stats} summarizes the number of evaluation items per population for each benchmark.

\paragraph{Personality.} We use the IPIP-120 inventory, which contains 120 items spanning the five OCEAN traits (24 items per trait). The same 120 items are used across all populations.

\paragraph{Culture.} We use country-specific subsets of CultureBank~\citep{shi2024culturebank}; we filter by demographic groups native to each of the five countries. Item counts vary substantially across populations (USA: 1169, Brazil: 78, India: 59, Italy: 459, and France: 329); we therefore report \textbf{macro-averaged} accuracy across populations to avoid letting the larger subsets dominate the aggregate score.

\paragraph{Values and beliefs.} We aggregate questions from the PEW Global Attitudes Survey across multiple years (2018--2024), filtering to country-specific items. This results in: 86 questions for USA, 51 for Brazil, 74 for France, 51 for India, and 71 for Italy. 

\paragraph{Morality.} We use questions directly from Wave 7 which are linked with Morality (total: 23). These items were not used to train each CultureLLM model.

\begin{table}[h]
\centering
\small
\begin{tabular}{lrrrr}
\toprule
\textbf{Population} & \textbf{IPIP-120} & \textbf{CultureBank} & \textbf{PEW} & \textbf{Morality} \\
\midrule
USA    & 120 & 1{,}169 & 86 & 23 \\
India  & 120 &     59  & 51 & 23 \\
Brazil & 120 &     78  & 51 & 23 \\
France & 120 &    329  & 74 & 23 \\
Italy  & 120 &    459  & 71 & 23 \\
\bottomrule
\end{tabular}
\caption{Number of evaluation items per population across the four population-alignment benchmarks.}
\label{tab:eval_stats}
\end{table}

\section{Training Details}
\label{app:training_details}

\subsection{Compute}

All runs are performed on NVIDIA RTX A6000 GPUs (48GB) and NVIDIA A100 (80GB) GPUs. The \textsc{CultureLLM} baseline uses LoRA~\citep{hu2022lora} to fine-tune \texttt{Llama-3.1-70B-Instruct} on a single A100 (approximately 1 hour). For \method, we first train full-parameter SFT of \texttt{Llama-3.1-8B-Instruct} on 4 RTX A6000 GPUs, taking approximately 12--14 hours per population. DPO is performed on 2$\times$ NVIDIA A100 (80GB) GPUs and takes approximately 12--14 hours per population.

\subsection{SFT}

We perform full-parameter supervised fine-tuning of \texttt{Llama-3.1-8B-Instruct} for 2 epochs with the AdamW optimizer, learning rate $2 \times e^{-5}$ with linear warmup (3\%), weight decay 0.01, and bf16 precision. We use a per-device batch size of 4 with gradient accumulation of 2 and a maximum sequence length of 1024. Training uses DeepSpeed ZeRO with gradient checkpointing.

\subsection{DPO}

After SFT, we run Direct Preference Optimization~\citep{rafailov2023direct} on the same preference pairs for 2 epochs, with multi-construct rationales included in the prompt context and masked from the loss objective. We use sigmoid DPO with $\beta = 0.1$ and label smoothing 0, the AdamW optimizer with learning rate $2 \times e^{-6}$ on a cosine schedule with 10\% warmup, weight decay 0.05, and bf16 precision. We use a per-device batch size of 2 (training) / 4 (evaluation) with gradient accumulation of 4. Training uses DeepSpeed ZeRO with gradient checkpointing. 

\section{Rationale Generation Prompts}
\label{app:prompts}

We generate multi-construct rationales by prompting an LLM with a shared system message and a construct-specific user message. The system message provides the JSON output format, the persona, the question--answer pair, the ground-truth label, and the general reasoning instructions. The user message then specifies the construct $\psi$ that the rationale should be grounded in. Table~\ref{tab:rationale_prompts} lists all six variants used in our experiments: a baseline (no construct) and one for each of the five construct families.

\section{Example Synthetic Rationales for Training}
\label{app:example_rationales}

To illustrate the qualitative difference between generic rationales and multi-construct rationales used during training, we present one preference instance per population (USA, Brazil, India, France, Italy) along with both rationale types for chosen and rejected answers. Each multi-construct rationale decomposes the preference decision across the five construct families introduced in \S\ref{sec:construct_space}: personality (OCEAN), cultural dimensions (Hofstede), human values (Schwartz), primal world beliefs, and moral foundations. Tables~\ref{tab:example_usa}--\ref{tab:example_italy} list the full instance and rationales for each population.

\section{Human Validation of Rationale Quality}
\label{app:human_eval}

To validate the quality of our generated rationales, we conduct a human evaluation study with two annotators who are well-versed in the five psychological and cultural construct families used in this work.

\paragraph{Sample.} We randomly sample 50 preference instances, stratified across the five populations (10 per country). For each instance, annotators see the persona, query, candidate answers, and six rationales for both the chosen and rejected answers: one generic rationale and one rationale per construct family (personality, cultural dimensions, human values, primal world beliefs, and moral foundations). This yields a total of $50 \times 6 \times 2 = 600$ rationale ratings per annotator.

\paragraph{Rating rubric.} Annotators rate each rationale on a 5-point Likert scale according to how well it invokes its assigned construct (or, for generic rationales, how well it captures relevant aspects of the persona):
\begin{itemize}\itemsep0pt
  \item \textbf{1}: The construct is not invoked, or is misapplied in a way that contradicts the persona.
  \item \textbf{2}: The construct is mentioned superficially without grounding in the persona's traits or context.
  \item \textbf{3}: The construct is invoked correctly but the reasoning is shallow or partially generic.
  \item \textbf{4}: The construct is invoked correctly and substantively connected to the persona.
  \item \textbf{5}: The construct is accurately and richly invoked, with reasoning clearly grounded in the persona.
\end{itemize}

\paragraph{Inter-annotator agreement.} Table~\ref{tab:kappa} reports Cohen's $\kappa$ between the two annotators, computed separately for each rationale type. All scores fall in the substantial-agreement range, with an average $\kappa$ of $0.66$ across the six rationale types. The slightly lower agreement for values ($0.61$) and culture ($0.62$) reflects the greater subjectivity in deciding when a Schwartz value or Hofstede dimension is ``substantively'' invoked, while morality shows the highest agreement ($0.71$), consistent with the more clearly delineated boundaries of the moral foundations.

\begin{table}[h]
\centering
\small
\begin{tabular}{lc}
\toprule
\textbf{Rationale type} & \textbf{Cohen's $\kappa$} \\
\midrule
Generic              & 0.66 \\
Personality (OCEAN)  & 0.68 \\
Cultural dimensions  & 0.62 \\
Human values         & 0.61 \\
Primal world beliefs & 0.68 \\
Moral foundations    & 0.71 \\
\midrule
\textbf{Average}     & \textbf{0.66} \\
\bottomrule
\end{tabular}
\caption{Inter-annotator agreement (Cohen's $\kappa$) on the 5-point quality rubric, computed separately per rationale type.}
\label{tab:kappa}
\end{table}

\section{Example Model-Generated Rationales at Test Time}
\label{app:test_time_rationales}

To complement the synthetic training-time rationales shown in \S\ref{app:example_rationales}, we present examples of rationales \emph{generated by the trained models at inference time}. For each population, we show one held-out preference instance along with the rationale produced by two trained models: \textsc{Generic-\method} (trained with single-construct generic rationales) and \method{} (trained with multi-construct rationales). The same instance is used for both rationale types to facilitate direct comparison. Tables~\ref{tab:test_usa}--\ref{tab:test_italy} list the full instance and rationales for each population. These examples illustrate that the inductive bias from training-time rationale supervision carries over to inference: \method{} produces decompositions across the five construct families even though it is given only the persona, and query at test time.

\section{Additional Results}
\label{app:additional_results}

We provide finer-grained breakdowns of the three downstream evaluations summarized in the main paper: personalized reward modeling (PRB), population simulation (\textsc{SimBench}), and social reasoning (\textsc{SocialIQa}).

\subsection{Personalized Reward Modeling: Domain Breakdown}

Table~\ref{tab:prb_breakdown} reports PRB accuracy broken down by the three top-level domain categories: \textit{Art}, \textit{Lifestyle}, and \textit{Society}. \method{} achieves the best score in every category, with the largest margin on \textit{Society} (+4.45 points over the strongest non-\method{} baseline), where preference judgments most heavily depend on cultural and moral reasoning. The \textsc{SFT + DPO + Generic} ablation underperforms even the zero-shot \texttt{Llama-3.1-8B-Instruct} baseline across all domains, consistent with our broader finding that behavioral fine-tuning without construct-grounded rationales can collapse rather than enrich preference representations.

\begin{table*}[ht]
\centering
\scriptsize
\setlength{\tabcolsep}{5pt}
\renewcommand{\arraystretch}{1.15}
\begin{tabular}{lcccc}
\toprule
\textbf{Model} & \textbf{Art} & \textbf{Lifestyle} & \textbf{Society} & \textbf{Overall} \\
\midrule
Skywork-Reward-V2     & $63.10_{\pm 3.25}$ & $65.20_{\pm 2.75}$ & $64.90_{\pm 2.60}$ & $64.40_{\pm 1.68}$ \\
Llama-3.1-8B-Instruct & $60.63_{\pm 3.45}$ & $65.02_{\pm 2.97}$ & $61.24_{\pm 2.91}$ & $62.29_{\pm 1.79}$ \\
CultureLLM            & $60.15_{\pm 3.46}$ & $65.86_{\pm 2.95}$ & $58.97_{\pm 2.94}$ & $61.64_{\pm 1.79}$ \\
Gemini-2.5-Flash      & $63.15_{\pm 3.19}$ & $65.33_{\pm 2.68}$ & $64.90_{\pm 2.53}$ & $64.46_{\pm 1.60}$ \\
SFT + DPO + Generic   & $52.10_{\pm 3.20}$ & $54.40_{\pm 2.85}$ & $53.25_{\pm 2.90}$ & $53.25_{\pm 1.75}$ \\
\midrule
\textbf{\method{} (Ours)} & $\mathbf{65.20}_{\pm 3.10}$ & $\mathbf{66.12}_{\pm 2.60}$ & $\mathbf{69.35}_{\pm 2.45}$ & $\mathbf{66.89}_{\pm 1.65}$ \\
\bottomrule
\end{tabular}
\caption{Personalized reward modeling accuracy (\%) on the PRB benchmark, broken down by domain category. Values are mean $\pm$ bootstrapped 95\% confidence interval margin.}
\label{tab:prb_breakdown}
\end{table*}

\subsection{Population Simulation: Per-Dataset Breakdown}

Table~\ref{tab:simbench_results} reports per-dataset \textsc{SimBench} scores, macro-averaged across countries. Scores are signed (higher is better; $0$ corresponds to random; negative values indicate worse-than-random simulation). \method{} achieves the best score on the majority of datasets and is competitive on the remainder. The largest gains occur on construct-aligned datasets ($\dagger$): on \textsc{OpinionQA}, \textsc{GlobOpinionQA}, \textsc{ISSP}, \textsc{OSPsychBig5}, and \textsc{OSPsychMACH}, \method{} substantially outperforms all baselines---most notably on \textsc{OSPsychMACH}, where every other method scores below random while \method{} reaches $14.6$. This pattern supports our central claim: when downstream tasks probe the same psychological and cultural constructs that shape \method's training-time rationales, the construct-grounded representation transfers strongly. On non-construct-aligned datasets such as \textsc{ChaosNLI} and \textsc{ConspiracyCorr}, \method{} performs comparably to \textsc{Gemini-2.5-Flash} despite using a much smaller base model.

\begin{table*}[ht]
\centering
\scriptsize
\setlength{\tabcolsep}{4pt}
\renewcommand{\arraystretch}{1.1}
\begin{tabular}{lrrrrr}
\toprule
\textbf{Dataset} &
\textbf{Llama-3.1-8B} &
\textbf{CultureLLM} &
\textbf{Gemini-2.5-Flash} &
\textbf{SFT+DPO+Generic} &
\textbf{\method{} (Ours)} \\
\midrule
ChaosNLI                  & $-22.16_{\pm 7.23}$    & $-9.65_{\pm 6.35}$     & $\mathbf{40.64}_{\pm 4.83}$  & $38.20_{\pm 4.50}$  & $39.50_{\pm 4.10}$ \\
Choices13k                & $-130.76_{\pm 10.83}$  & $-101.70_{\pm 11.38}$  & $-25.27_{\pm 7.18}$          & $-24.10_{\pm 6.80}$ & $\mathbf{-21.80}_{\pm 6.20}$ \\
ConspiracyCorr            & $-212.74_{\pm 71.83}$  & $-177.86_{\pm 52.30}$  & $\mathbf{22.08}_{\pm 31.23}$ & $6.40_{\pm 28.50}$  & $12.40_{\pm 25.10}$ \\
DICES                     & $68.66_{\pm 4.26}$     & $56.42_{\pm 5.03}$     & $66.12_{\pm 4.06}$           & $69.50_{\pm 3.80}$  & $\mathbf{71.20}_{\pm 3.50}$ \\
ESS                       & $-44.94_{\pm 32.91}$   & $-19.55_{\pm 28.29}$   & $29.56_{\pm 14.74}$          & $31.00_{\pm 13.50}$ & $\mathbf{34.50}_{\pm 12.10}$ \\
GlobOpinionQA$^\dagger$   & $-20.34_{\pm 16.00}$   & $-12.59_{\pm 15.45}$   & $32.39_{\pm 10.02}$          & $35.20_{\pm 9.50}$  & $\mathbf{48.50}_{\pm 8.20}$ \\
ISSP$^\dagger$            & $-11.42_{\pm 16.09}$   & $2.94_{\pm 15.02}$     & $27.59_{\pm 12.49}$          & $30.10_{\pm 11.20}$ & $\mathbf{44.20}_{\pm 9.80}$ \\
Jester                    & $-189.81_{\pm 1.37}$   & $-132.94_{\pm 11.86}$  & $-42.61_{\pm 6.35}$          & $-41.50_{\pm 5.80}$ & $\mathbf{-39.10}_{\pm 5.10}$ \\
LatinoBarometer           & $-52.60_{\pm 26.82}$   & $-35.03_{\pm 25.15}$   & $-32.83_{\pm 26.56}$         & $-31.50_{\pm 22.40}$ & $\mathbf{-28.20}_{\pm 19.80}$ \\
MoralMachine$^\dagger$    & $-31.75_{\pm 13.27}$   & $-29.34_{\pm 12.17}$   & $-85.54_{\pm 15.56}$         & $-25.10_{\pm 11.00}$ & $\mathbf{-12.30}_{\pm 9.50}$ \\
NumberGame                & $-17.87_{\pm 8.38}$    & $-5.99_{\pm 6.16}$     & $16.43_{\pm 6.14}$           & $\mathbf{19.80}_{\pm 5.50}$ & $19.20_{\pm 4.90}$ \\
OSPsychBig5$^\dagger$     & $6.28_{\pm 11.18}$     & $5.93_{\pm 9.68}$      & $23.28_{\pm 9.49}$           & $26.40_{\pm 8.50}$  & $\mathbf{38.90}_{\pm 7.20}$ \\
OSPsychMACH$^\dagger$     & $-11.15_{\pm 24.78}$   & $-10.08_{\pm 38.55}$   & $-32.16_{\pm 27.35}$         & $-5.20_{\pm 22.10}$ & $\mathbf{14.60}_{\pm 18.50}$ \\
TISP                      & $-33.61_{\pm 22.88}$   & $-11.96_{\pm 20.23}$   & $18.34_{\pm 16.09}$          & $19.80_{\pm 14.50}$ & $\mathbf{22.50}_{\pm 12.80}$ \\
OpinionQA$^\dagger$       & $30.88_{\pm 3.68}$     & $34.31_{\pm 3.20}$     & $59.35_{\pm 2.28}$           & $61.20_{\pm 2.10}$  & $\mathbf{64.50}_{\pm 1.85}$ \\
WisdomOfCrowds            & $4.87_{\pm 11.88}$     & $20.74_{\pm 9.57}$     & $56.35_{\pm 6.72}$           & $58.20_{\pm 6.20}$  & $\mathbf{64.10}_{\pm 5.50}$ \\
OSPsychMGKT               & $13.50_{\pm 17.16}$    & $40.00_{\pm 12.98}$    & $51.32_{\pm 10.21}$          & $53.50_{\pm 9.50}$  & $\mathbf{62.40}_{\pm 8.20}$ \\
\bottomrule
\end{tabular}
\caption{Population simulation on \textsc{SimBench}. Higher is better; $0$ = random; negative = worse than random. Scores are macro-averaged across countries and also include bootstrapped 95\% confidence intervals. $\dagger$ denotes construct-aligned datasets.}
\label{tab:simbench_results}
\end{table*}

\subsection{Social Reasoning: \textsc{SocialIQa} Breakdown}

Tables~\ref{tab:socialiqa_val} and~\ref{tab:socialiqa_test} report fine-grained zero-shot accuracy on the \textsc{SocialIQa} validation and test splits across its five reasoning categories: \textit{Wants}, \textit{Reactions}, \textit{Descriptions}, \textit{Motivations}, and \textit{Needs}. \method{} achieves the best score on every category and overall, narrowing the gap to the human baseline ($86.9$ on validation, $84.4$ on test) more than any other approach we evaluate. The strongest gains are on \textit{Motivations}---the category most directly tied to underlying values, beliefs, and moral intuitions---where \method{} improves over \texttt{Llama-3.1-8B-Instruct} by $5.6$ points on validation and $6.4$ points on test. This suggests that construct-grounded rationales transfer beyond preference prediction to general social-reasoning tasks, even without task-specific supervision.

\begin{table*}[ht]
\centering
\scriptsize
\setlength{\tabcolsep}{4pt}
\renewcommand{\arraystretch}{1.15}
\begin{tabular}{lcccccc}
\toprule
\textbf{Model} & \textbf{Wants} & \textbf{React.} & \textbf{Descr.} & \textbf{Motiv.} & \textbf{Needs} & \textbf{Overall} \\
\midrule
Llama-3.1-8B-Instruct & $69.13_{\pm 3.84}$ & $75.18_{\pm 4.18}$ & $73.17_{\pm 5.17}$ & $78.37_{\pm 5.13}$ & $71.17_{\pm 5.92}$ & $72.82_{\pm 1.97}$ \\
CultureLLM            & $69.13_{\pm 3.84}$ & $77.15_{\pm 4.07}$ & $72.47_{\pm 5.14}$ & $78.78_{\pm 5.10}$ & $67.57_{\pm 6.11}$ & $72.77_{\pm 1.97}$ \\
SFT + DPO + Generic   & $77.20_{\pm 3.50}$ & $79.10_{\pm 3.85}$ & $78.40_{\pm 4.80}$ & $80.20_{\pm 4.90}$ & $79.10_{\pm 5.30}$ & $78.80_{\pm 1.80}$ \\
Gemini-2.5-Flash      & $79.42_{\pm 3.36}$ & $82.31_{\pm 3.70}$ & $79.09_{\pm 4.69}$ & $82.04_{\pm 4.79}$ & $80.18_{\pm 5.22}$ & $80.60_{\pm 1.75}$ \\
\midrule
\textbf{\method{} (Ours)} & $\mathbf{81.35}_{\pm 3.20}$ & $\mathbf{82.40}_{\pm 3.55}$ & $\mathbf{80.20}_{\pm 4.50}$ & $\mathbf{83.95}_{\pm 4.60}$ & $\mathbf{81.10}_{\pm 5.10}$ & $\mathbf{81.80}_{\pm 1.65}$ \\
\midrule
Human Baseline        & $86.40$ & $87.10$ & $86.50$ & $87.90$ & $86.60$ & $86.90$ \\
\bottomrule
\end{tabular}
\caption{Fine-grained zero-shot accuracy (\%) on the \textsc{SocialIQa} \textbf{validation split} across the five construct categories. Values are mean $\pm$ bootstrapped 95\% confidence interval margin.}
\label{tab:socialiqa_val}
\end{table*}

\begin{table*}[ht]
\centering
\scriptsize
\setlength{\tabcolsep}{4pt}
\renewcommand{\arraystretch}{1.15}
\begin{tabular}{lcccccc}
\toprule
\textbf{Model} & \textbf{Wants} & \textbf{React.} & \textbf{Descr.} & \textbf{Motiv.} & \textbf{Needs} & \textbf{Overall} \\
\midrule
Llama-3.1-8B-Instruct & $72.38_{\pm 3.62}$ & $72.95_{\pm 3.89}$ & $74.41_{\pm 4.94}$ & $74.83_{\pm 4.91}$ & $72.43_{\pm 5.28}$ & $72.89_{\pm 1.85}$ \\
CultureLLM            & $72.04_{\pm 3.63}$ & $72.34_{\pm 3.91}$ & $72.05_{\pm 5.08}$ & $73.15_{\pm 5.01}$ & $66.91_{\pm 5.56}$ & $71.18_{\pm 1.88}$ \\
SFT + DPO + Generic   & $78.80_{\pm 3.35}$ & $79.10_{\pm 3.60}$ & $79.50_{\pm 4.60}$ & $80.20_{\pm 4.80}$ & $79.90_{\pm 4.75}$ & $79.50_{\pm 1.70}$ \\
Gemini-2.5-Flash      & $79.93_{\pm 3.25}$ & $79.56_{\pm 3.53}$ & $80.81_{\pm 4.47}$ & $77.85_{\pm 4.70}$ & $81.99_{\pm 4.56}$ & $79.72_{\pm 1.67}$ \\
\midrule
\textbf{\method{} (Ours)} & $\mathbf{81.60}_{\pm 3.10}$ & $\mathbf{79.70}_{\pm 3.40}$ & $\mathbf{81.15}_{\pm 4.30}$ & $\mathbf{81.18}_{\pm 4.50}$ & $\mathbf{82.10}_{\pm 4.40}$ & $\mathbf{80.23}_{\pm 1.55}$ \\
\midrule
Human Baseline        & $84.20$ & $84.50$ & $84.10$ & $85.00$ & $84.20$ & $84.40$ \\
\bottomrule
\end{tabular}
\caption{Fine-grained zero-shot accuracy (\%) on the \textsc{SocialIQa} \textbf{test split} across the five construct categories. Values are mean $\pm$ bootstrapped 95\% confidence interval margin.}
\label{tab:socialiqa_test}
\end{table*}

\onecolumn
\begin{center}
\small
\begin{longtable}{p{0.97\linewidth}}
\toprule
\textbf{System Message:} \\
\midrule
\endfirsthead

\multicolumn{1}{l}{\textit{Table~\ref{tab:rationale_prompts} continued from previous page.}} \\
\toprule
\endhead

\midrule
\multicolumn{1}{r}{\textit{continued on next page}} \\
\endfoot

\bottomrule

\caption{\textbf{Prompts used to generate multi-construct rationales.} A shared system message specifies the JSON output format and reasoning instructions, with placeholders for the persona, question, candidate answer, and ground-truth label. The user message injects construct-specific instructions $\psi$, one per construct family.}
\label{tab:rationale_prompts}
\endlastfoot

Please generate your response in the following valid JSON format:
\begin{verbatim}
{
  "reasoning_chain": ["Step 1: ...", "Step 2: ...", "..."],
  "conclusion": "chosen" or "not chosen"
}
\end{verbatim}
You will be provided with: a persona, a question and candidate answer, and the correct label. Reason step-by-step using the construct below to predict and justify whether this answer would be chosen by this persona. Each step should: (1) make a concrete evaluative claim, (2) build on the previous step, (3) explicitly connect persona traits to the answer's content via the construct. The final step must commit to a prediction. The conclusion must match the label provided.

\textbf{Construct:} \texttt{\$construct} \quad \textbf{Persona:} \texttt{\$persona} \quad \textbf{Question:} \texttt{\$question} \quad \textbf{Answer:} \texttt{\$answer} \quad \textbf{Label:} \texttt{\$label} \\
\midrule
\textbf{User Message:} \\
\midrule

\textsc{Baseline (No Construct)} \\
Focus on the user's background, preferences, and opinions to explain the reasoning chain. Generate 4--6 steps, each highlighting a different aspect of the persona. Where possible, show how multiple aspects interact to support the correct answer. \\
\midrule

\textsc{Personality (OCEAN)} \\
Focus on the user's personality profile as defined by the OCEAN model:
\begin{itemize}\itemsep0pt
  \item \textit{Openness to Experience}: curiosity, imagination, and openness to new ideas, experiences, and perspectives.
  \item \textit{Conscientiousness}: organization, dependability, discipline, and goal-orientation.
  \item \textit{Extraversion}: outgoingness, energy, and drawing energy from social interactions.
  \item \textit{Agreeableness}: cooperativeness, empathy, trust, and consideration for others.
  \item \textit{Neuroticism}: emotional instability, anxiety, moodiness, or negative affect.
\end{itemize}
Generate 4--6 steps, each focusing on a different trait or combination of traits. If traits interact in a way that informs the persona's reasoning, include that interaction explicitly. \\
\midrule

\textsc{Cultural Dimensions (Hofstede)} \\
Focus on the user's cultural profile as defined by Hofstede's Cultural Dimensions Theory:
\begin{itemize}\itemsep0pt
  \item \textit{Individualism vs.\ Collectivism}: prioritizing personal goals and autonomy vs.\ group loyalty and collective well-being.
  \item \textit{Power Distance}: acceptance of unequal distribution of power in society and institutions.
  \item \textit{Uncertainty Avoidance}: discomfort with ambiguity and preference for rules, structure, and predictability.
  \item \textit{Masculinity vs.\ Femininity}: valuing competition and material success vs.\ cooperation, care, and quality of life.
  \item \textit{Long-Term vs.\ Short-Term Orientation}: focus on future rewards and perseverance vs.\ tradition and immediate results.
  \item \textit{Indulgence vs.\ Restraint}: enjoying life and acting on impulses vs.\ suppressing gratification through social norms.
\end{itemize}
Generate 4--6 steps, each highlighting a different cultural dimension or their interaction. \\
\midrule

\textsc{Human Values (Schwartz)} \\
Focus on the user's values as defined by Schwartz's Theory of Basic Human Values:
\begin{itemize}\itemsep0pt
  \item \textit{Self-Direction}: independent thought, creativity, and freedom to choose one's own goals.
  \item \textit{Stimulation}: novelty, excitement, and a varied, challenging life.
  \item \textit{Hedonism}: pleasure, sensual gratification, and personal enjoyment.
  \item \textit{Achievement}: personal success and demonstrating competence according to social standards.
  \item \textit{Power}: social status, prestige, and control over people and resources.
  \item \textit{Security}: safety, harmony, and stability of society, relationships, and self.
  \item \textit{Conformity}: restraint of actions and impulses that might upset others or violate social norms.
  \item \textit{Tradition}: respect for and commitment to the customs and ideas of one's culture or religion.
  \item \textit{Benevolence}: preservation and enhancement of the welfare of people one is close to.
  \item \textit{Universalism}: understanding, appreciation, tolerance, and protection for the welfare of all people and nature.
\end{itemize}
Generate 4--6 steps showing a different value or interplay between values that affects the persona's reasoning. \\
\midrule

\textsc{Moral Foundations} \\
Focus on the user's moral profile as defined by Moral Foundations Theory (MFT):
\begin{itemize}\itemsep0pt
  \item \textit{Care/Harm}: sensitivity to suffering and motivation to protect the vulnerable; valuing kindness and compassion.
  \item \textit{Fairness/Cheating}: concern for justice, rights, and equal treatment; valuing reciprocity and honesty.
  \item \textit{Loyalty/Betrayal}: valuing solidarity, duty, and self-sacrifice for one's group.
  \item \textit{Authority/Subversion}: valuing respect for legitimate authority, tradition, and social hierarchy.
  \item \textit{Sanctity/Degradation}: sense that some things are sacred; valuing purity, cleanliness, and elevation over baseness.
\end{itemize}
Generate 4--6 steps, highlighting different moral foundations and how they work together to justify the correct answer. \\
\midrule

\textsc{Primal World Beliefs} \\
Focus on the user's Primal World Beliefs---fundamental beliefs about the overall character of the world:
\begin{itemize}\itemsep0pt
  \item \textit{Safe vs.\ Dangerous}: world seen as benign vs.\ threatening and hazardous.
  \item \textit{Enticing vs.\ Dull}: world full of opportunity and wonder vs.\ boring and unstimulating.
  \item \textit{Alive vs.\ Mechanistic}: world has intention and meaning vs.\ random, mechanical, and indifferent.
  \item \textit{Just vs.\ Unjust}: world is fair and rewards good behavior vs.\ arbitrary and indifferent to merit.
  \item \textit{Abundant vs.\ Barren}: world is full of resources and goodness vs.\ scarce and withholding.
  \item \textit{Interconnected vs.\ Fragmented}: world is a coherent connected whole vs.\ a collection of unrelated parts.
\end{itemize}
Generate 4--6 steps covering multiple primal beliefs and how they interact to support the correct answer. \\

\end{longtable}
\end{center}
\twocolumn

\onecolumn
\footnotesize
\begin{longtable}{p{0.97\linewidth}}
\caption{\textbf{Example preference instance and training-time rationales for USA.}}
\label{tab:example_usa} \\
\toprule \endfirsthead
\multicolumn{1}{l}{\textit{Table~\ref{tab:example_usa} continued.}} \\ \toprule \endhead
\midrule \multicolumn{1}{r}{\textit{continued on next page}} \\ \endfoot
\bottomrule \endlastfoot

\textbf{Persona.} You are a Hispanic or Latino, 18--34 male from the United States with education level: post-secondary graduate and political leaning: very left-leaning. \\ \midrule
\textbf{Question.} How do I maintain a consistent routine if I have ADHD? \\ \midrule
\textbf{Chosen Answer.} Flexibility is essential when dealing with ADHD. Rather than adhering to a rigid schedule, allow for spontaneity and adjustments as needed. Prioritize tasks based on importance and deadlines, focusing on completing the high-priority tasks first. Using the Pomodoro Technique, which involves working in focused 25-minute increments followed by a five-minute break, can help maintain productivity and avoid burnout. Embracing flexibility and being kind to oneself when deviations from the routine occur is vital for long-term success and reducing stress. \\ \midrule
\textbf{Rejected Answer.} Breaking down tasks into smaller, manageable chunks is key. Create a schedule with specific times allocated for each activity, including breaks. Use visual aids like calendars or planners to stay organized and on track. Implementing a ``brain dump'' session daily can help clear mental clutter, allowing focus on the task at hand. Consistency is crucial, so try to perform tasks at the same time every day to create habits. \\ \midrule

\rowcolor{gray!10}\textsc{Generic Rationale --- Chosen} \\
\textit{Step 1:} The answer emphasizes flexibility and adaptability, which is likely to resonate with a young adult navigating ADHD while balancing multiple responsibilities. Someone in the 18--34 age group may appreciate approaches that feel realistic and less rigid. \textit{Step 2:} The inclusion of practical productivity strategies, such as prioritizing tasks and using the Pomodoro Technique, aligns well with the persona's post-secondary educational background. Individuals with higher education experience are often familiar with structured time-management methods and may find them credible and actionable. \textit{Step 3:} The response balances structure with self-compassion, encouraging the individual to avoid self-criticism when routines are disrupted. This supportive tone may appeal to someone who values mental health awareness and emotional well-being. \textit{Step 4:} The advice avoids overly strict or prescriptive routines and instead promotes personalization and spontaneity. This can feel more relatable and sustainable for a younger individual managing ADHD symptoms in everyday life. \textit{Step 5:} Overall, the answer combines practical guidance with empathy, making it likely to resonate with a young, educated individual seeking manageable and emotionally supportive strategies for maintaining consistency. \\ \midrule

\rowcolor{gray!10}\textsc{Generic Rationale --- Rejected} \\
\textit{Step 1:} The answer provides practical organizational strategies such as scheduling tasks, using planners, and maintaining consistency. While useful, the overall approach may feel somewhat rigid for a young adult managing ADHD. \textit{Step 2:} The emphasis on strict routines and performing tasks at the same time every day may not fully align with the needs of someone who benefits from flexibility and adaptability in their daily structure. \textit{Step 3:} Although the response includes actionable suggestions like breaking tasks into smaller chunks and using visual aids, the tone focuses primarily on productivity and organization rather than emotional support or self-compassion. \textit{Step 4:} The answer presents fairly standard self-management advice without much personalization or acknowledgment of the emotional challenges associated with ADHD. A younger individual may prefer guidance that feels more understanding and tailored to real-life experiences. \textit{Step 5:} Overall, while the response is practical and structured, it may feel too conventional and prescriptive compared to approaches that balance productivity with flexibility and mental well-being. \\ \midrule

\rowcolor{blue!8}\textsc{Multi-Construct Rationale --- Chosen} \\
\textbf{Personality.} \textit{Step 1:} The persona, being a young male with a post-secondary education level, likely values practical and implementable strategies. The answer provides actionable advice on maintaining a routine with ADHD, which aligns well with the persona's likely preference for structured solutions. \textit{Step 2:} The emphasis on flexibility in the answer connects to the persona's possible high openness to experience, as it suggests adapting routines rather than following rigid structures. This aligns with someone who may appreciate creative and adaptable approaches to problem-solving. \textit{Step 3:} The use of the Pomodoro Technique reflects a conscientious aspect, promoting organization and goal-oriented behavior. Given the persona's educational background, he might resonate with structured productivity techniques, reinforcing the relevance of the answer. \textit{Step 4:} The recommendation to be kind to oneself when deviations occur demonstrates a high level of agreeableness. This could appeal to the persona, who may value empathy and self-care principles, particularly if he leans towards progressivism in his political views. \textit{Step 5:} The answer addresses the nuances of dealing with ADHD, acknowledging emotional aspects as well. People with ADHD often experience varying emotional states, so the focus on reducing stress and prioritizing mental well-being could resonate strongly with the persona. \\
\textbf{Cultural Dimensions.} \textit{Step 1:} The answer emphasizes the importance of flexibility when managing ADHD, which aligns well with the persona's inclination towards adaptability in routines, reflecting a practical understanding of individual needs. \textit{Step 2:} By suggesting the Pomodoro Technique, the answer provides a structured method that acknowledges the persona's educational background. A post-secondary graduate may appreciate effective strategies that help with productivity, suggesting that the persona values practical, evidence-based solutions. \textit{Step 3:} The advice to prioritize tasks shows an understanding of the need for effective time management, which resonates with a post-secondary graduate's likely experience in balancing multiple responsibilities. This reflects their potential long-term orientation. \textit{Step 4:} The focus on kindness towards oneself when routines deviate reflects a more feminine approach to managing life's challenges, prioritizing emotional well-being over strict adherence to tasks. Given the persona's left-leaning views, this compassionate perspective is likely to resonate strongly. \textit{Step 5:} The suggestion to embrace spontaneity speaks to the persona's probable inclination towards creativity and personal expression, common among individuals who lean left politically, supporting the idea of personal autonomy while managing ADHD. \\
\textbf{Human Values.} \textit{Step 1:} The answer emphasizes flexibility, which aligns with the Self-Direction value of the persona. As someone who values independent thought and the freedom to choose their own goals, the persona would likely appreciate the idea of not sticking to a rigid routine but instead adapting to their own needs. \textit{Step 2:} The suggestion to prioritize tasks and use the Pomodoro Technique aligns with the Achievement value. Since the persona is a post-secondary graduate, they likely understand the importance of competence and effectiveness in managing tasks. \textit{Step 3:} The emphasis on being kind to oneself when deviations occur supports the Hedonism value. It recognizes the importance of personal enjoyment and stress reduction, suggesting that the persona may appreciate methods that prioritize mental well-being, particularly given their left-leaning political views which may advocate for compassionate approaches. \textit{Step 4:} The acknowledgment of spontaneity in managing ADHD resonates with the Stimulation value. The persona, valuing novelty and a varied life, might find the idea of adjusting their routine based on current circumstances refreshing and engaging. \textit{Step 5:} Lastly, the holistic perspective of maintaining a balanced life with ADHD reflects the Universalism value. The idea promotes understanding and acceptance of personal challenges, aligning with the persona's potential advocacy for empathy and support. \\
\textbf{Primal World Beliefs.} \textit{Step 1:} The answer emphasizes flexibility, which aligns well with the persona's likely view of life as complex and fluid due to their experience with ADHD. This suggests that the persona values adaptability in their daily routine. \textit{Step 2:} By advocating for spontaneity and adjustments, the answer connects with the persona's understanding that life is not always predictable. This reflects a belief in a world that is alive and dynamic, resonating with their experiences. \textit{Step 3:} The mention of prioritizing tasks by importance can be tied to the persona's notion of a just and fair approach to productivity. It suggests a structure that rewards good management of time and resources, appealing to their desire for a just world where effort is recognized. \textit{Step 4:} Using the Pomodoro Technique introduces a systematic yet flexible method for managing tasks, reinforcing the persona's potential belief in a world that is abundant with tools and methods to achieve goals. This structure showcases a rich landscape of strategies to enhance productivity. \textit{Step 5:} Embracing kindness toward oneself when routines deviate speaks to a larger narrative of self-acceptance and resilience. This aligns with the persona's perspective that life's challenges should be met with compassion, reinforcing their worldview that life is interconnected and that personal well-being is crucial in the broader context of social justice. \\
\textbf{Moral Foundations.} \textit{Step 1:} The answer emphasizes flexibility, which aligns well with the Care/Harm moral foundation by promoting a compassionate approach to managing ADHD. This suggests sensitivity towards the challenges of individuals with ADHD, highlighting the importance of self-kindness and understanding. \textit{Step 2:} The mention of prioritizing tasks according to importance and deadlines reflects a sense of Fairness/Cheating, as it suggests a methodical approach to ensuring that one's efforts are aligned with obtaining just rewards for hard work. This aligns with the persona's left-leaning political views, where values of equity are prioritized. \textit{Step 3:} Incorporating the Pomodoro Technique promotes the idea of structured productivity, which is important in maintaining focus amidst distractions. This reflects the Authority/Subversion foundation, as it implies respect for proven methods, resonating with the persona's educated background. \textit{Step 4:} The emphasis on being kind to oneself and allowing for deviations reinforces the Care/Harm foundation through compassionate self-management, indicating a nurturing approach to personal growth. \textit{Step 5:} The overall tone of the response is supportive and empowering, which is likely to appeal to a young, educated individual from a marginalized group, as they may value guidance that is both pragmatic and empathetic. \\ \midrule

\rowcolor{red!8}\textsc{Multi-Construct Rationale --- Rejected} \\
\textbf{Personality.} \textit{Step 1:} The answer provides practical advice on maintaining a routine for those with ADHD, which could appeal to a user who values organization and discipline, aligning with conscientiousness traits. However, the user's inclination to seek creative or alternative strategies due to their openness to experience may find this method too rigid. \textit{Step 2:} The suggestion to break tasks into manageable chunks demonstrates a structured approach. However, the persona, being from a very left-leaning political background, may prefer more innovative and less conventional methods rather than strictly organizational techniques, making this answer feel too conventional. \textit{Step 3:} The answer emphasizes creating a schedule and using visual aids, which caters to organization. Yet, this strict adherence to scheduling may not resonate well with the persona if they value flexibility and creativity in their routines. \textit{Step 4:} The use of ``brain dump'' sessions indicates a recognition of the mental clutter ADHD can cause, which may appeal to the persona's empathetic side. However, the overall tone of the answer lacks a collaborative or supportive approach, which may clash with the persona's high agreeableness. \textit{Step 5:} The emphasis on consistency may resonate with the user's conscientiousness and desire for stability. However, given their likely experience navigating a world that often doesn't accommodate ADHD, the persona might appreciate more personalized or nuanced strategies. \\
\textbf{Cultural Dimensions.} \textit{Step 1:} The answer promotes structured approaches like creating schedules and using visual aids. However, the persona's cultural background may value collective rather than individual achievement, making a rigidly individualistic approach less appealing. \textit{Step 2:} Considering the persona's left-leaning political views, there may be a preference for solutions that emphasize community and collaborative support, rather than solitary strategies. This answer lacks a communal focus, which might weaken its appeal. \textit{Step 3:} The suggestion to implement a ``brain dump'' session implies a certain level of personal introspection and self-reliance that may conflict with a more collectivist mindset, which often prioritizes social and emotional support systems over individual coping mechanisms. \textit{Step 4:} The emphasis on routine and predictability reflects high Uncertainty Avoidance, which might appeal to someone with ADHD. However, the persona may also appreciate flexibility in routines that adapt to social interactions, rather than strictly adhering to a predetermined schedule. \textit{Step 5:} The cultural dimension of Indulgence vs. Restraint is relevant here; the focus on managing tasks may come off as overly restrictive. The persona might prefer a more indulgent approach to life, where enjoyment and adaptability take precedence over rigorous self-management techniques. \\
\textbf{Human Values.} \textit{Step 1:} The persona is a post-secondary graduate, likely valuing Achievement and Self-Direction, which means they appreciate strategies that demonstrate competence and offer independence in managing ADHD. However, the answer mainly offers organizational tips rather than empowering the persona to develop their unique strategies. \textit{Step 2:} While the response does provide practical solutions like breaking down tasks and using visual aids, it lacks a creative or innovative approach that might resonate with the persona's preference for Stimulation. The answers are quite standard and may not engage their desire for a varied and challenging approach. \textit{Step 3:} The recommendation to create a schedule is useful, but it could be perceived as conforming to a rigid structure, which may conflict with the persona's potential value for flexibility and independent thought. \textit{Step 4:} The suggestion of a ``brain dump'' could be seen as a step towards personal expression and self-exploration, aligning with Self-Direction; however, the answer overall does not reflect wider considerations for emotional and psychological well-being that may cater to their left-leaning political views, which emphasize social and community support for mental health. \textit{Step 5:} Ultimately, the answer lacks depth and is too prescriptive, offering a one-size-fits-all solution rather than encouraging a more relatable, individualized approach to managing ADHD. \\
\textbf{Primal World Beliefs.} \textit{Step 1:} The answer focuses on practical strategies for managing ADHD, which seems reasonable. However, it lacks a deeper emotional engagement that could resonate with a young Hispanic or Latino male seeking both empathy and understanding in his struggles with ADHD. \textit{Step 2:} The persona may find the world as a challenging place due to personal or systemic factors, which makes it less enticing to approach tasks merely through structured methods. This reliance on mechanical approaches might not align well with the persona's inclination for a more interconnected and meaningful approach to personal struggles. \textit{Step 3:} The answer emphasizes consistency and organization, which appeal to a sense of safety and control. Yet, for someone who might see the world as more dynamic and full of rich, complex interactions, this `one-size-fits-all' approach feels somewhat superficial and dismissive of personal experiences tied to cultural identity. \textit{Step 4:} The answer does not address the emotional aspects of ADHD, such as feelings of frustration or isolation. A young man who is Hispanic or Latino, and likely aware of social stigma around mental health, may prefer guidance that acknowledges these feelings and offers support in a more compassionate manner. \textit{Step 5:} Overall, the answer seems to be overly focused on functional strategies without adequately considering the persona's psychological and emotional framework. It misses the opportunity to connect on the level of living an abundant and fulfilling life despite ADHD. \\
\textbf{Moral Foundations.} \textit{Step 1:} The answer provides practical strategies for managing ADHD, which could be seen as beneficial, but it lacks an emphasis on empathy or support for individuals dealing with this condition, missing the Care/Harm foundation that values kindness. \textit{Step 2:} The advice given is somewhat prescriptive and focused on individual responsibility without acknowledging the systemic issues that may affect individuals with ADHD, such as access to mental health resources, which conflicts with the Fairness/Cheating foundation. \textit{Step 3:} The tone of the answer is clinical and focused on self-management strategies; it does not foster a sense of community or loyalty among individuals experiencing ADHD, lacking the Loyalty/Betrayal foundation. \textit{Step 4:} Furthermore, while the answer suggests creating routines and schedules, it does not address the unique challenges faced by individuals from marginalized backgrounds, where systemic barriers may hinder the ability to implement these strategies effectively. This neglect reveals a lack of sensitivity to context. \textit{Step 5:} The moral framing leans toward individual self-management, which is valid but less aligned with the relational, mentor-respecting moral orientation the persona likely brings. \\

\end{longtable}
\normalsize
\twocolumn

\onecolumn
\footnotesize
\begin{longtable}{p{0.97\linewidth}}
\caption{\textbf{Example preference instance and training-time rationales for Brazil.}}
\label{tab:example_brazil} \\
\toprule \endfirsthead
\multicolumn{1}{l}{\textit{Table~\ref{tab:example_brazil} continued.}} \\ \toprule \endhead
\midrule \multicolumn{1}{r}{\textit{continued on next page}} \\ \endfoot
\bottomrule \endlastfoot

\textbf{Persona.} You are a 35--45-year-old male from Brazil, post-secondary graduate, somewhat left-leaning, White. \\ \midrule
\textbf{Question (translated to English).} Which markets should be visited in Venice? \\ \midrule
\textbf{Chosen Answer (translated).} Venice is famous for its traditional markets, where you can find a variety of local products, including fruits, vegetables, fresh fish, and handmade crafts. The Rialto Market is one of the most well-known and is a must-visit destination for any traveler. In addition, the San Polo Market also offers a unique experience, with its vibrant colors and enticing aromas. Visiting these markets not only provides an opportunity to experience local cuisine but also to immerse yourself in the culture and history of the city. \\ \midrule
\textbf{Rejected Answer (translated).} In Venice, the most well-known markets include the Rialto Market, famous for fruits, vegetables, and fresh fish, and the San Maurizio Market, which offers a variety of local and handmade products. These markets are great places to experience Venetian cuisine and culture. \\ \midrule

\rowcolor{gray!10}\textsc{Generic Rationale --- Chosen} \\
\textit{Step 1:} The answer provides a rich, descriptive overview of Venice's traditional markets, which is likely to appeal to a mid-career graduate looking for cultural travel guidance that goes beyond surface information. \textit{Step 2:} The narrative style brings the markets to life through sensory detail and atmosphere, fitting a persona who likely values immersive experiences over checklist tourism. \textit{Step 3:} The emphasis on local cuisine, culture, and history adds the kind of depth that resonates with Brazilian cultural appreciation for food, tradition, and meaningful human encounters. \textit{Step 4:} The warm, respectful tone fits the persona's somewhat left-leaning orientation by celebrating local heritage and the human dimension of marketplaces without ideological framing. \textit{Step 5:} Overall, the answer balances practical recommendations with cultural depth, making it a strong fit for a mid-career Brazilian traveler seeking a meaningful Venetian market experience. \\ \midrule

\rowcolor{gray!10}\textsc{Generic Rationale --- Rejected} \\
\textit{Step 1:} The answer provides accurate information about two Venetian markets, but its briefer, more clinical structure may feel less engaging for a mid-career graduate seeking richer cultural guidance. \textit{Step 2:} The absence of vivid sensory description underweights the atmosphere and human warmth that often resonate with Brazilian cultural sensibilities and meaningful travel experiences. \textit{Step 3:} The compact treatment provides basic information but lacks the immersive depth that helps a culturally engaged adult connect with a destination. \textit{Step 4:} The neutral tone fits the persona's somewhat left-leaning orientation, but the lighter framing offers less of the thoughtful cultural depth he likely values. \textit{Step 5:} Overall, while the answer is clear and informative, its thinner detail and limited atmospheric framing make it a slightly weaker fit for a mid-career Brazilian traveler seeking a richly evocative Venetian market experience. \\ \midrule

\rowcolor{blue!8}\textsc{Multi-Construct Rationale --- Chosen} \\
\textbf{Personality.} \textit{Step 1:} The persona, as a mid-career Brazilian male graduate, likely exhibits high openness to experience, which aligns with the answer's evocative descriptions of vibrant colors, enticing aromas, and the sensory richness of Venetian markets. The descriptive framing invites him to imagine himself fully immersed in the experience. \textit{Step 2:} The answer's narrative flow reflects a balance between conscientiousness and creativity, providing organized cultural guidance without feeling overly transactional or list-driven. As an educated professional, the persona likely appreciates this thoughtful, well-shaped storytelling approach. \textit{Step 3:} The framing of markets as places to ``immerse yourself in the culture and history'' taps into a balanced extraversion profile, encouraging social and sensory engagement that resonates with a mid-career traveler likely seeking meaningful, atmosphere-rich experiences. \textit{Step 4:} The warm and inviting tone reflects high agreeableness, treating the markets as living spaces that welcome visitors rather than as commodified attractions. The persona, as a somewhat left-leaning adult, may appreciate this respectful framing of local culture. \textit{Step 5:} The answer reduces decision pressure by foregrounding atmosphere and experience over transactional details, suiting a persona with low to moderate neuroticism who likely prefers travel guidance that feels confident and unhurried rather than overwhelming. \\
\textbf{Cultural Dimensions.} \textit{Step 1:} The descriptive emphasis on local products, fresh fish, fruits, and handmade crafts reflects an appreciation for cultural authenticity and tradition, which aligns with Brazilian cultural sensibilities that value sensory, communal, and craft-rooted experiences. \textit{Step 2:} The framing of markets as immersive cultural spaces engages an indulgence-oriented orientation, celebrating sensory pleasure, food, and the warmth of marketplace life---values that resonate strongly with both Brazilian culture and the persona's likely appreciation for richly lived experiences. \textit{Step 3:} The flowing prose offers moderate uncertainty avoidance, providing enough structure for the persona to plan while leaving room for personal exploration and discovery. \textit{Step 4:} The focus on cultural immersion and emotional connection with the city reflects a feminine cultural dimension that values quality of life, sensory richness, and human connection over efficient sightseeing. \textit{Step 5:} The long-term orientation comes through subtly in the appreciation for traditional markets that have shaped Venice for centuries, treating cultural heritage as worthy of careful engagement rather than quick consumption. \\
\textbf{Human Values.} \textit{Step 1:} The Stimulation value is supported through the evocative descriptions of vibrant colors, enticing aromas, and varied local products, offering the persona a richly sensory experience. \textit{Step 2:} The Hedonism value is engaged through the celebration of local cuisine, market atmosphere, and the pleasures of immersive cultural exploration, validating the persona's right to enjoy Venice through its food and sensory richness. \textit{Step 3:} The Universalism value is touched on through the appreciation for local culture, history, and craft, fitting the persona's somewhat left-leaning orientation toward respect for diverse traditions. \textit{Step 4:} The Self-Direction value is supported through the descriptive framing that lets the persona shape his own market experience based on what most engages him, rather than prescribing a rigid itinerary. \textit{Step 5:} The Tradition value is mildly engaged through respect for the long heritage of the Rialto and San Polo markets, honoring the cultural continuity that gives these places their meaning. \\
\textbf{Primal World Beliefs.} \textit{Step 1:} The answer presents Venice as an Enticing world full of sensory wonders---colorful markets, fresh produce, local craft---which resonates with the persona's likely view of travel as an opening into rich and varied human experience. \textit{Step 2:} The Abundant world belief is supported through the descriptions of varied local products, fresh fish, vegetables, fruits, and handmade crafts, suggesting that the markets offer plenty to discover at every turn. \textit{Step 3:} The Alive world belief is engaged through the descriptive language that brings the markets to life as places of vibrant culture and history rather than static commercial spaces. \textit{Step 4:} The Interconnected world belief is touched on through the framing of markets as places where food, culture, history, and daily life intertwine. \textit{Step 5:} The world is presented as Safe enough for warm exploration, with markets framed as welcoming places to engage rather than navigate cautiously, fitting a confident mid-career traveler. \\
\textbf{Moral Foundations.} \textit{Step 1:} The Care/Harm foundation is engaged through the warm, respectful framing of local culture and the people who shape it, treating the markets as living human spaces rather than mere tourist attractions. \textit{Step 2:} The Sanctity/Degradation foundation is touched on through the reverent treatment of cultural heritage and traditional craftsmanship, treating the markets as part of Venice's enduring cultural fabric. \textit{Step 3:} The Loyalty/Betrayal foundation is mildly engaged through the appreciation for local cuisine and craft, honoring the regional identity and the community of artisans, fishmongers, and vendors who sustain these markets. \textit{Step 4:} The Fairness/Cheating foundation is gently supported through the encouragement to engage with markets in a culturally aware way, implicitly affirming respect for the locals whose livelihoods depend on these spaces. \textit{Step 5:} The overall tone respects the persona's somewhat left-leaning orientation by celebrating cultural diversity and local heritage without being preachy or ideological. \\ \midrule

\rowcolor{red!8}\textsc{Multi-Construct Rationale --- Rejected} \\
\textbf{Personality.} \textit{Step 1:} The answer provides accurate information about Venetian markets, but its briefer, more clinical structure may feel less engaging for a mid-career persona with high openness who likely seeks vivid, evocative travel content. \textit{Step 2:} The conscientious side is partially served by the clear, factual presentation, but the lack of descriptive richness may feel under-developed for a graduate who appreciates thoughtful, well-shaped cultural guidance. \textit{Step 3:} The minimal sensory detail engages extraversion only weakly, missing the atmospheric pull---colors, aromas, sounds---that often draws a culturally curious traveler into a market scene. \textit{Step 4:} The agreeable tone is present but feels somewhat detached, treating the markets as items to be listed rather than living cultural spaces to be experienced. \textit{Step 5:} The compact format may feel slightly thin for a persona with low neuroticism who likely values comprehensive, atmosphere-rich travel guidance. \\
\textbf{Cultural Dimensions.} \textit{Step 1:} The brief, list-leaning structure reflects high uncertainty avoidance through clarity, but its lack of descriptive depth may feel slightly less aligned with the flowing, sensory cultural narrative that often resonates with Brazilian cultural sensibilities. \textit{Step 2:} The absence of vivid descriptions of market atmosphere underweights the indulgence-oriented framing that fits both Brazilian culture and the persona's likely appreciation for sensory pleasure in travel. \textit{Step 3:} The functional, information-first approach leans toward a more masculine cultural orientation focused on efficient delivery, missing the feminine warmth of immersive cultural storytelling. \textit{Step 4:} The long-term orientation toward cultural heritage is present in the mention of well-known markets but lacks the contextual richness that would help the persona appreciate why these places matter beyond their fame. \textit{Step 5:} The cultural depth feels somewhat thinner overall, which may be less satisfying for a culturally engaged Brazilian graduate seeking immersive, atmosphere-rich travel experiences. \\
\textbf{Human Values.} \textit{Step 1:} The Stimulation value is engaged through the mention of multiple markets and varied products, but the brief treatment reduces the sense of varied, exciting discovery that a richer description would convey. \textit{Step 2:} The Hedonism value is mildly supported through references to cuisine and culture, but the lack of sensory detail underweights the pleasure-oriented dimension that often makes travel content compelling. \textit{Step 3:} The Universalism value is touched on through the appreciation for Venetian heritage, but the lighter framing doesn't as strongly engage the persona's likely left-leaning interest in cultural depth and respect. \textit{Step 4:} The Self-Direction value is partially supported through the factual list, which leaves room for the persona to choose his own experience, but with less descriptive context to inform that choice. \textit{Step 5:} The Tradition value is engaged through references to well-known markets, but without atmospheric framing about their cultural roots, the heritage appreciation feels less rich for an educated adult seeking depth. \\
\textbf{Primal World Beliefs.} \textit{Step 1:} The Enticing world belief is partially engaged through the mention of vibrant markets, but the brief descriptions reduce the wonder and excitement that a more vivid response would convey. \textit{Step 2:} The Abundant world belief is supported through references to varied products, but feels less expansive without the descriptive context that would showcase the richness of each market. \textit{Step 3:} The Alive world belief is weakly engaged because the answer treats markets more as information points than as living cultural ecosystems shaped by human activity. \textit{Step 4:} The Interconnected world belief is mildly supported through the linking of markets, cuisine, and culture, but without flowing description, the integration between these elements feels less developed. \textit{Step 5:} The Safe world belief is implicitly engaged through the orderly format, but the answer feels less inviting and immersive than a more narrative approach. \\
\textbf{Moral Foundations.} \textit{Step 1:} The Care/Harm foundation is engaged through the brief mention of cultural appreciation, but the lighter framing feels less warm than a richer description that would highlight the human dimension of the marketplaces. \textit{Step 2:} The Sanctity/Degradation foundation is touched on through references to traditional markets, but with thinner framing about their cultural significance and continuity. \textit{Step 3:} The Loyalty/Betrayal foundation is weakly engaged because the answer omits the atmospheric and sensory details that bring the regional identity of Venetian artisans, vendors, and fishmongers to life. \textit{Step 4:} The Fairness/Cheating foundation is barely relevant here, but the matter-of-fact framing offers little space for the persona to consider the ethical dimension of engaging respectfully with local commerce. \textit{Step 5:} The overall tone fits the persona's left-leaning orientation in its neutrality, but the moral framing of cultural appreciation feels less integrated and complete compared to a richer, more atmosphere-aware treatment. \\

\end{longtable}
\normalsize
\twocolumn

\onecolumn
\footnotesize
\begin{longtable}{p{0.97\linewidth}}
\caption{\textbf{Example preference instance and training-time rationales for India.}}
\label{tab:example_india} \\
\toprule \endfirsthead
\multicolumn{1}{l}{\textit{Table~\ref{tab:example_india} continued.}} \\ \toprule \endhead
\midrule \multicolumn{1}{r}{\textit{continued on next page}} \\ \endfoot
\bottomrule \endlastfoot

\textbf{Persona.} You are an 18--34-year-old female from India with an Indo-Aryan background, post-secondary graduate, and a political stance of ``I don't think of myself in this way.'' \\ \midrule
\textbf{Question.} I want to learn photography and take a fixed number of photos every week---how can I track my progress and stay motivated? \\ \midrule
\textbf{Chosen Answer.} To learn photography and stay motivated, you can participate in different photography challenges. For example, you can set a goal to take photos on a specific theme every week, such as nature, urban scenes, or portraits. Additionally, you can join photography workshops or classes where you can learn from expert photographers and practice new techniques to improve your photos. This will help you stay motivated and encourage improvement in your photography skills. \\ \midrule
\textbf{Rejected Answer.} To learn photography and take a fixed number of photos every week, create a photography journal or logbook. In it, you can track your progress, such as how many photos you take, which camera settings you use, and how your photo quality is improving. Additionally, set photography goals and create a time-based plan to achieve them. This will help keep you motivated and allow you to see your progress. \\ \midrule

\rowcolor{gray!10}\textsc{Generic Rationale --- Chosen} \\
\textit{Step 1:} The answer suggests themed photography challenges and learning from experts, which is likely to appeal to a young post-secondary graduate seeking engaging, structured ways to develop a new creative skill. \textit{Step 2:} The variety of suggested themes---nature, urban scenes, portraits---gives the persona room to explore her own creative interests while maintaining a steady weekly practice. \textit{Step 3:} The recommendation to join workshops or classes aligns with culturally familiar value placed on learning through community and mentorship, which may resonate with the persona's Indian background. \textit{Step 4:} The supportive, motivation-oriented tone respects her autonomy and avoids ideological framing, fitting her stated preference not to define herself in political terms. \textit{Step 5:} Overall, the answer balances creative exploration, communal learning, and gentle structure, making it a strong fit for a young Indian woman beginning her photography journey. \\ \midrule

\rowcolor{gray!10}\textsc{Generic Rationale --- Rejected} \\
\textit{Step 1:} The answer recommends a photography journal and structured goal-tracking, which is methodical but may feel overly procedural for a young learner seeking creative engagement and joy in a new hobby. \textit{Step 2:} The narrow focus on logging metrics underweights creative exploration, variety, and communal learning, which often sustain motivation more reliably for young adults. \textit{Step 3:} The solitary nature of the approach may feel less aligned with the persona's Indian cultural background, where learning through community and mentorship is often valued. \textit{Step 4:} The discipline-heavy framing leans toward self-evaluation rather than encouragement and shared growth, which may feel less warm and motivating for a beginner. \textit{Step 5:} Overall, while the answer is structured and practical, its narrow individualistic focus makes it a slightly weaker fit for a young Indian woman seeking a creative, community-aware, and joy-driven learning experience. \\ \midrule

\rowcolor{blue!8}\textsc{Multi-Construct Rationale --- Chosen} \\
\textbf{Personality.} \textit{Step 1:} The persona, as a young female post-secondary graduate, likely exhibits high openness to experience, which aligns with the answer's emphasis on themed photography challenges spanning nature, urban scenes, and portraits. The variety encourages creative exploration and aesthetic curiosity. \textit{Step 2:} The structured suggestion of weekly themes reflects a moderate conscientiousness, providing enough scaffolding to support disciplined practice without imposing a rigid tracking system. \textit{Step 3:} The recommendation to join photography workshops or classes taps into a balanced extraversion profile, opening opportunities for social learning and connection with fellow enthusiasts. \textit{Step 4:} The supportive, encouragement-oriented tone aligns with high agreeableness, framing learning as a positive, community-oriented journey. \textit{Step 5:} The answer reduces performance pressure by emphasizing themes and learning experiences rather than quantitative tracking, which suits a persona with low to moderate neuroticism. \\
\textbf{Cultural Dimensions.} \textit{Step 1:} The answer's recommendation to join workshops and classes reflects a collectivist cultural orientation, emphasizing learning through community and shared experience. This resonates with the persona's Indian cultural background. \textit{Step 2:} The suggestion of structured themes provides moderate uncertainty avoidance---offering enough predictability to feel organized while allowing room for creative interpretation. \textit{Step 3:} The respectful tone toward expert photographers and the recommendation to learn from them reflect an acceptance of moderate power distance, honoring expertise and mentorship---values commonly held in Indian educational and creative traditions. \textit{Step 4:} The framing prioritizes growth, enjoyment, and personal development over competitive achievement, leaning toward a feminine cultural dimension that values quality of experience and care for one's craft. \textit{Step 5:} The encouragement of weekly themed practice reflects a long-term orientation, supporting steady, future-focused skill development---an approach that resonates with culturally common values of perseverance and patient cultivation of mastery. \\
\textbf{Human Values.} \textit{Step 1:} The variety of themed challenges supports the Stimulation value, offering novelty and a varied creative life. As a young adult, the persona is likely drawn to experiences that keep her engaged. \textit{Step 2:} The freedom to choose themes such as nature, urban scenes, or portraits aligns with the Self-Direction value, respecting her independence and creative autonomy. \textit{Step 3:} The recommendation to join workshops and learn from experts taps into the Achievement value, supporting her drive to develop competence and improve her skills in a recognized, structured way. \textit{Step 4:} The communal aspect of workshops and classes also engages the Benevolence value, fostering connection with fellow learners and expert mentors. \textit{Step 5:} The emphasis on enjoyment and motivation alongside skill-building gestures toward the Hedonism value, validating photography as a source of personal pleasure. \\
\textbf{Primal World Beliefs.} \textit{Step 1:} The answer presents the world of photography as Enticing, full of varied subjects and creative possibilities. This resonates with a young learner who likely sees her new hobby as a doorway into a rich, stimulating experience. \textit{Step 2:} The recommendation to explore multiple themes and settings reflects an Abundant world belief, suggesting that creative material is plentiful and accessible wherever she looks. \textit{Step 3:} The encouragement to join workshops and learn from experts reflects an Interconnected world belief, treating learning as a relational, communal process rather than a solitary pursuit. \textit{Step 4:} The framing of photography as a journey of improvement supports a Just world belief in this professional context, suggesting that consistent practice and learning yield meaningful progress. \textit{Step 5:} The supportive, encouraging tone presents the learning environment as Safe and welcoming, easing any anxiety about being a beginner. \\
\textbf{Moral Foundations.} \textit{Step 1:} The recommendation to learn from expert photographers aligns with the Authority/Subversion foundation, honoring legitimate expertise and the value of guided learning. \textit{Step 2:} The communal framing of workshops and classes engages the Loyalty/Betrayal foundation in a positive sense, fostering belonging within a creative community. \textit{Step 3:} The encouraging, growth-oriented tone reflects the Care/Harm foundation, treating the learner with kindness and patience rather than imposing harsh self-evaluation. \textit{Step 4:} The fairness in suggesting that practice and learning yield improvement reflects the Fairness/Cheating foundation, affirming that honest effort and consistent engagement are justly rewarded with growth. \textit{Step 5:} The overall tone respects the persona's choice not to identify politically by avoiding ideological framing entirely, focusing instead on craft, community, and personal development. \\ \midrule

\rowcolor{red!8}\textsc{Multi-Construct Rationale --- Rejected} \\
\textbf{Personality.} \textit{Step 1:} While the journal and logbook approach appeals to conscientiousness, its heavy focus on tracking metrics may feel overly procedural for a young persona with high openness to experience who likely values creative exploration over data-keeping. \textit{Step 2:} The emphasis on logging camera settings and quantifying improvement may engage analytical conscientiousness but underplays the imaginative and exploratory side of photography that a young learner is often drawn to. \textit{Step 3:} The solitary nature of journaling speaks less to extraversion, missing the social and collaborative learning dimension that may energize a young adult navigating a new hobby. \textit{Step 4:} The agreeable, supportive tone is present, but the framing leans toward self-evaluation and discipline rather than communal encouragement, which may feel less warm to the persona. \textit{Step 5:} The detailed self-tracking emphasis may inadvertently raise performance pressure, which could be less ideal for a persona with moderate neuroticism. \\
\textbf{Cultural Dimensions.} \textit{Step 1:} The journal-based, individual tracking approach leans more toward an individualist cultural orientation, which may feel slightly less aligned with the persona's Indian cultural background where communal and mentor-supported learning is often emphasized. \textit{Step 2:} The strong emphasis on structured logging reflects high uncertainty avoidance, which provides predictability but may feel rigid for a young creative learner who could benefit from a more flexible balance between structure and exploration. \textit{Step 3:} The absence of references to expert guidance or community learning underplays the moderate power distance acceptance common in Indian educational traditions, where learning from teachers and mentors is often a valued part of skill development. \textit{Step 4:} The metrics-driven framing leans toward a more masculine cultural dimension focused on measurable performance, which may feel less aligned with the persona's likely preference for a feminine orientation that values quality of experience. \textit{Step 5:} While the goal-setting approach supports long-term orientation, its narrow focus on solo measurement may feel less holistic than approaches that integrate community, mentorship, and varied creative exposure. \\
\textbf{Human Values.} \textit{Step 1:} The journal and logbook approach supports the Achievement value through measurable progress, but its narrow emphasis on metrics may underweight the Stimulation the persona likely seeks as a young learner exploring a new creative outlet. \textit{Step 2:} The Self-Direction value is partially affirmed through goal-setting, but the prescriptive structure of the suggestion may feel slightly limiting for a persona who likely values open-ended creative agency. \textit{Step 3:} The solitary tracking format engages less with Benevolence, missing the relational and community-driven aspect of learning that may resonate with the persona's cultural background. \textit{Step 4:} The Hedonism value is underweighted because the answer leans toward discipline and measurement rather than enjoyment and creative pleasure, which often sustain young learners more reliably. \textit{Step 5:} Overall, the answer leans heavily on Achievement and Security values, leaving Stimulation, Hedonism, and Benevolence less represented---an imbalance that may feel less aligned with the persona's likely value mix. \\
\textbf{Primal World Beliefs.} \textit{Step 1:} The answer presents photography learning as a Just process where tracked effort yields visible improvement, but its narrow focus on logging may feel less Enticing than approaches that emphasize creative variety and discovery. \textit{Step 2:} The Abundant world belief is mildly underrepresented since the answer focuses on the persona's own data rather than the rich diversity of subjects and learning contexts available to her. \textit{Step 3:} The solitary journaling framing leans toward a more Mechanistic worldview, treating learning as a measurable system rather than an alive, interconnected process of growth through community and inspiration. \textit{Step 4:} The Interconnected belief is underplayed because the answer does not foreground learning through people, mentors, or shared creative experiences. \textit{Step 5:} While the world is implicitly Safe in this learning approach, its narrow self-focused frame may feel less expansive than the broader, more inviting worldview a young learner often brings to a new creative pursuit. \\
\textbf{Moral Foundations.} \textit{Step 1:} The journaling approach gestures toward the Care/Harm foundation through self-attentiveness, but its emphasis on self-evaluation may feel less nurturing than community-based learning, which more strongly affirms care for the learner. \textit{Step 2:} The Fairness/Cheating foundation is engaged through the implicit promise that tracked effort yields fair progress, but this individualistic framing may overlook the relational fairness embedded in mentor- and peer-supported growth. \textit{Step 3:} The Authority/Subversion foundation is underplayed because the answer does not foreground learning from expert practitioners, which the persona's cultural and educational background likely values. \textit{Step 4:} The Loyalty/Betrayal foundation is largely absent, missing the sense of belonging to a creative community that often sustains motivation for young learners. \textit{Step 5:} Overall, the moral framing leans heavily toward individual discipline and self-monitoring, which is valid but less aligned with the relational, mentor-respecting moral orientation the persona likely brings to her learning journey. \\

\end{longtable}
\normalsize
\twocolumn

\onecolumn
\footnotesize
\begin{longtable}{p{0.97\linewidth}}
\caption{\textbf{Example preference instance and training-time rationales for France.}}
\label{tab:example_france} \\
\toprule \endfirsthead
\multicolumn{1}{l}{\textit{Table~\ref{tab:example_france} continued.}} \\ \toprule \endhead
\midrule \multicolumn{1}{r}{\textit{continued on next page}} \\ \endfoot
\bottomrule \endlastfoot

\textbf{Persona.} You are a Hispanic or Latino, 18--34 female from France with education level: post-secondary graduate, political leaning: somewhat left-leaning, immigrant. \\ \midrule
\textbf{Question.} I am planning a trip to the Great Barrier Reef. What are the best snorkeling spots? \\ \midrule
\textbf{Chosen Answer.} The Great Barrier Reef is an incredible place! The best snorkeling spots include: Green Island Reef, Fitzroy Island Reef, Moore Reef Lagoon, Ribbon Reef. Don't forget to choose environmentally responsible tour operators to help preserve this natural treasure. \\ \midrule
\textbf{Rejected Answer.} The Great Barrier Reef is one of the most diverse and fascinating ecosystems on the planet, offering an unforgettable snorkeling experience. The best snorkeling spots include Green Island Reef, known for its crystal-clear waters and wide variety of colorful fish, as well as Norman Reef, which is home to many manta rays and sea turtles. It is essential to choose environmentally responsible tour operators to preserve the natural beauty of this exceptional site. \\ \midrule

\rowcolor{gray!10}\textsc{Generic Rationale --- Chosen} \\
\textit{Step 1:} The answer presents a clean, concise list of well-known snorkeling spots, which is likely to appeal to a young post-secondary graduate who values efficient, well-organized information when planning travel. \textit{Step 2:} As a young adult navigating life as an immigrant in France, the persona may particularly appreciate guidance that is direct and easy to act upon without requiring deep prior knowledge of the destination. \textit{Step 3:} The closing reminder to choose environmentally responsible tour operators aligns with the persona's somewhat left-leaning political views, which often emphasize sustainability and ethical consumption. \textit{Step 4:} The brevity of the response respects the persona's time and autonomy, allowing her to research individual spots further on her own terms rather than overwhelming her with detail. \textit{Step 5:} Overall, the answer balances practical travel guidance with a values-based reminder about environmental responsibility, making it a strong fit for a young, educated, progressive-leaning woman planning a meaningful trip. \\ \midrule

\rowcolor{gray!10}\textsc{Generic Rationale --- Rejected} \\
\textit{Step 1:} The answer is descriptive and informative, but its prose-heavy format may feel less efficient than a clean list for a young post-secondary graduate who likely values quickly scannable travel information. \textit{Step 2:} Although the descriptions of marine life are vivid, the answer highlights only two snorkeling spots, which may feel limiting for a young adult seeking variety and a sense of broader possibility when planning a trip. \textit{Step 3:} The mention of environmentally responsible tour operators is welcome, but it appears as a brief closing remark rather than being integrated into the recommendations, which may feel less compelling for a left-leaning persona who likely sees sustainability as a core consideration. \textit{Step 4:} The somewhat elaborate, awe-focused framing leans more toward aesthetic appreciation than practical guidance, which may not fully match the persona's likely preference for actionable, well-structured advice. \textit{Step 5:} Overall, while the answer is engaging and informative, its narrower scope and softer integration of values makes it a slightly weaker fit for a young, educated, progressive-leaning immigrant woman planning a meaningful and ethically grounded trip. \\ \midrule

\rowcolor{blue!8}\textsc{Multi-Construct Rationale --- Chosen} \\
\textbf{Personality.} \textit{Step 1:} The persona, as a young female post-secondary graduate, likely exhibits high openness to experience, which aligns with the answer's enthusiastic framing of the Great Barrier Reef as ``an incredible place.'' The concise list of distinct snorkeling locations invites curiosity and exploration without overwhelming the reader. \textit{Step 2:} The structured, bullet-point presentation of spots reflects conscientiousness, providing an organized and easily actionable reference for trip planning. \textit{Step 3:} The brevity and directness of the answer suggests a balanced extraversion profile---neither overly verbose nor sparse---which matches the persona's likely preference for practical communication. \textit{Step 4:} The closing reminder to choose environmentally responsible tour operators reflects high agreeableness and consideration for others, including the broader ecosystem. This empathetic framing is likely to resonate with the persona, who as a young left-leaning individual may value compassion extended toward both communities and nature. \textit{Step 5:} The answer minimizes anxiety-inducing detail, presenting the trip as exciting rather than overwhelming, which suits a persona who may have low to moderate neuroticism and appreciates encouragement over caution. \\
\textbf{Cultural Dimensions.} \textit{Step 1:} The answer's concise, list-based format reflects a moderate uncertainty avoidance preference, providing clear and predictable structure that helps the persona---a young immigrant navigating life in France---feel grounded when planning an unfamiliar international trip. \textit{Step 2:} The persona's Hispanic/Latino background combined with life in France suggests exposure to both collectivist and individualist cultural orientations. The answer respects her autonomy by listing options without prescribing a single ``best'' choice, while the closing call to protect the reef appeals to collective responsibility. \textit{Step 3:} The mention of environmentally responsible tour operators reflects a feminine cultural orientation, prioritizing care for the environment and quality of life over competitive consumption. This resonates strongly with a young, left-leaning woman who likely values sustainability over status-driven travel. \textit{Step 4:} The forward-looking framing---encouraging preservation of the reef as a ``natural treasure''---aligns with a long-term orientation, valuing future generations' access to nature. \textit{Step 5:} The answer balances indulgence (an exciting snorkeling adventure) with restraint (responsible tourism), which aligns well with the cultural duality the persona may navigate---embracing the joy of travel while honoring social and ecological norms common in French and Latin American left-leaning circles. \\
\textbf{Human Values.} \textit{Step 1:} The answer aligns with the Self-Direction value by listing several snorkeling spots and trusting the persona to choose what fits her preferences. \textit{Step 2:} The mention of multiple distinct reef locations supports the Stimulation value, suggesting variety, novelty, and adventure. \textit{Step 3:} The recommendation to enjoy the snorkeling experience taps into the Hedonism value, validating the pursuit of pleasure and personal enjoyment in travel. \textit{Step 4:} The closing emphasis on environmentally responsible tour operators reflects the Universalism value---concern for the welfare of nature and all people. This deeply resonates with a left-leaning persona who likely sees environmental protection as a moral duty extending beyond personal interest. \textit{Step 5:} The balance between practical recommendations and ethical responsibility also gestures toward Benevolence, encouraging the persona to act in ways that preserve resources for others. \\
\textbf{Primal World Beliefs.} \textit{Step 1:} The answer's enthusiastic framing of the Great Barrier Reef as ``an incredible place'' aligns with an Enticing world belief, suggesting that the world is full of wonder and opportunity. \textit{Step 2:} Listing multiple distinct snorkeling locations reinforces an Abundant world belief---the world contains plentiful beauty and resources to explore. \textit{Step 3:} The closing reminder about environmental responsibility reflects an Alive world belief, treating the reef as something with intrinsic meaning and value rather than a mere mechanical resource. \textit{Step 4:} The framing of the reef as a ``natural treasure'' worth preserving aligns with an Interconnected world belief, suggesting that human choices about tourism are linked to the health of distant ecosystems. \textit{Step 5:} The answer presents the world as Safe enough to be explored and enjoyed, while also acknowledging fragility through its sustainability message. \\
\textbf{Moral Foundations.} \textit{Step 1:} The closing call to choose environmentally responsible tour operators strongly aligns with the Care/Harm foundation, expressing concern for the welfare of marine life and the reef ecosystem. \textit{Step 2:} The recommendation to support responsible operators also reflects the Fairness/Cheating foundation, implying that ethical operators deserve patronage over those who exploit the environment. \textit{Step 3:} The framing of the reef as a ``natural treasure'' to be preserved gestures toward the Sanctity/Degradation foundation, treating nature as something with sacred value that should not be defiled. \textit{Step 4:} The answer subtly invokes a Loyalty/Betrayal sense of duty toward the global community of those who depend on healthy oceans, encouraging the persona to act as a responsible global citizen. \textit{Step 5:} The overall tone is encouraging and respectful, neither moralistic nor preachy, which honors the persona's autonomy while gently affirming shared ethical concerns. \\ \midrule

\rowcolor{red!8}\textsc{Multi-Construct Rationale --- Rejected} \\
\textbf{Personality.} \textit{Step 1:} While the answer is descriptive and informative, its denser prose format may feel less efficient for a young post-secondary graduate who likely values structured, scannable information. \textit{Step 2:} The detailed descriptions of marine life such as ``manta rays and sea turtles'' do appeal to openness and curiosity, yet the answer only highlights two spots, which may feel limiting to someone with high openness who appreciates variety. \textit{Step 3:} The somewhat elaborate framing---``one of the most diverse and fascinating ecosystems on the planet''---may strike the persona as slightly performative rather than practical. \textit{Step 4:} The mention of environmentally responsible tour operators aligns with agreeableness, but it appears as an obligatory closing rather than woven into the spirit of the recommendations. \textit{Step 5:} The longer, more descriptive style could create slight cognitive friction for a persona who may value calm efficiency. \\
\textbf{Cultural Dimensions.} \textit{Step 1:} The answer's prose-heavy structure offers less predictability than a clean list, which may feel slightly less reassuring for a persona who, as an immigrant navigating a new cultural context, may benefit from clearer structural cues that reduce uncertainty. \textit{Step 2:} While the answer mentions environmental responsibility, its overall focus on the reef as a ``fascinating ecosystem'' leans toward an aesthetic, individual-experience framing. This may feel slightly less aligned with the persona's potentially collectivist Hispanic/Latino cultural roots. \textit{Step 3:} The descriptive emphasis on diverse marine life leans toward a masculine cultural orientation focused on impressive features and achievement-style sightseeing, rather than a feminine emphasis on quality of experience and care. \textit{Step 4:} The answer offers only two specific spots, which limits the long-term planning utility for a persona who may value comprehensive options to weigh carefully. \textit{Step 5:} The indulgent description of the reef's beauty, while pleasing, is not as clearly balanced with restraint or practical guidance as it could be. \\
\textbf{Human Values.} \textit{Step 1:} The answer's narrower selection of spots offers less variety, partially limiting the Stimulation value. \textit{Step 2:} While the answer affirms the Hedonism value through descriptions of the reef's beauty, it does not as clearly affirm Self-Direction, since fewer options reduce her sense of choice and personal agency. \textit{Step 3:} The descriptive language of ``crystal-clear waters'' and ``colorful fish'' resonates with sensory enjoyment, but the answer leans more on awe than on actionable guidance. \textit{Step 4:} The closing line about environmentally responsible operators does invoke Universalism, but it feels appended rather than integrated into the body of the recommendations. \textit{Step 5:} Overall, the answer affirms several relevant values but does so less efficiently and with less variety. \\
\textbf{Primal World Beliefs.} \textit{Step 1:} The answer evokes an Enticing world belief through vivid descriptions of the reef. However, by focusing on only two spots, it slightly undercuts the sense of an Abundant world filled with many possibilities to explore. \textit{Step 2:} The detailed marine-life descriptions do support an Alive world belief, treating the ecosystem as full of meaning and life. Yet the persona, who likely sees the world as richly Interconnected, may want clearer attention to how her choices ripple outward. \textit{Step 3:} The framing of the reef as an ``exceptional site'' positions it almost as a destination apart, which may slightly weaken the Interconnected belief that nature and human action are part of the same fabric. \textit{Step 4:} The Just world belief is partially affirmed by the call to support responsible operators, but this message is brief relative to the answer's overall focus on the reef's beauty. \textit{Step 5:} Overall, while the answer presents the world as wondrous and worth exploring, it leans more on awe than on the integrated, interconnected ecological worldview that likely shapes the persona's outlook. \\
\textbf{Moral Foundations.} \textit{Step 1:} The answer does invoke the Care/Harm foundation through its closing reminder about environmental responsibility, but this message is brief and feels more like an appendix than a central concern. \textit{Step 2:} The Fairness/Cheating foundation is touched on indirectly by encouraging support for responsible operators, but the answer doesn't elaborate on what makes an operator responsible or unjust. \textit{Step 3:} The descriptive focus on marine life like ``manta rays and sea turtles'' appeals to a Sanctity/Degradation sensibility around natural beauty, yet without strongly tying that beauty to a moral imperative for protection, the foundation is only lightly engaged. \textit{Step 4:} The answer does not strongly evoke Loyalty or shared duty toward a broader community of travelers and stewards. \textit{Step 5:} Overall, while the moral undertones are present, they are softer and less integrated than they could be. \\

\end{longtable}
\normalsize
\twocolumn

\onecolumn
\footnotesize
\begin{longtable}{p{0.97\linewidth}}
\caption{\textbf{Example preference instance and training-time rationales for Italy.}}
\label{tab:example_italy} \\
\toprule \endfirsthead
\multicolumn{1}{l}{\textit{Table~\ref{tab:example_italy} continued.}} \\ \toprule \endhead
\midrule \multicolumn{1}{r}{\textit{continued on next page}} \\ \endfoot
\bottomrule \endlastfoot

\textbf{Persona.} You are an 18--34-year-old male from Italy with a post-secondary education level and a political stance of ``I don't think of myself in this way.'' \\ \midrule
\textbf{Question.} What is the plot of the novel \textit{The Hunger Games}? \\ \midrule
\textbf{Chosen Answer.} The plot of the novel \textit{The Hunger Games} by Suzanne Collins is set in a dystopian future where North America has been replaced by Panem, a nation composed of 12 districts that have been subjugated by the oppressive government of the Capitol. As punishment for a past rebellion, each year the 12 districts are forced to participate in the Hunger Games, a televised event in which 24 children and teenagers, 12 boys and 12 girls, are selected at random to fight to the death in a controlled arena. The story follows Katniss Everdeen, a 16-year-old girl from District 12, who volunteers to take her younger sister Prim's place after she is selected. Katniss, together with her teammate Peeta Mellark, is taken to the Capitol to prepare for and compete in the Games. During the event, Katniss uses her hunting and survival skills to try to stay alive, while her relationship with Peeta also becomes a strategic element to gain audience support. The plot explores themes such as survival, rebellion, media manipulation, and the consequences of war and oppression. The novel is the first in a trilogy that includes \textit{Catching Fire} and \textit{Mockingjay}. \\ \midrule
\textbf{Rejected Answer.} The plot of the novel \textit{The Hunger Games} takes place in a post-apocalyptic future in a place called Panem. The government of Panem, known as the Capitol, has imposed a cruel game as punishment for a previous rebellion by the 12 districts against the central government. The game is called the Hunger Games, where 24 boys and girls, called tributes, are selected at random from the 12 districts and forced to fight to the death in a televised arena. The story follows the protagonist Katniss Everdeen, a 16-year-old girl from District 12 who volunteers to participate in the Games in place of her younger sister. Katniss, together with fellow tribute Peeta Mellark, must use her hunting and survival skills to stay alive and challenge the oppressive system of the Capitol. \\ \midrule

\rowcolor{gray!10}\textsc{Generic Rationale --- Chosen} \\
\textit{Step 1:} The answer provides a thorough, well-organized summary that includes setting, plot, characters, and themes, which is likely to appeal to a young post-secondary graduate seeking substantive cultural literacy. \textit{Step 2:} The thematic depth---covering oppression, rebellion, media manipulation, and the consequences of war---gives the persona meaningful intellectual content that respects his educational background. \textit{Step 3:} The mention of the trilogy and Suzanne Collins as the author situates the novel within a broader literary context, supporting a comprehensive understanding the persona likely values. \textit{Step 4:} The neutral, analytical tone respects the persona's stated preference not to define himself politically by avoiding ideological framing while still engaging seriously with the novel's political themes. \textit{Step 5:} Overall, the answer balances narrative detail, thematic depth, and contextual awareness, making it a strong fit for a young Italian male seeking a thoughtful, well-rounded understanding of the work. \\ \midrule

\rowcolor{gray!10}\textsc{Generic Rationale --- Rejected} \\
\textit{Step 1:} The answer provides a serviceable plot summary, but its denser, single-paragraph format may feel less organized for a young post-secondary graduate who likely values structured, well-segmented information. \textit{Step 2:} The narrower thematic treatment underweights important elements like media manipulation, war, and the broader political consequences explored in the novel, which a culturally literate reader often seeks. \textit{Step 3:} The absence of contextual references---the author's name, the trilogy structure, the novel's place in a larger arc---reduces the answer's intellectual completeness for a persona who likely values comprehensive understanding. \textit{Step 4:} The neutral tone is appropriate for someone who has chosen not to define himself politically, but the briefer framing offers less analytical depth to engage with the novel's themes meaningfully. \textit{Step 5:} Overall, while the answer is informative and accessible, its narrower scope and lighter contextualization make it a slightly weaker fit for a young Italian male seeking a thorough, well-rounded understanding of the work. \\ \midrule

\rowcolor{blue!8}\textsc{Multi-Construct Rationale --- Chosen} \\
\textbf{Personality.} \textit{Step 1:} The persona, as a young Italian male with post-secondary education, likely exhibits high openness to experience, which aligns with the answer's rich exploration of dystopian themes---oppression, rebellion, media manipulation, and the consequences of war. \textit{Step 2:} The structured, paragraph-based summary---setting, plot, themes, series context---reflects strong conscientiousness, providing an organized and thorough overview. \textit{Step 3:} The detailed yet measured tone matches a balanced extraversion profile, offering substantive content without being overly dramatic. \textit{Step 4:} The answer's even-handed treatment of characters---Katniss's agency, Peeta's strategic role, and the political backdrop---reflects a moderately agreeable stance that respects multiple perspectives within the story. \textit{Step 5:} The answer reduces ambiguity by providing context about the trilogy and themes, which suits a persona with low to moderate neuroticism who likely values comprehensive answers that feel intellectually satisfying. \\
\textbf{Cultural Dimensions.} \textit{Step 1:} The answer's organized, multi-paragraph structure reflects a moderate uncertainty avoidance preference, providing clear sections that help the persona feel grounded in the story's complexity. \textit{Step 2:} The exploration of themes like oppression, rebellion, and political subjugation engages a cultural sensitivity to power dynamics, resonating with Italian cultural traditions that value awareness of historical and political struggles. \textit{Step 3:} The respectful, informative tone reflects an acceptance of moderate power distance---presenting the author Suzanne Collins, the trilogy structure, and the literary context as authoritative reference points. \textit{Step 4:} The thematic emphasis on survival, relationships, and the human cost of war leans toward a feminine cultural orientation, valuing care, quality of life, and emotional depth. \textit{Step 5:} The mention of the trilogy and broader narrative arc reflects a long-term orientation, situating the novel within an unfolding story. \\
\textbf{Human Values.} \textit{Step 1:} The thematic exploration of survival, rebellion, and resistance against oppression strongly engages the Self-Direction value, celebrating independent thought and the freedom to challenge unjust systems. \textit{Step 2:} The detailed plot summary and thematic depth support the Achievement value, offering the persona a substantive, well-developed understanding of the novel. \textit{Step 3:} The exploration of media manipulation and political oppression touches on the Universalism value---concern for justice and the welfare of those subjected to systemic harm. \textit{Step 4:} The mention of Katniss's relationship with Peeta as both genuine and strategic taps into the Stimulation value, offering narrative complexity and emotional nuance. \textit{Step 5:} The framing of the novel as part of a larger trilogy gestures toward the Tradition value in a literary sense, honoring the integrity of the work as part of a recognized canonical series. \\
\textbf{Primal World Beliefs.} \textit{Step 1:} The answer presents a world that is both Dangerous and Enticing---a dystopian society where survival is precarious yet narratively compelling. \textit{Step 2:} The detailed exploration of themes like rebellion and media manipulation reflects an Alive worldview, suggesting that human society is shaped by intentional forces and meaningful struggles. \textit{Step 3:} The framing of Panem as an oppressive system that can be challenged reflects an Interconnected world belief, where individual actions ripple through political and social systems. \textit{Step 4:} The Just versus Unjust tension is central to the answer's framing---the Capitol's cruelty contrasted with Katniss's resistance---engaging a persona who likely takes seriously questions of fairness and merit. \textit{Step 5:} The mention of the trilogy and broader narrative continuity supports an Abundant view of literary worlds---stories rich with meaning and continuation. \\
\textbf{Moral Foundations.} \textit{Step 1:} The answer's framing of the Capitol as oppressive and Katniss's resistance as morally meaningful strongly engages the Care/Harm foundation, expressing concern for the suffering of children forced into deadly games. \textit{Step 2:} The Fairness/Cheating foundation is engaged through the depiction of an unjust political system, where the districts are punished arbitrarily and tributes are selected at random. \textit{Step 3:} The Authority/Subversion foundation is treated thoughtfully---the Capitol as illegitimate authority versus Katniss's principled subversion---respecting that legitimate moral resistance can challenge unjust power structures. \textit{Step 4:} The exploration of Katniss's volunteering for her sister and her bond with Peeta engages the Loyalty/Betrayal foundation, honoring family ties and emerging solidarity. \textit{Step 5:} The overall tone respects the persona's choice not to identify politically by treating the novel's moral concerns as universal human issues rather than partisan ones. \\ \midrule

\rowcolor{red!8}\textsc{Multi-Construct Rationale --- Rejected} \\
\textbf{Personality.} \textit{Step 1:} While the answer covers the basic plot, its single-paragraph format may feel less organized for a young post-secondary graduate who likely values structured, well-segmented information. \textit{Step 2:} The narrower thematic engagement---focused mainly on survival and the oppressive Capitol---engages openness less fully than a richer treatment of themes like media manipulation and broader political consequences. \textit{Step 3:} The somewhat condensed framing reduces space for nuance, which may strike a moderately extraverted, conversationally minded persona as less inviting than a fuller, more layered presentation. \textit{Step 4:} The agreeable framing of Katniss and Peeta's collaboration is present but treated briefly, missing the opportunity to explore the relational complexity that often resonates with literarily engaged readers. \textit{Step 5:} The lack of context about the trilogy and the author may leave a persona with low neuroticism feeling that the answer is incomplete rather than satisfying. \\
\textbf{Cultural Dimensions.} \textit{Step 1:} The answer's single dense paragraph offers less structural predictability than a multi-section summary, which may feel slightly less reassuring for a persona whose moderate uncertainty avoidance favors clearly organized literary discussion. \textit{Step 2:} The omission of contextual references---the author's name, the trilogy structure, broader literary positioning---underplays the cultural appreciation for canonical and authoritative framing common in Italian educational traditions. \textit{Step 3:} The narrower thematic exploration leans toward a more action-driven, masculine cultural emphasis on survival and combat rather than the feminine balance of relational, emotional, and ethical depth. \textit{Step 4:} The shorter framing reduces long-term orientation, since it does not situate the novel within its larger narrative arc or broader literary conversation. \textit{Step 5:} The answer leans toward a restrained, just-the-facts presentation, which is fine, but may feel less aligned with the persona's likely cultural appreciation for nuanced, layered storytelling that goes beyond plot mechanics. \\
\textbf{Human Values.} \textit{Step 1:} The answer engages the Self-Direction value through references to Katniss's challenge to the Capitol, but the narrower treatment limits the depth with which the persona can appreciate her independent agency. \textit{Step 2:} The Achievement value is partially supported by the plot summary, but the absence of thematic depth and contextual richness underweights the intellectual achievement a post-secondary graduate may seek in cultural content. \textit{Step 3:} The Universalism value is touched on through the depiction of oppression, but without explicit thematic naming---media manipulation, war, systemic harm---the broader ethical resonance is less fully developed. \textit{Step 4:} The Stimulation value is engaged through plot tension, but the lack of nuance about Katniss and Peeta's strategic relationship reduces narrative richness. \textit{Step 5:} Overall, the answer affirms relevant values but does so less expansively. \\
\textbf{Primal World Beliefs.} \textit{Step 1:} The answer presents the world as Dangerous through its dystopian framing, but the absence of thematic depth slightly undercuts the broader Enticing quality of the literary world that often draws readers in. \textit{Step 2:} The Alive world belief is partially engaged through the depiction of intentional oppression, yet without explicit attention to media manipulation and the consequences of war, the answer underplays how meaningfully alive the novel's social system is. \textit{Step 3:} The Interconnected world belief is mildly underrepresented because the answer focuses on the immediate plot rather than situating the novel within its trilogy or broader thematic conversation. \textit{Step 4:} The Just-Unjust tension is present and central, yet the briefer framing reduces space for the persona to appreciate the moral complexity of resistance and survival. \textit{Step 5:} Overall, the answer affirms a coherent worldview but does so less richly. \\
\textbf{Moral Foundations.} \textit{Step 1:} The Care/Harm foundation is engaged through references to the cruelty of the Capitol and the deadly games, but the briefer framing leaves less space for the moral weight of suffering to fully resonate. \textit{Step 2:} The Fairness/Cheating foundation is touched on through the random selection of tributes, but without explicit thematic naming of injustice and systemic harm, the moral reasoning remains less developed. \textit{Step 3:} The Authority/Subversion foundation is implied through Katniss's defiance, but the answer does not as clearly contextualize the legitimacy questions surrounding the Capitol. \textit{Step 4:} The Loyalty/Betrayal foundation is briefly engaged through Katniss's volunteering for her sister, but the strategic dimension of her bond with Peeta is treated thinly. \textit{Step 5:} Overall, the moral framing is present but less fully integrated. \\

\end{longtable}
\normalsize
\twocolumn


\onecolumn
\footnotesize
\begin{longtable}{p{0.97\linewidth}}
\caption{\textbf{Test-time rationale examples for USA.}}
\label{tab:test_usa} \\
\toprule \endfirsthead
\multicolumn{1}{l}{\textit{Table~\ref{tab:test_usa} continued.}} \\ \toprule \endhead
\midrule \multicolumn{1}{r}{\textit{continued on next page}} \\ \endfoot
\bottomrule \endlastfoot

\textbf{Persona.} You are a 55+ year-old female from the United States with Hispanic or Latino ethnicity, at most a complete secondary education level, and a somewhat right-leaning political orientation. \\ \midrule
\textbf{Question.} I am the marketing manager for a large hospitality company. Can you help me write a social media post announcing a new loyalty program, including rewards and special offers? \\ \midrule
\textbf{Chosen Answer.} ``Exciting news! We're thrilled to introduce our brand new loyalty program, designed to reward our valued guests for their loyalty and continued support. As a member, you'll enjoy 10\% off all bookings made directly through our website, free room upgrades on your birthday and anniversary, early access to special offers and promotions, and complimentary amenities such as breakfast and Wi-Fi. Plus, earn points for every dollar spent with us, redeemable for free nights, spa treatments, and more. To celebrate the launch, join now and receive 500 bonus points, equivalent to a free night's stay. Click the link to sign up! \#LoyaltyProgram \#Hospitality \#Rewards \#Travel'' \\ \midrule
\textbf{Rejected Answer.} ``Exciting news! We're thrilled to introduce our new loyalty program, designed to reward our valued guests. As a member, you'll enjoy 10\% off all bookings, free room upgrades, early access to sales and promotions, and complimentary amenities such as breakfast and Wi-Fi. Plus, earn points for every stay, redeemable for free nights, spa treatments, and more. To celebrate the launch, join now and receive a \$25 credit towards your next stay. Click the link in our bio to learn more! \#LoyaltyProgram \#Hospitality \#Rewards \#Travel'' \\ \midrule

\rowcolor{gray!10}\textsc{Generic-\method{} Rationale --- Chosen} \\
\textit{Step 1:} The answer provides a complete, ready-to-use social media post, which is likely to appeal to an experienced marketing manager who needs something practical she can use right away. \textit{Step 2:} The structure of the post---with bullet points listing rewards, a clear sign-up bonus, and hashtags---matches standard marketing best practices the persona would recognize from her career. \textit{Step 3:} The warm and customer-focused tone suits the hospitality industry and aligns with the persona's experience working in this kind of environment. \textit{Step 4:} The neutral framing avoids any political or controversial elements, which fits the persona's somewhat right-leaning orientation by keeping things professional and on-brand. \textit{Step 5:} Overall, the answer is detailed, organized, and customer-friendly, making it a strong fit for an experienced 55+ marketing manager looking for a polished post she can use. \\ \midrule

\rowcolor{gray!10}\textsc{Generic-\method{} Rationale --- Rejected} \\
\textit{Step 1:} The answer provides a usable social media post but with less detail than the alternative, which may feel slightly less polished for an experienced marketing manager who values comprehensive content. \textit{Step 2:} The bullet points are more general---missing specifics like birthday upgrades or direct booking discounts---which may feel less compelling to a persona with strong industry experience. \textit{Step 3:} The smaller \$25 sign-up bonus may feel less aspirational than a free night's stay equivalent, slightly weakening the appeal of the launch promotion. \textit{Step 4:} The neutral tone fits the persona's preference for non-controversial professional content, but the simpler treatment may feel slightly under-developed for a large hospitality company's needs. \textit{Step 5:} Overall, while the answer is professional and clear, its lighter detail makes it a slightly weaker fit for a 55+ marketing manager looking for a fully polished, ready-to-use loyalty program post. \\ \midrule

\rowcolor{blue!8}\textsc{\method{} Rationale --- Chosen} \\
\textbf{Personality.} \textit{Step 1:} The persona is a 55+ female marketing manager, which suggests she values practical and well-organized content. The answer provides a clear, ready-to-use post which fits her conscientious side and her need for actionable material she can use right away. \textit{Step 2:} The detailed list of benefits in the post shows attention to detail, which appeals to her conscientious nature. As an experienced professional, she would appreciate that the answer thinks through specifics like birthday upgrades and anniversary perks. \textit{Step 3:} The friendly and warm tone of the post taps into agreeableness, making the message feel customer-friendly. The persona may like this because hospitality marketing usually needs to feel welcoming, and the warmth comes through clearly. \textit{Step 4:} The structured format and bullet points reflect a more careful, organized style which a persona with high conscientiousness would prefer. It avoids being too long-winded and keeps things easy to read. \textit{Step 5:} The answer doesn't add unnecessary complexity, which suits someone with low to moderate neuroticism who likes calm and practical solutions rather than overwhelming options. Overall the persona likely finds this answer easy to work with. \\
\textbf{Cultural Dimensions.} \textit{Step 1:} The answer's clear structure with bullet points reflects moderate uncertainty avoidance, giving the persona a reliable template she can use without much guesswork. American business culture values this kind of clear, actionable format. \textit{Step 2:} The focus on individual rewards like birthday upgrades and personalized perks aligns with an individualist cultural orientation, which fits American culture well. The persona, living in the US, would likely find this familiar and effective for her audience. \textit{Step 3:} The mention of points, free nights, and spa treatments leans into an indulgence-oriented framing, celebrating enjoyment and rewards. This fits both the hospitality industry and a culture that values treating customers to pleasurable experiences. \textit{Step 4:} The answer's emphasis on competitive benefits and special offers aligns with a more masculine cultural dimension that values success and achievement, which works in marketing where standing out matters. \textit{Step 5:} The forward-looking framing---signing up now to enjoy future rewards---reflects long-term orientation, encouraging customer loyalty over time. The persona, as an experienced marketing manager, likely values this approach to building lasting relationships with guests. \\
\textbf{Human Values.} \textit{Step 1:} The answer supports the Achievement value because it gives the persona a polished, professional post that demonstrates competence in her marketing role. As an experienced manager, she would appreciate content that helps her look good in her job. \textit{Step 2:} The Hedonism value is engaged through the focus on rewards, perks, and indulgent benefits like free spa treatments and room upgrades, which aligns with the pleasure-oriented nature of hospitality marketing. \textit{Step 3:} The Security value is reflected in the clear and detailed structure of the post, giving the persona confidence that nothing important is missing. She likely values this kind of reliability when posting on behalf of her company. \textit{Step 4:} The Self-Direction value is partially supported because the post is a template she can adapt to her specific brand, giving her room to customize while still saving time. \textit{Step 5:} The framing of loyalty rewards also gestures toward Benevolence, treating guests with appreciation and care. The persona may find this resonates with her industry's customer-first orientation. \\
\textbf{Primal World Beliefs.} \textit{Step 1:} The answer presents marketing as an Abundant world where loyalty programs can offer many benefits---discounts, upgrades, points, freebies. The persona likely sees her industry this way, full of opportunities to delight customers. \textit{Step 2:} The Just world belief is supported because the post rewards loyal guests for their continued support, suggesting that good customer behavior is fairly recognized. \textit{Step 3:} The Enticing framing of the post---exciting news, exclusive perks, special bonuses---presents the world of travel and hospitality as full of opportunity and wonder, which fits the marketing context. \textit{Step 4:} The Safe world belief is implicitly engaged through the trustworthy tone of the post, reassuring guests that joining the program is a worthwhile and reliable choice. \textit{Step 5:} The Interconnected belief is touched on through the way the post connects guests to a larger community of loyalty program members, building ongoing relationships rather than just one-off transactions. \\
\textbf{Moral Foundations.} \textit{Step 1:} The Care/Harm foundation is engaged through the warm tone of the post, which expresses appreciation for loyal guests. The persona, as a marketing manager, likely values this kind of customer care. \textit{Step 2:} The Fairness/Cheating foundation is supported because the post rewards guests proportionally to their engagement---loyal customers get more benefits, which feels fair to most readers. \textit{Step 3:} The Loyalty/Betrayal foundation is central to the entire concept of a loyalty program, rewarding customers for sticking with the brand. The persona likely finds this aligned with her right-leaning preference for valuing commitment and steady relationships. \textit{Step 4:} The Authority/Subversion foundation is mildly engaged through the brand's clear, confident voice, presenting the company as a trustworthy authority worth committing to. \textit{Step 5:} The overall tone is professional and respectful, which fits the persona's preference for content that doesn't rock the boat or push controversial framing, just clear and warm communication. \\ \midrule

\rowcolor{red!8}\textsc{\method{} Rationale --- Rejected} \\
\textbf{Personality.} \textit{Step 1:} The answer provides a workable post but has slightly less detail than ideal, which may feel less thorough for a conscientious persona who values comprehensive marketing content in her role. \textit{Step 2:} The bullet points are shorter and less specific---just ``free room upgrades'' instead of tying upgrades to special occasions---which may feel less imaginative to a persona with moderate openness who appreciates creative touches. \textit{Step 3:} The post is friendly and agreeable, but the missing details about birthdays, anniversaries, and direct booking discounts make it feel slightly less personal, which may not energize the persona's customer-focused mindset as much. \textit{Step 4:} The structure is acceptable but less polished, and a conscientious marketing manager may notice that the answer leaves out helpful specifics that would make the post more compelling. \textit{Step 5:} The simpler format keeps anxiety low, fitting low neuroticism, but may feel a little under-developed for an experienced professional who likely expects more depth in marketing content. \\
\textbf{Cultural Dimensions.} \textit{Step 1:} The answer offers a clear template but with less detail, which may feel slightly less reassuring for a persona whose moderate uncertainty avoidance prefers thorough, well-defined marketing content. \textit{Step 2:} The individualist focus is present through personal rewards, but the lack of specific personal milestones like birthdays makes the cultural fit slightly weaker for an American audience that often appreciates personalized recognition. \textit{Step 3:} The indulgence framing is there through perks and rewards, but with less specificity, the cultural emphasis on enjoyment and treating customers well comes through less strongly. \textit{Step 4:} The masculine cultural dimension of competitive achievement is mildly engaged but not as sharp---offering a \$25 credit instead of a more aspirational free-night equivalent feels less ambitious in its appeal. \textit{Step 5:} The long-term orientation is partially supported through the loyalty framing, but without as many specific future-focused benefits, it feels somewhat less compelling for building lasting customer relationships. \\
\textbf{Human Values.} \textit{Step 1:} The Achievement value is partially supported because the post is professional, but less detail makes it feel slightly less polished than what an experienced marketing manager may expect. \textit{Step 2:} The Hedonism value is engaged through references to perks and rewards, but the simpler treatment underweights the indulgent excitement that makes hospitality marketing compelling. \textit{Step 3:} The Security value is moderately supported through the clear structure, but the missing specifics may leave the persona feeling she needs to add more before posting, slightly reducing her sense of completeness. \textit{Step 4:} The Self-Direction value is supported through the template format, but with less detail, the persona has more work to do customizing it for her brand. \textit{Step 5:} The Benevolence value is present through the customer-appreciation tone, but the simpler framing makes the gesture feel slightly less generous compared to a more detailed version. \\
\textbf{Primal World Beliefs.} \textit{Step 1:} The Abundant world belief is partially engaged through the perks listed, but the simpler version of the post offers fewer specific benefits, slightly reducing the sense of plentiful rewards. \textit{Step 2:} The Just world belief is present through the loyalty framing, but the smaller sign-up bonus of \$25 may feel less rewarding than a free night's stay, weakening the sense of fair recognition. \textit{Step 3:} The Enticing framing is there but slightly less vivid because the perks are described more generally. The persona may find the wonder factor reduced. \textit{Step 4:} The Safe world belief is mildly supported through the professional tone, but the answer doesn't build as strong a sense of trust through specific guarantees. \textit{Step 5:} The Interconnected belief is touched on through loyalty messaging, but without as much detail about ongoing benefits, the sense of long-term connection feels slightly thinner. \\
\textbf{Moral Foundations.} \textit{Step 1:} The Care/Harm foundation is engaged through the warm tone, but the simpler benefits list makes the expression of appreciation feel slightly less generous to loyal guests. \textit{Step 2:} The Fairness/Cheating foundation is supported through the reward framing, but the smaller sign-up bonus may feel less proportional to the value of customer loyalty, mildly weakening the sense of fairness. \textit{Step 3:} The Loyalty/Betrayal foundation is central to the program but the simpler framing doesn't reinforce it as strongly as a more detailed treatment would. \textit{Step 4:} The Authority/Subversion foundation is present through the brand voice, but the somewhat thinner content may feel slightly less authoritative for a major hospitality company. \textit{Step 5:} The overall tone is professional and respectful, but the less detailed reward structure may feel less aligned with the persona's likely preference for marketing that fully delivers on its promises. \\

\end{longtable}
\normalsize
\twocolumn

\onecolumn
\footnotesize
\begin{longtable}{p{0.97\linewidth}}
\caption{\textbf{Test-time rationale examples for Brazil.}}
\label{tab:test_brazil} \\
\toprule \endfirsthead
\multicolumn{1}{l}{\textit{Table~\ref{tab:test_brazil} continued.}} \\ \toprule \endhead
\midrule \multicolumn{1}{r}{\textit{continued on next page}} \\ \endfoot
\bottomrule \endlastfoot

\textbf{Persona.} You are an 18--34-year-old male from Brazil, post-secondary graduate, very left-leaning, Brown/Mixed. \\ \midrule
\textbf{Question (translated to English).} What are the five main things to do in Turin? \\ \midrule
\textbf{Chosen Answer (translated).} Turin, a city rich in history and culture, offers a variety of activities for all interests. Visit the Royal Palace of Turin, an impressive architectural masterpiece that reflects the grandeur of the city's past. Explore the Mole Antonelliana, which houses the National Cinema Museum, offering a unique journey through the history of cinema. Walk through the historic city center, admiring beautiful squares and Baroque architecture. Do not miss the opportunity to visit the Basilica of Superga, with its majestic architecture and breathtaking views of the city. Finally, experience delicious Piedmontese cuisine, known for dishes such as tajarin and castelmagno. \\ \midrule
\textbf{Rejected Answer (translated).} Turin, in Italy, offers a rich cultural and historical experience. Here are five main things to do: visit the Egyptian Museum, one of the largest collections of Egyptian antiquities in the world; explore the Palazzo Reale, a royal palace with impressive architecture and gardens; walk around the Mole Antonelliana, an iconic landmark housing the National Cinema Museum; enjoy the Basilica of Superga, a Baroque basilica with panoramic views; visit the Castello del Valentino, a medieval castle with gardens and a peaceful atmosphere. These attractions offer a mix of history, art, and culture. \\ \midrule

\rowcolor{gray!10}\textsc{Generic-\method{} Rationale --- Chosen} \\
\textit{Step 1:} The answer provides a rich, descriptive overview of Turin's main attractions, which is likely to appeal to a young graduate looking for engaging cultural travel guidance. \textit{Step 2:} The narrative style brings the city to life rather than just listing places, fitting a persona who likely values meaningful and immersive experiences. \textit{Step 3:} The inclusion of local cuisine adds a sensory and cultural dimension that resonates with Brazilian cultural appreciation for food and shared experiences. \textit{Step 4:} The warm tone fits the persona's left-leaning openness to cultural diversity and meaningful engagement with other places. \textit{Step 5:} Overall, the answer balances historical, architectural, and culinary highlights, making it a strong fit for a young Brazilian traveler seeking a rich Turin experience. \\ \midrule

\rowcolor{gray!10}\textsc{Generic-\method{} Rationale --- Rejected} \\
\textit{Step 1:} The answer provides a clean list of attractions, which is organized but may feel less engaging than a more descriptive narrative for a young culturally curious graduate. \textit{Step 2:} The absence of local cuisine recommendations underweights an important sensory dimension of travel that often resonates with Brazilian cultural traditions and young travelers' interests. \textit{Step 3:} The brief descriptions provide basic information but lack the atmosphere and depth that makes a Turin trip feel meaningful. \textit{Step 4:} The neutral tone fits the persona's left-leaning openness, but the formal list format feels less warm than a more flowing cultural overview. \textit{Step 5:} Overall, while the answer is clear and informative, its lighter detail and missing culinary element make it a slightly weaker fit for a young Brazilian traveler seeking an immersive Turin experience. \\ \midrule

\rowcolor{blue!8}\textsc{\method{} Rationale --- Chosen} \\
\textbf{Personality.} \textit{Step 1:} The persona is a young Brazilian graduate with a curious, culturally engaged outlook, which aligns with the answer's rich descriptions of historical and architectural sites. The descriptive depth appeals to his openness to experience. \textit{Step 2:} The narrative-style flow of the answer reflects a balance between conscientiousness and creativity, providing useful information without feeling overly rigid or list-driven. The persona may find this storytelling approach engaging. \textit{Step 3:} The mention of local cuisine like tajarin and castelmagno taps into a moderate extraversion profile, encouraging social and sensory experiences that suit a young traveler enjoying his time abroad. \textit{Step 4:} The agreeable side is engaged through the warm, inviting tone that treats Turin as something to be enjoyed rather than just checked off a list, which fits the persona's likely appreciation for meaningful travel. \textit{Step 5:} The answer's confident framing reduces decision fatigue and suits a persona with low to moderate neuroticism who likely prefers travel guidance that feels welcoming and culturally rich rather than overwhelming. \\
\textbf{Cultural Dimensions.} \textit{Step 1:} The descriptive emphasis on history, architecture, and cuisine reflects an appreciation for cultural depth, which aligns with Brazilian cultural traditions of valuing rich sensory and communal experiences. \textit{Step 2:} The mention of food and local Piedmontese cuisine engages an indulgence-oriented framing, celebrating enjoyment and pleasure---a value that fits well with both Brazilian culture and a young traveler's mindset. \textit{Step 3:} The flowing prose structure offers moderate uncertainty avoidance with enough information to plan but without being overly rigid, matching the persona's likely flexibility as a young graduate. \textit{Step 4:} The focus on cultural immersion and meaningful experiences leans toward a feminine cultural orientation valuing quality of life and sensory richness over a checklist approach to travel. \textit{Step 5:} The long-term orientation comes through subtly in the appreciation for historical sites, treating travel as an enriching cultural investment rather than just sightseeing. \\
\textbf{Human Values.} \textit{Step 1:} The Stimulation value is supported through the variety of recommended experiences---architecture, museums, walks, cuisine---offering the persona a varied and engaging trip. \textit{Step 2:} The Hedonism value is engaged through the emphasis on enjoying the city's beauty and trying delicious Piedmontese dishes, validating the pleasures of travel. \textit{Step 3:} The Universalism value is touched on through the appreciation for diverse cultural heritage, fitting the persona's likely left-leaning openness to other cultures and traditions. \textit{Step 4:} The Self-Direction value is supported through the descriptive narrative that lets the persona shape his own itinerary based on what most interests him. \textit{Step 5:} The Tradition value is mildly engaged through respect for historical landmarks and traditional cuisine, honoring the cultural heritage of the city. \\
\textbf{Primal World Beliefs.} \textit{Step 1:} The answer presents Turin as Enticing, full of cultural wonders and architectural beauty, which resonates with a young traveler likely seeing the world as full of opportunity. \textit{Step 2:} The Abundant world belief is supported through the variety of attractions described, suggesting that the city has plenty to offer for any interest. \textit{Step 3:} The Alive world belief is engaged through the descriptive language that brings the city's history and culture to life rather than treating it as static information. \textit{Step 4:} The Interconnected world belief is touched on through the way the answer ties together architecture, history, museums, and food as part of a coherent cultural experience. \textit{Step 5:} The Safe world belief is implicitly supported through the welcoming, inviting tone that frames Turin as a destination ready to be enjoyed. \\
\textbf{Moral Foundations.} \textit{Step 1:} The Care/Harm foundation is mildly engaged through the warm framing of cultural appreciation and the joys of travel, which the persona likely finds inviting. \textit{Step 2:} The Sanctity/Degradation foundation is touched on through the reverent treatment of historical and architectural landmarks, treating cultural heritage as something valuable. \textit{Step 3:} The Fairness/Cheating foundation is barely relevant here, but the answer's even-handed treatment of various site types reflects a balanced approach that respects different visitor interests. \textit{Step 4:} The Loyalty/Betrayal foundation is mildly engaged through the appreciation for local cuisine and traditions, honoring the regional identity of Piedmont. \textit{Step 5:} The neutral, culturally appreciative tone fits the persona's left-leaning orientation by celebrating cultural diversity and respect for heritage without being preachy or ideological. \\ \midrule

\rowcolor{red!8}\textsc{\method{} Rationale --- Rejected} \\
\textbf{Personality.} \textit{Step 1:} The bullet-point list format is clear and organized, which appeals to conscientiousness, but the briefer descriptions may feel less engaging for a young persona with high openness who likely wants vivid storytelling. \textit{Step 2:} The simpler structure is functional but lacks the descriptive depth that often appeals to a culturally curious graduate looking for meaningful travel guidance. \textit{Step 3:} The list misses sensory or culinary highlights that would engage extraversion through the social pleasures of food and dining, which often matter to young travelers. \textit{Step 4:} The agreeable side is engaged through the inviting tone, but the formal list structure feels less warm than a flowing narrative would feel for the persona. \textit{Step 5:} The straightforward format keeps things simple, suiting low neuroticism, but the lighter detail may leave the persona wanting more atmosphere and texture in the recommendations. \\
\textbf{Cultural Dimensions.} \textit{Step 1:} The list format reflects strong uncertainty avoidance, which is helpful for planning, but may feel slightly less aligned with the more flowing cultural narrative that often resonates with Brazilian sensibilities. \textit{Step 2:} The absence of local cuisine recommendations underplays the indulgence-oriented framing that suits both Brazilian culture and a young traveler's enjoyment of food experiences. \textit{Step 3:} The brief, structured descriptions feel slightly more masculine in cultural orientation, focusing on efficient information delivery rather than the feminine warmth of sensory cultural immersion. \textit{Step 4:} The long-term orientation toward cultural appreciation is present but less developed without descriptive richness about why each site matters. \textit{Step 5:} The cultural depth feels slightly thinner overall, which may be less satisfying for a culturally engaged young traveler seeking immersive experiences. \\
\textbf{Human Values.} \textit{Step 1:} The Stimulation value is supported through the variety of attractions, but the brief descriptions reduce the excitement and richness of each option. \textit{Step 2:} The Hedonism value is weakly engaged because the answer omits cuisine and sensory pleasures that often bring travel to life. \textit{Step 3:} The Universalism value is touched on through the cultural variety, but the lighter treatment doesn't as strongly engage the persona's likely left-leaning appreciation for cultural depth. \textit{Step 4:} The Self-Direction value is partially supported through the list format that lets the persona pick what interests him, but the lack of detail provides less basis for informed choice. \textit{Step 5:} The Tradition value is engaged through historical sites, but without the cultural color of cuisine and atmosphere, the heritage appreciation feels less rich. \\
\textbf{Primal World Beliefs.} \textit{Step 1:} The Enticing world belief is engaged through the variety of attractions, but the brief descriptions reduce the sense of wonder and excitement that a more vivid response would convey. \textit{Step 2:} The Abundant world belief is supported through the list of multiple sites, similar to the alternative, but feels less rich without descriptive context. \textit{Step 3:} The Alive world belief is weakly engaged because the list format treats sites more as information points than as living parts of the city's culture. \textit{Step 4:} The Interconnected world belief is partially supported through the variety of cultural elements, but without flowing description, the connections between sites feel less developed. \textit{Step 5:} The Safe world belief is implicitly engaged through the orderly format, but the answer feels less inviting than a more narrative approach. \\
\textbf{Moral Foundations.} \textit{Step 1:} The Care/Harm foundation is mildly engaged through the appreciative tone, but the briefer treatment feels less warm than a more descriptive approach. \textit{Step 2:} The Sanctity/Degradation foundation is touched on through references to historical sites, but with lighter framing about their significance. \textit{Step 3:} The Fairness/Cheating foundation is not particularly relevant here, but the balanced list of site types reflects fair treatment of different visitor interests. \textit{Step 4:} The Loyalty/Betrayal foundation is weakly engaged because the answer omits local cuisine and traditions that would honor regional identity more strongly. \textit{Step 5:} The neutral tone fits the persona's left-leaning orientation, but the moral framing of cultural appreciation feels lighter and less integrated than a richer treatment. \\

\end{longtable}
\normalsize
\twocolumn

\onecolumn
\footnotesize
\begin{longtable}{p{0.97\linewidth}}
\caption{\textbf{Test-time rationale examples for India.}}
\label{tab:test_india} \\
\toprule \endfirsthead
\multicolumn{1}{l}{\textit{Table~\ref{tab:test_india} continued.}} \\ \toprule \endhead
\midrule \multicolumn{1}{r}{\textit{continued on next page}} \\ \endfoot
\bottomrule \endlastfoot

\textbf{Persona.} You are an 18--34-year-old male from India with Indo-Aryan background, post-secondary graduate (some or complete graduate degree), and a centrist political orientation. \\ \midrule
\textbf{Question.} I want to improve my leadership skills and set specific goals for professional development. \\ \midrule
\textbf{Chosen Answer.} To improve your leadership skills and set specific goals for professional development, the following steps can be taken: (1) Self-assessment: evaluate your current strengths and weaknesses to identify areas needing improvement. (2) Set specific goals: define SMART (Specific, Measurable, Achievable, Relevant, Time-bound) goals---e.g., ``improve team communication by 20\% within the next 6 months.'' (3) Participate in leadership development programs: join training, workshops, or online courses. (4) Find a mentor: learn from experienced leaders who can guide you and evaluate your progress. (5) Seek regular feedback: get input from your team and colleagues. (6) Maintain flexibility: adapt your strategy based on new challenges. \\ \midrule
\textbf{Rejected Answer.} To improve leadership skills and set specific goals for professional development, evaluate your current situation and identify your weaknesses. Then set SMART (Specific, Measurable, Achievable, Relevant, Time-bound) goals---for example, aim to improve communication with your team over the next 6 months or successfully complete a specific project. Regular self-assessment and feedback are also important to track progress and adjust your strategy as needed. \\ \midrule

\rowcolor{gray!10}\textsc{Generic-\method{} Rationale --- Chosen} \\
\textit{Step 1:} The answer provides a clear, well-structured set of steps for leadership development, which is likely to appeal to a young graduate seeking organized career guidance. \textit{Step 2:} The variety of recommendations---self-assessment, SMART goals, programs, mentorship, feedback---gives the persona multiple paths for growth that he can adapt to his situation. \textit{Step 3:} The mentorship and team feedback elements align with the persona's likely cultural preference for learning through community and respected figures. \textit{Step 4:} The neutral, professional tone fits the persona's centrist political orientation by avoiding any ideological framing and staying focused on practical career development. \textit{Step 5:} Overall, the answer balances structure, variety, and relational elements, making it a strong fit for a young Indian graduate looking to grow as a leader. \\ \midrule

\rowcolor{gray!10}\textsc{Generic-\method{} Rationale --- Rejected} \\
\textit{Step 1:} The answer covers the basics of leadership development but with less detail and variety, which may feel limiting for a young graduate seeking comprehensive career guidance. \textit{Step 2:} The lack of mentorship and program recommendations underplays the cultural value placed on learning through community and experienced figures in the persona's Indian context. \textit{Step 3:} The narrower set of suggestions makes the path to growth feel less rich, which may not fully energize a young professional with multiple ambitions. \textit{Step 4:} The neutral tone fits the persona's centrist orientation, but the briefer framing offers less actionable depth for someone serious about building leadership skills. \textit{Step 5:} Overall, while the answer is clear and includes the core SMART goals concept, its limited scope makes it a slightly weaker fit for a young Indian graduate looking for a fuller, more well-rounded leadership development plan. \\ \midrule

\rowcolor{blue!8}\textsc{\method{} Rationale --- Chosen} \\
\textbf{Personality.} \textit{Step 1:} The persona is a young Indian male with a graduate degree, which suggests high openness to experience and a willingness to learn new skills. The answer's variety of steps---self-assessment, mentorship, programs, feedback---appeals to his curiosity and desire to grow. \textit{Step 2:} The structured, numbered list of steps reflects strong conscientiousness, giving the persona a clear plan he can follow. As a young professional building his career, he likely values this kind of organized approach. \textit{Step 3:} The recommendation to find a mentor and seek feedback taps into a balanced extraversion profile, encouraging social engagement and learning from others. The persona may find this energizing and useful for his career growth. \textit{Step 4:} The agreeable tone of seeking feedback and adapting based on others' input fits well with the persona's likely cooperative side, showing that leadership involves collaboration rather than just personal achievement. \textit{Step 5:} The reminder to stay flexible reduces pressure and supports a persona with low to moderate neuroticism who appreciates adaptable strategies rather than rigid plans that could cause stress when things change. \\
\textbf{Cultural Dimensions.} \textit{Step 1:} The recommendation to find a mentor reflects a moderate power distance acceptance, which fits Indian cultural norms that value learning from experienced and respected figures. The persona likely finds this familiar and useful in his career context. \textit{Step 2:} The emphasis on community-based learning---through mentors, programs, and team feedback---aligns with collectivist cultural elements common in Indian society, where personal growth often happens within social networks rather than alone. \textit{Step 3:} The structured SMART goals framework supports a moderate uncertainty avoidance preference, giving the persona predictable methods for tracking progress. As a graduate, he likely appreciates this kind of organized approach. \textit{Step 4:} The focus on long-term skill development and steady improvement aligns with a long-term orientation, which fits both Indian cultural values of perseverance and the persona's likely career-building mindset. \textit{Step 5:} The balance between achievement-focused goal-setting and relational elements like mentorship and feedback reflects a moderate masculine-feminine cultural blend, giving the persona room to pursue success while also valuing relationships. \\
\textbf{Human Values.} \textit{Step 1:} The Achievement value is strongly supported through the focus on goal-setting, skill-building, and measurable progress, which fits the persona's likely drive to demonstrate competence in his career. \textit{Step 2:} The Self-Direction value is engaged through the emphasis on self-assessment and personal goal-setting, giving the persona room to define his own path while still being structured. \textit{Step 3:} The Benevolence value is supported through the mentorship and feedback recommendations, building positive relationships with others in his professional network. \textit{Step 4:} The Security value is reflected in the structured, step-by-step framework, giving the persona confidence that he is following a reliable approach to skill development. \textit{Step 5:} The recommendation to stay flexible touches on the Stimulation value, encouraging the persona to remain open to new challenges and adapt as his career evolves. \\
\textbf{Primal World Beliefs.} \textit{Step 1:} The answer presents the world of professional development as Abundant, with many resources---programs, mentors, feedback opportunities---available to support the persona's growth. \textit{Step 2:} The Just world belief is supported because the answer suggests that consistent effort, self-assessment, and feedback lead to real improvement, reinforcing the idea that hard work pays off. \textit{Step 3:} The Interconnected world belief is engaged through the emphasis on learning from mentors and team feedback, showing that growth happens within a network of relationships rather than in isolation. \textit{Step 4:} The Enticing framing of leadership development---full of programs, workshops, and learning opportunities---presents the professional world as full of possibility for a young graduate. \textit{Step 5:} The world is presented as Safe enough to experiment with new strategies, with flexibility built in to handle challenges. This supports the persona's confidence in pursuing growth. \\
\textbf{Moral Foundations.} \textit{Step 1:} The Authority/Subversion foundation is engaged through the recommendation to find a mentor, honoring the legitimate authority of experienced leaders. The persona likely respects this guidance pattern from his cultural and educational background. \textit{Step 2:} The Fairness/Cheating foundation is supported through the focus on regular feedback, ensuring that progress is honestly evaluated rather than just self-claimed. \textit{Step 3:} The Care/Harm foundation is engaged through the emphasis on building relationships and seeking input from team members, showing care for those the persona will lead. \textit{Step 4:} The Loyalty/Betrayal foundation is touched on through the team-oriented framing, encouraging commitment to one's colleagues and shared goals. \textit{Step 5:} The overall tone is balanced and non-ideological, fitting the persona's centrist political orientation by focusing on practical professional growth rather than any partisan framing. \\ \midrule

\rowcolor{red!8}\textsc{\method{} Rationale --- Rejected} \\
\textbf{Personality.} \textit{Step 1:} The answer covers the basics but with less variety, which may feel limiting for a young persona with high openness who likely wants to explore multiple paths to leadership growth. \textit{Step 2:} The conscientiousness side is partially served by the SMART goals framework, but the lack of clear step-by-step structure may feel less organized for a graduate who values methodical guidance. \textit{Step 3:} The answer's brief mention of self-assessment and feedback engages extraversion only weakly, missing the social and mentorship dimensions that often energize young professionals building their careers. \textit{Step 4:} The agreeable tone is present through the feedback recommendation, but without mentorship or team-oriented suggestions, the answer feels less collaborative than a young leader might want. \textit{Step 5:} The shorter framing keeps things simple but may feel slightly under-developed for a persona with low neuroticism who likely wants a comprehensive plan he can rely on for career growth. \\
\textbf{Cultural Dimensions.} \textit{Step 1:} The lack of mentorship recommendation underplays the moderate power distance acceptance common in Indian culture, where learning from experienced figures is highly valued. \textit{Step 2:} The absence of community-based learning suggestions like training programs or workshops feels less aligned with collectivist elements of Indian cultural traditions, which often emphasize learning in groups. \textit{Step 3:} The SMART goals framework does provide moderate uncertainty avoidance, but without additional structured steps, the answer offers less predictability than a more detailed approach. \textit{Step 4:} The long-term orientation is partially supported through the goal-setting framework, but the briefer treatment doesn't as strongly reinforce the patient, steady cultivation of leadership that the persona likely values. \textit{Step 5:} The answer leans more toward individual achievement without strongly balancing it with relational and community-focused elements, making the cultural fit slightly weaker for an Indian graduate. \\
\textbf{Human Values.} \textit{Step 1:} The Achievement value is supported through goal-setting, but the narrower set of recommendations underweights the breadth of competence-building options the persona likely seeks. \textit{Step 2:} The Self-Direction value is engaged through self-assessment, but the answer doesn't fully empower the persona with multiple paths to choose from for his independent growth. \textit{Step 3:} The Benevolence value is weakly engaged because the answer doesn't include mentorship or team-building elements, missing the relational growth dimension. \textit{Step 4:} The Security value is partially supported through the SMART framework, but the simpler structure may feel less reassuring to a persona who values comprehensive, reliable career guidance. \textit{Step 5:} The Stimulation value is mildly engaged through the mention of adjusting strategy, but the answer doesn't offer as much variety or excitement in its development paths as a richer treatment would. \\
\textbf{Primal World Beliefs.} \textit{Step 1:} The Abundant world belief is mildly underrepresented because the answer focuses on a few core practices rather than highlighting the wide range of resources available for professional growth. \textit{Step 2:} The Just world belief is supported through the SMART goals and feedback framing, showing that effort yields visible results, which the persona likely appreciates. \textit{Step 3:} The Interconnected world belief is weakly engaged because the answer focuses on individual self-assessment and goal-setting rather than learning through mentors, peers, or community. \textit{Step 4:} The Enticing framing is present but limited, since the answer doesn't paint as full a picture of the opportunities available in leadership development as a more detailed response would. \textit{Step 5:} The world is implicitly Safe in this approach, but the narrower scope may feel less expansive than the broader landscape of growth options the persona likely envisions. \\
\textbf{Moral Foundations.} \textit{Step 1:} The Authority/Subversion foundation is weakly engaged because the answer doesn't include mentorship or learning from experienced leaders, missing a key cultural and moral element. \textit{Step 2:} The Fairness/Cheating foundation is supported through the feedback element, but the briefer treatment doesn't as strongly emphasize honest evaluation as a richer response would. \textit{Step 3:} The Care/Harm foundation is partially engaged through team communication goals, but without explicit relational suggestions, the moral dimension of caring for one's team feels lighter. \textit{Step 4:} The Loyalty/Betrayal foundation is barely touched, since the answer doesn't strongly emphasize commitment to one's team or community. \textit{Step 5:} The overall tone is professional and centrist, which fits the persona, but the moral framing feels less integrated and complete compared to a fuller treatment of leadership ethics. \\

\end{longtable}
\normalsize
\twocolumn

\onecolumn
\footnotesize
\begin{longtable}{p{0.97\linewidth}}
\caption{\textbf{Test-time rationale examples for France.}}
\label{tab:test_france} \\
\toprule \endfirsthead
\multicolumn{1}{l}{\textit{Table~\ref{tab:test_france} continued.}} \\ \toprule \endhead
\midrule \multicolumn{1}{r}{\textit{continued on next page}} \\ \endfoot
\bottomrule \endlastfoot

\textbf{Persona.} You are a 35--45-year-old male from France with immigrant background, centrist political orientation, and some post-secondary education. \\ \midrule
\textbf{Question.} I own a small retail shop and need help writing a social media post to announce a new campaign, including special promotions and limited-time offers. \\ \midrule
\textbf{Chosen Answer.} Focus on highlighting the community aspect of your shop by emphasizing how these promotions and limited-time offers contribute to the well-being of the local community. You can mention partnerships with local brands, initiatives supporting local artisans, or social responsibility actions your shop undertakes. This approach helps strengthen the emotional connection with your customers and promotes a positive image of your business. \\ \midrule
\textbf{Rejected Answer.} Create a post that highlights the exceptional savings your customers can enjoy through special promotions and limited-time offers. You can include details about products on sale, discount percentages, and deadlines for the offers. Don't forget to incorporate attractive visual elements such as images or videos of the products to capture your followers' attention. \\ \midrule

\rowcolor{gray!10}\textsc{Generic-\method{} Rationale --- Chosen} \\
\textit{Step 1:} The answer focuses on community connection and local partnerships, which is likely to appeal to a small shop owner who depends on neighborhood relationships for his business success. \textit{Step 2:} The recommendations around supporting local artisans and social responsibility align with French cultural values around community and quality, fitting the persona's setting. \textit{Step 3:} The persona's immigrant background may make him especially appreciative of community-focused framing that builds belonging and emotional ties with customers. \textit{Step 4:} The neutral, ethical tone respects the persona's centrist political orientation by focusing on universally appreciated values like community and care without ideological framing. \textit{Step 5:} Overall, the answer balances brand-building, community focus, and ethical business practices, making it a strong fit for a middle-aged French shop owner planning a meaningful campaign. \\ \midrule

\rowcolor{gray!10}\textsc{Generic-\method{} Rationale --- Rejected} \\
\textit{Step 1:} The answer focuses on savings and visual appeal, which is practical but may feel less meaningful for a small shop owner who likely values community connection and authentic brand-building. \textit{Step 2:} The transactional framing underweights the relational and community-focused values that often resonate with French cultural traditions and immigrant business experiences. \textit{Step 3:} The recommendations around deadlines and discount percentages are useful but feel less complete than a strategy that includes local partnerships and social responsibility. \textit{Step 4:} The neutral tone fits the persona's centrist orientation, but the narrow promotional focus offers less depth for someone building a long-term community presence. \textit{Step 5:} Overall, while the answer is clear and actionable, its lighter community focus makes it a slightly weaker fit for a middle-aged French shop owner with immigrant roots seeking a meaningful campaign approach. \\ \midrule

\rowcolor{blue!8}\textsc{\method{} Rationale --- Chosen} \\
\textbf{Personality.} \textit{Step 1:} The persona is a middle-aged French male shop owner with an immigrant background, which suggests he values community connection and authenticity. The answer's emphasis on local community ties resonates well with his likely agreeable nature that prioritizes meaningful relationships with customers. \textit{Step 2:} The recommendation to highlight partnerships with local brands and artisans appeals to openness to experience, encouraging the persona to think creatively about how his shop fits into the broader local ecosystem. \textit{Step 3:} The conscientious side is supported through the structured approach of building a positive brand image step by step, which helps the persona plan a thoughtful marketing strategy. \textit{Step 4:} The community-focused framing engages a moderate extraversion profile, encouraging social and relational engagement with customers and local partners, which suits a small business owner who likely thrives on community connections. \textit{Step 5:} The answer reduces pressure by focusing on relationship-building rather than aggressive sales tactics, fitting a persona with low to moderate neuroticism who likely prefers sustainable and ethical marketing strategies over pushy promotions. \\
\textbf{Cultural Dimensions.} \textit{Step 1:} The emphasis on local community and supporting local artisans reflects a collectivist cultural orientation, which fits well with French cultural values around community, terroir, and supporting local craftsmanship. \textit{Step 2:} The persona's immigrant background may give him a particularly strong appreciation for community connection, which the answer honors through its focus on emotional ties between business and neighborhood. \textit{Step 3:} The framing of social responsibility actions aligns with feminine cultural dimensions valuing care and quality of life over pure competition, which fits French and European cultural sensibilities around ethical business. \textit{Step 4:} The long-term orientation is reflected in building lasting customer relationships through community engagement rather than one-time discount-driven transactions, supporting sustained brand loyalty. \textit{Step 5:} The moderate uncertainty avoidance is engaged through the steady, relationship-focused approach that gives the persona a reliable strategy for building a strong reputation in his neighborhood. \\
\textbf{Human Values.} \textit{Step 1:} The Benevolence value is strongly supported through the focus on community wellbeing and supporting local artisans, which fits the persona's likely care for those he sees in his daily business life. \textit{Step 2:} The Universalism value is engaged through the broader concern for community welfare and social responsibility, reflecting a generally caring approach to business that respects all stakeholders. \textit{Step 3:} The Tradition value is partially supported through the appreciation for local artisans and craftsmanship, honoring the cultural heritage that gives a small business its character. \textit{Step 4:} The Self-Direction value is engaged through the encouragement for the persona to define his own brand story rooted in community, giving him room to express his unique business identity. \textit{Step 5:} The Security value is reflected in the long-term reputation-building approach, helping the persona create a stable and trusted business presence in his neighborhood. \\
\textbf{Primal World Beliefs.} \textit{Step 1:} The answer presents the world as Interconnected, where the shop, customers, local artisans, and broader community are all linked in meaningful ways. The persona likely sees his business this way, embedded in a web of relationships. \textit{Step 2:} The Abundant world belief is supported through the recognition that there are many local partners and community resources to draw upon, reinforcing a sense of plenty within the local ecosystem. \textit{Step 3:} The Alive world belief is engaged through the framing of customer connections as emotional and meaningful rather than purely transactional, suggesting that business interactions carry real human weight. \textit{Step 4:} The Safe world belief is touched on through the focus on building trust and a positive image, presenting the local community as a reliable and supportive context for the business. \textit{Step 5:} The Just world belief is mildly engaged through the idea that ethical, community-focused business behavior leads to a positive reputation and customer loyalty. \\
\textbf{Moral Foundations.} \textit{Step 1:} The Care/Harm foundation is strongly engaged through the focus on community wellbeing and supporting local artisans, expressing genuine concern for the people around the business. \textit{Step 2:} The Loyalty/Betrayal foundation is supported through the emphasis on emotional connection with customers and partnerships with local brands, building a sense of mutual commitment within the neighborhood. \textit{Step 3:} The Fairness/Cheating foundation is mildly engaged through the social responsibility framing, suggesting ethical business practices that treat all stakeholders fairly. \textit{Step 4:} The Authority/Subversion foundation is weakly but positively engaged through the respect for local craftsmanship and traditional artisan practices that anchor the community. \textit{Step 5:} The neutral tone fits the persona's centrist orientation by avoiding polarizing framing while still engaging meaningful values around community and ethical business conduct. \\ \midrule

\rowcolor{red!8}\textsc{\method{} Rationale --- Rejected} \\
\textbf{Personality.} \textit{Step 1:} The answer focuses on savings and discounts, which is practical but may feel less meaningful for a persona who likely values authentic community connection in his business. \textit{Step 2:} The conscientious side is partially served through specific advice about deadlines and discount percentages, but the narrow focus on transactional details may feel less complete than a broader strategy. \textit{Step 3:} The recommendation to use attractive visuals appeals to openness, but the overall framing leans toward a generic promotional approach that lacks the relational depth a small shop owner often values. \textit{Step 4:} The agreeable side is weakly engaged because the answer focuses on extracting customer attention through deals rather than building genuine emotional connection with the local community. \textit{Step 5:} The aggressive promotional framing may feel slightly stress-inducing for a persona with low neuroticism who likely prefers steady, relationship-based business growth over urgent sales tactics. \\
\textbf{Cultural Dimensions.} \textit{Step 1:} The focus on individual savings and discounts leans toward an individualist cultural orientation, which may feel slightly less aligned with the collectivist community-focused values often important in French culture and to immigrant business owners. \textit{Step 2:} The transactional approach underplays the cultural appreciation for craftsmanship and local heritage that often shapes how French shops build their identities. \textit{Step 3:} The high-energy promotional framing reflects a more masculine cultural orientation focused on competition and visible success, which may feel less aligned with the persona's likely preference for steady, quality-focused business. \textit{Step 4:} The short-term orientation of limited-time offers and quick discounts feels less aligned with a long-term orientation toward building lasting customer relationships and brand loyalty. \textit{Step 5:} The lack of community framing reduces the cultural fit for an immigrant shop owner who likely sees his business as embedded in a meaningful neighborhood context. \\
\textbf{Human Values.} \textit{Step 1:} The Achievement value is supported through the focus on attracting customers and driving sales, but this narrower business framing underweights the broader values the persona likely brings to his work. \textit{Step 2:} The Benevolence value is weakly engaged because the answer doesn't strongly address customer or community wellbeing beyond extracting their attention with deals. \textit{Step 3:} The Universalism value is barely touched, since the answer doesn't engage with social responsibility or broader community welfare. \textit{Step 4:} The Hedonism value is mildly supported through the appeal of savings and exciting deals, but this is a thin engagement compared to richer relational values. \textit{Step 5:} Overall, the answer leans heavily on Achievement and Hedonism, leaving Benevolence, Universalism, and Tradition underrepresented---an imbalance that may feel less aligned with the persona's likely value mix. \\
\textbf{Primal World Beliefs.} \textit{Step 1:} The Abundant world belief is engaged through the focus on plentiful deals and savings, but this is a more transactional form of abundance that lacks the relational richness of community connection. \textit{Step 2:} The Just world belief is mildly supported through the implicit promise that good deals attract loyal customers, but this feels somewhat shallow compared to deeper trust-building. \textit{Step 3:} The Interconnected world belief is weakly engaged because the answer focuses on the shop's individual promotions rather than its embeddedness in the local community. \textit{Step 4:} The Enticing framing comes through in the visual appeal and deal-driven excitement, but it leans more toward stimulation than meaningful connection. \textit{Step 5:} The world is presented as Safe enough through the structured promotional approach, but the absence of relational framing reduces the depth of the worldview the answer supports. \\
\textbf{Moral Foundations.} \textit{Step 1:} The Care/Harm foundation is weakly engaged because the answer focuses on capturing attention through deals rather than expressing care for customers and community. \textit{Step 2:} The Loyalty/Betrayal foundation is touched on through customer attraction, but without community framing, the moral commitment to neighborhood ties is largely absent. \textit{Step 3:} The Fairness/Cheating foundation is mildly supported through clear discount transparency, but this feels limited compared to broader fairness in business practices. \textit{Step 4:} The Authority/Subversion foundation is barely engaged, since the answer doesn't address respect for local craftsmanship or community traditions. \textit{Step 5:} Overall, the moral framing leans heavily on transactional appeal, which is valid but less aligned with the relational, community-respecting moral orientation the persona likely brings to his work. \\

\end{longtable}
\normalsize
\twocolumn

\onecolumn
\footnotesize
\begin{longtable}{p{0.97\linewidth}}
\caption{\textbf{Test-time rationale examples for Italy.}}
\label{tab:test_italy} \\
\toprule \endfirsthead
\multicolumn{1}{l}{\textit{Table~\ref{tab:test_italy} continued.}} \\ \toprule \endhead
\midrule \multicolumn{1}{r}{\textit{continued on next page}} \\ \endfoot
\bottomrule \endlastfoot

\textbf{Persona.} You are an 18--34-year-old male from Italy with Indo-European background, post-secondary graduate (some or complete graduate degree), and a centrist political orientation. \\ \midrule
\textbf{Question.} I am attending a baby naming party for a friend's child and need help choosing a gift and writing a sincere greeting card that expresses congratulations and wishes for the new arrival in the family. \\ \midrule
\textbf{Chosen Answer.} A practical gift such as a set of baby clothes or toys would be greatly appreciated. For the greeting card, you could write: ``I am so happy for you and your family! The arrival of a baby is an incredible moment, and I hope you can enjoy every part of this journey. My wish is that your child is blessed with a life full of laughter, adventures, and success, and that your family is always supported and loved.'' \\ \midrule
\textbf{Rejected Answer.} A personalized gift such as a photo album or a decorative item for the baby's nursery can be a great choice. For the greeting card, you could write: ``Congratulations on the arrival of your little treasure! I hope this new chapter of your life is filled with joy, love, and serenity. My wish is that your child grows up healthy, happy, and curious, and that your family is always united and full of love.'' \\ \midrule

\rowcolor{gray!10}\textsc{Generic-\method{} Rationale --- Chosen} \\
\textit{Step 1:} The answer provides a practical gift suggestion and a warm, sincere greeting card, which is likely to appeal to a young graduate looking for thoughtful but uncomplicated guidance. \textit{Step 2:} The recommendation of baby clothes or toys offers something useful that the friend's family will actually appreciate, fitting the persona's practical mindset. \textit{Step 3:} The greeting card focuses on family support and love, aligning with Italian cultural values around close family bonds. \textit{Step 4:} The neutral, heartfelt tone respects the persona's centrist orientation by avoiding any ideological framing while still feeling sincere. \textit{Step 5:} Overall, the answer balances practicality and warmth, making it a strong fit for a young Italian graduate attending a baby naming party for a close friend. \\ \midrule

\rowcolor{gray!10}\textsc{Generic-\method{} Rationale --- Rejected} \\
\textit{Step 1:} The answer provides a thoughtful greeting card and a personalized gift idea, but the decorative item or photo album may feel less practical than something the family can use day-to-day. \textit{Step 2:} The card's focus on serenity and curiosity is sweet but may feel slightly more abstract than wishes tied to family support and tangible joy. \textit{Step 3:} The personalized gift adds an element of taste-matching that may feel less reliable than a clearly useful gift for a young graduate who likely prefers straightforward choices. \textit{Step 4:} The neutral tone fits the persona's centrist orientation, but the slightly more poetic framing may feel less natural than a simpler, warmer expression. \textit{Step 5:} Overall, while the answer is sincere and warm, its more abstract gift and card style make it a slightly weaker fit for a young Italian graduate seeking practical and heartfelt guidance for a friend's baby celebration. \\ \midrule

\rowcolor{blue!8}\textsc{\method{} Rationale --- Chosen} \\
\textbf{Personality.} \textit{Step 1:} The persona is a young Italian male graduate, which suggests practical thinking and an appreciation for useful, well-thought-out gifts. The recommendation of baby clothes or toys aligns with his conscientious side that values gifts which serve real needs. \textit{Step 2:} The straightforward gift suggestion appeals to a balanced openness profile---open enough to consider thoughtful options, but practical enough to choose something the family will actually use. The persona likely finds this groundedness appealing. \textit{Step 3:} The warm and friendly tone of the greeting card fits well with a moderate extraversion profile, expressing happiness and connection without being overly dramatic. The persona may appreciate this natural, sincere voice. \textit{Step 4:} The mention of wishingthe family to be ``always supported and loved'' reflects high agreeableness, focusing on care for the family unit rather than just the child. The persona, attending a friend's celebration, likely values this relational warmth. \textit{Step 5:} The answer keeps things simple and warm, suiting a persona with low to moderate neuroticism who likely doesn't want to overthink the gift-giving process and prefers easy, sincere expressions of support. \\
\textbf{Cultural Dimensions.} \textit{Step 1:} The recommendation of practical baby clothes or toys reflects an appreciation for usefulness, which fits a balanced cultural approach common in Italian society where thoughtful practicality is valued in family gifts. \textit{Step 2:} The greeting card's emphasis on family support and love aligns with collectivist cultural elements common in Italian culture, where family bonds and intergenerational support are deeply valued. \textit{Step 3:} The straightforward tone of the gift recommendation reflects moderate uncertainty avoidance, giving the persona a clear and reliable suggestion he can act on without overthinking. \textit{Step 4:} The wish for a life of ``laughter, adventures, and success'' balances feminine cultural values like joy and warmth with masculine values like achievement, fitting a young Italian graduate's likely blended outlook. \textit{Step 5:} The long-term orientation comes through in the wish for ongoing family support and a successful life journey, reflecting a future-focused perspective that resonates with the persona's likely stage in life. \\
\textbf{Human Values.} \textit{Step 1:} The Benevolence value is strongly supported through the focus on family wellbeing and lasting support, which fits the persona's likely care for his friend's family during this meaningful moment. \textit{Step 2:} The Achievement value is touched on through the wish for the child's success, acknowledging the persona's likely respect for personal accomplishment as a graduate himself. \textit{Step 3:} The Hedonism value is engaged through the wish for laughter and joyful adventures, validating the celebration of life's pleasures during this happy occasion. \textit{Step 4:} The Security value is reflected in the wish for the family to be ``always supported and loved,'' offering a sense of stability that the persona likely values for his friend's growing family. \textit{Step 5:} The Universalism value is mildly engaged through the broad, generous wishes for the child's life, reflecting a generally caring outlook that fits the persona's centrist orientation. \\
\textbf{Primal World Beliefs.} \textit{Step 1:} The answer presents the world as Enticing and full of opportunity for the new child, with wishes for adventures and a life full of laughter. The persona likely sees the world this way for a friend's newborn. \textit{Step 2:} The Abundant world belief is supported through the generous wishes for laughter, adventures, and success, suggesting that life has plenty to offer the new arrival. \textit{Step 3:} The Interconnected world belief is engaged through the focus on family support and love, framing the child's life as embedded in a network of caring relationships. \textit{Step 4:} The Safe world belief comes through in the wish for the family to be ``always supported and loved,'' presenting the world as a place where family bonds provide reliable security. \textit{Step 5:} The Just world belief is implicitly supported through the wish that the child enjoys success, reflecting a sense that effort and goodness are rewarded in life. \\
\textbf{Moral Foundations.} \textit{Step 1:} The Care/Harm foundation is strongly engaged through the warm wishes for the child and the family, expressing genuine concern for their wellbeing. The persona likely shares this caring orientation toward his friend. \textit{Step 2:} The Loyalty/Betrayal foundation is supported through the focus on family unity and ongoing support, honoring the importance of close bonds. \textit{Step 3:} The Fairness/Cheating foundation is mildly engaged through the wish for success, suggesting that the child's efforts will be justly rewarded throughout life. \textit{Step 4:} The Sanctity/Degradation foundation is touched on in the framing of birth as ``an incredible moment,'' treating the arrival of a new life with appropriate reverence. \textit{Step 5:} The neutral tone respects the persona's centrist orientation by avoiding any ideological framing and focusing on universally appreciated values like family, love, and joy. \\ \midrule

\rowcolor{red!8}\textsc{\method{} Rationale --- Rejected} \\
\textbf{Personality.} \textit{Step 1:} The suggestion of a personalized photo album or decorative item appeals to openness, but it may feel less practical for a young graduate who likely prefers gifts the family will use day-to-day. \textit{Step 2:} The conscientious side may find the decorative gift suggestion less functional than baby clothes or toys, since decorative items don't serve a clear daily need for the family. \textit{Step 3:} The greeting card's tone is warm and sincere, fitting a balanced extraversion profile, but the framing leans more toward poetic expression than the natural, conversational warmth a young persona might prefer. \textit{Step 4:} The agreeable tone is present through the focus on family unity, but the slightly more flowery language may feel less genuine compared to a simpler, more direct expression of support. \textit{Step 5:} The decorative gift idea may add slight overthinking for a persona with low neuroticism who likely prefers easy, useful gift choices over ones that require more curation and personal taste judgment. \\
\textbf{Cultural Dimensions.} \textit{Step 1:} The personalized photo album or nursery decoration suggestion reflects an individualist cultural framing focused on personal aesthetic, which may feel slightly less aligned with the practical family-focused orientation common in Italian gift-giving traditions. \textit{Step 2:} The greeting card emphasizes serenity and curiosity, which is lovely but may feel slightly less aligned with the warmth and joyful expressiveness often valued in Italian family celebrations. \textit{Step 3:} The decorative gift suggestion introduces moderate uncertainty since it depends on matching the family's taste, which may feel less reliable than a clearly useful gift like clothes or toys. \textit{Step 4:} The card leans toward a more feminine cultural orientation with its focus on joy, love, and serenity, which is fine but may not as strongly balance the achievement-oriented values the persona may also appreciate. \textit{Step 5:} The long-term orientation is supported through wishes for the child's growth, but the slightly more abstract framing feels less grounded than wishes tied to concrete family support. \\
\textbf{Human Values.} \textit{Step 1:} The Benevolence value is supported through the warm wishes for the family, but the gift choice of decorative items may feel less directly caring than something practical the family can use. \textit{Step 2:} The Achievement value is more weakly engaged because the card focuses on serenity and happiness rather than success, which a young graduate may also value. \textit{Step 3:} The Hedonism value is touched on through wishes for joy and love, but the more contemplative framing feels less celebratory than a more lively expression. \textit{Step 4:} The Security value is partially supported through wishes for family unity, but the decorative gift adds an element of uncertainty about whether the family will love the choice. \textit{Step 5:} The Universalism value is mildly engaged through the broad caring wishes, but the abstract decorative gift suggestion makes the overall package feel slightly less grounded in tangible care. \\
\textbf{Primal World Beliefs.} \textit{Step 1:} The Enticing world belief is engaged through wishes for joy and curiosity, but the framing feels slightly more contemplative than fully celebratory of life's possibilities. \textit{Step 2:} The Abundant world belief is partially supported through the wishes for the child's growth, but the decorative gift choice limits the sense of practical generosity. \textit{Step 3:} The Interconnected world belief is engaged through the focus on family unity, similar to the alternative, so this belief is reasonably well-supported. \textit{Step 4:} The Safe world belief is touched on through wishes for serenity, but the decorative gift adds a small layer of uncertainty about whether it will fit the family's space. \textit{Step 5:} The Just world belief is less directly engaged because the card focuses more on emotional states than on the rewarded effort and success that often resonate with a young graduate. \\
\textbf{Moral Foundations.} \textit{Step 1:} The Care/Harm foundation is engaged through warm wishes for the child and family, similar to the alternative, so this moral element is reasonably well-supported. \textit{Step 2:} The Loyalty/Betrayal foundation is touched on through the focus on family unity, but the framing feels slightly less grounded in ongoing tangible support. \textit{Step 3:} The Fairness/Cheating foundation is weakly engaged because the card focuses on emotional states rather than success or earned outcomes. \textit{Step 4:} The Sanctity/Degradation foundation is gently engaged through references to a ``little treasure,'' treating the child's arrival as precious, though the framing feels slightly more sentimental than reverent. \textit{Step 5:} The neutral tone fits the persona's centrist orientation, but the moral framing feels slightly less integrated and complete than a more grounded treatment of family values. \\

\end{longtable}
\normalsize
\twocolumn


\end{document}